\documentclass[mnsc]{informs3b} 

\OneAndAHalfSpacedXI 

\usepackage{natbib}
 \bibpunct[, ]{(}{)}{,}{a}{}{,}%
 \def\bibfont{\small}%

\usepackage[resetlabels]{multibib}
\newcites{sec}{References}

\newcites{tworevse}{References}
\newcites{tworevae}{References}
\newcites{tworevone}{References}
\newcites{tworevtwo}{References}
\newcites{tworevthree}{References}

\TheoremsNumberedThrough     
\ECRepeatTheorems

\EquationsNumberedThrough    

\MANUSCRIPTNO{ISR-2022-125.R3}

\usepackage{caption}

\usepackage[bottom]{footmisc}
\usepackage{afterpage}

\usepackage{xcolor}

\usepackage[bookmarks, hidelinks]{hyperref}
\hypersetup{
	bookmarksnumbered=true,     
	bookmarksopen=true,
	bookmarksopenlevel=2,    
	colorlinks=true,
	linkcolor={red!50!black},
	citecolor={blue!50!black},
	urlcolor={blue!80!black},
}

\usepackage{bookmark}
\usepackage{etoolbox}
\usepackage{tocloft}
\makeatletter
\pretocmd\endfigure{%
\addtocontents{lof}{\protect{%
    \bookmark[
    rellevel=1,
    keeplevel,
    dest=\@currentHref,
    ]{Figure \thefigure: \@currentlabelname}}}%
\bookmark[
rellevel=1,
keeplevel,
dest=\@currentHref,
]{Figure \thefigure: \@currentlabelname}%
}{}{\errmessage{Patching \noexpand\endfigure failed}}
\pretocmd\endtable{%
\addtocontents{lof}{\protect{%
    \bookmark[
    rellevel=1,
    keeplevel,
    dest=\@currentHref,
    ]{Table \thetable: \@currentlabelname}}}%
\bookmark[
rellevel=1,
keeplevel,
dest=\@currentHref,
]{Table \thetable: \@currentlabelname}%
}{}{\errmessage{Patching \noexpand\endtable failed}}
\makeatother

\usepackage{tabularx}
\usepackage{multirow}
\usepackage{cases}
\usepackage{threeparttable}
\usepackage{makecell}

\usepackage{booktabs}

\usepackage{comment}

\usepackage{algorithm} 
\usepackage[noend]{algpseudocode} 
\usepackage{tikz}
\usetikzlibrary{matrix}

\makeatletter
\newcount\dirtree@lvl 
\newcount\dirtree@plvl
\newcount\dirtree@clvl
\def\dirtree@growth{%
  \ifnum\tikznumberofcurrentchild=1\relax
  \global\advance\dirtree@plvl by 1
  \expandafter\xdef\csname dirtree@p@\the\dirtree@plvl\endcsname{\the\dirtree@lvl}
  \fi
  
  \global\advance\dirtree@lvl by 1\relax
  \dirtree@clvl=\dirtree@lvl
  \advance\dirtree@clvl by -\csname dirtree@p@\the\dirtree@plvl\endcsname
  \pgf@xa=1cm\relax
  \pgf@ya=-0.6cm\relax
  \pgf@ya=\dirtree@clvl\pgf@ya
  \pgftransformshift{\pgfqpoint{\the\pgf@xa}{\the\pgf@ya}}%
  
  \ifnum\tikznumberofcurrentchild=\tikznumberofchildren
  \global\advance\dirtree@plvl by -1
  \fi
}
\tikzset{
  dirtree/.style={
    growth function=\dirtree@growth,
    growth parent anchor=south west,
    parent anchor=south west,
    every child node/.style={anchor=west},
    edge from parent path={([xshift=2ex] \tikzparentnode\tikzparentanchor) 
                           |- (\tikzchildnode\tikzchildanchor)},
  }
}
\makeatother

\usepackage[toc]{appendix}

\usepackage[nomessages]{fp}

\usepackage{enumitem}

\DeclareMathOperator{\dplus}{+\kern -0.4em+}

\newcommand{\G}{\mathcal G}
\newcommand{\M}{\mathcal X}
\newcommand{\Q}{\mathcal Q}

\newcommand{\J}{\mathcal J}

\newcommand{\I}{\mathcal I}
\newcommand{\PP}{\mathcal P}
\newcommand{\real}{\mathbb{R}}
\newcommand{\intg}{\mathbb{Z}}

\usepackage{float}
\usepackage{dsfont}
\usepackage{caption}
\usepackage{graphicx}
\usetikzlibrary{matrix}

\usepackage[toc]{appendix}

\begin{document}

\newcolumntype{L}[1]{>{\raggedright\arraybackslash}p{#1}}
\newcolumntype{C}[1]{>{\centering\arraybackslash}p{#1}}
\newcolumntype{R}[1]{>{\raggedleft\arraybackslash}p{#1}}



\RUNTITLE{Proactive Resource Request for Disaster Response}

\TITLE{Proactive Resource Request for Disaster Response: A Deep Learning-based Optimization Model}

\ARTICLEAUTHORS{%

\AUTHOR{ \large Accepted by Information Systems Research}
\AFF{}
\AUTHOR{Hongzhe Zhang\thanks{Equal Contribution}}
\AFF{School of Management and Economics and Shenzhen Finance Institute, The Chinese University of Hong Kong, Shenzhen (CUHK-Shenzhen), China} 
\AUTHOR{Xiaohang Zhao\footnotemark[1]}
\AFF{School of Information Management and Engineering,
Shanghai University of Finance and Economics, Shanghai, China}
\AUTHOR{Xiao Fang\thanks{Corresponding Author: Xiao Fang, \href{xfang@udel.edu}{xfang@udel.edu}}, Bintong Chen}
\AFF{Lerner College of Business and Economics, University of Delaware, Newark, Delaware}
}

\ABSTRACT{Disaster response is critical to save lives and reduce damages in the aftermath of a disaster. Fundamental to disaster response operations is the management of disaster relief resources. 
To this end, a local agency (e.g., a local emergency resource distribution center) collects demands from local communities affected by a disaster, dispatches available resources to meet the demands, and requests more resources from a central emergency management agency (e.g., Federal Emergency Management Agency in the U.S.). 
Prior resource management research for disaster response overlooks the problem of deciding optimal quantities of resources requested by a local agency. In response to this research gap, we define a new resource management problem that proactively decides optimal quantities of requested resources by considering both currently unfulfilled demands and future demands. To solve the problem, we take salient characteristics of the problem into consideration and develop a novel deep learning method for future demand prediction. We then formulate the problem as a stochastic optimization model, analyze key properties of the model, and propose an effective solution method to the problem based on the analyzed properties. We demonstrate the superior performance of our method over prevalent existing methods using both real world and simulated data. We also show its superiority over prevalent existing methods in a multi-stakeholder and multi-objective setting through simulations.
}



\maketitle
 
%


\section{Introduction}
\label{sec:intro}

We are living in the century of destructive disasters, which have claimed thousands of lives and caused tremendous economic losses.\footnote{See \href{https://slate.com/technology/2011/05/tornado-in-missouri-why-disasters-are-becoming-more-frequent-and-what-we-can-do-about-it.html}{https://slate.com/technology/2011/05/tornado-in-missouri-why-disasters-are-becoming-more-frequent-and-what-we-can-do-about-it.html} (last accessed on March 23, 2023)} In 2020 alone, 389 natural disasters affected 98.4 million people and costed 171.3 billion US dollars worldwide.\footnote{ See \href{https://reliefweb.int/report/world/cred-crunch-newsletter-issue-no-62-may-2021-disaster-year-review-2020-global-trends-and}{https://reliefweb.int/report/world/cred-crunch-newsletter-issue-no-62-may-2021-disaster-year-review-2020-global-trends-and} (last accessed on March 23, 2023)} 
Disaster management is critical to reduce adverse impacts of disasters and thus has drawn attentions from fields such as Information Systems and Operations Management \citep{park_disaster_2015, gupta_disaster_2016}. 
The life cycle of disaster management consists of four phases: mitigation, preparedness, response, and recovery \citep{altay_orms_2006}. Disaster mitigation and preparedness take place before a disaster. The former aims to prevent disasters or reduce their impacts whereas the latter makes preparations before a disaster strikes. Disaster response and recovery occur after a disaster. 
Disaster recovery focuses on restoring the affected community to the status before a disaster. In particular, disaster response plays an irreplaceable role in reducing fatalities and damages caused by a disaster \citep{fiedrich_optimized_2000}.
It encompasses all operations conducted to save lives and reduce damages in the aftermath of a disaster, including providing resources to disaster-affected people, conducting search and rescue missions, and ensuring continuity of critical services.\footnote{See \href{https://www.fema.gov/sites/default/files/2020-05/CPG_101_V2_30NOV2010_FINAL_508.pdf}{https://www.fema.gov/sites/default/files/2020-05/CPG\_101\_V2\_30NOV2010\_FINAL\_508.pdf} (last accessed on March 23, 2023)} Fundamental to these operations is resource management.

\vspace{-0.2cm}
\begin{figure}[H]
	\FIGURE
	{\includegraphics[scale=0.6]{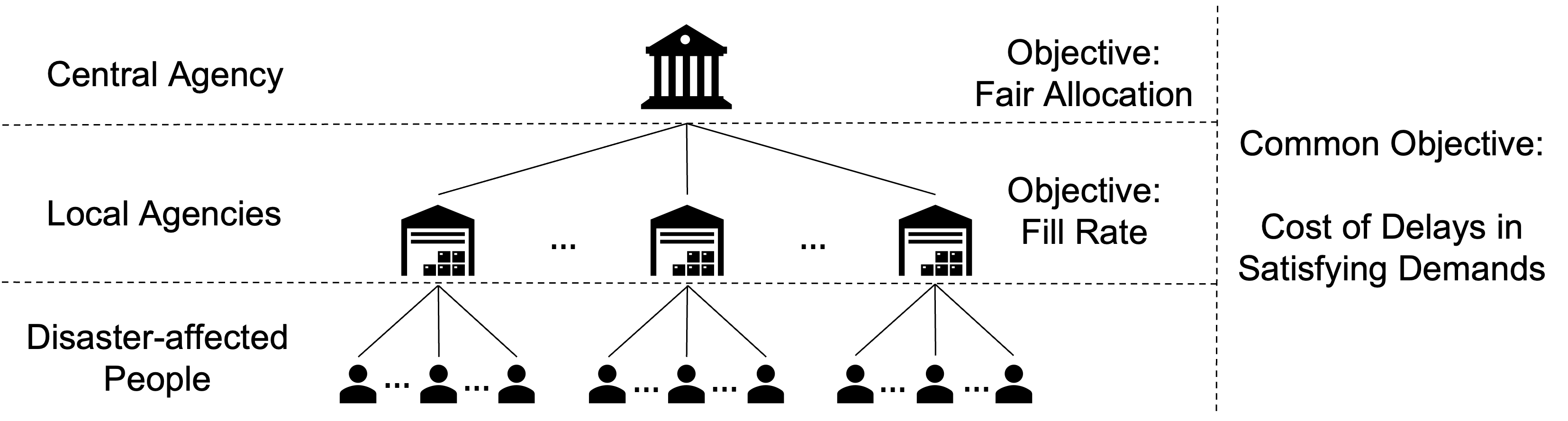}}
	{A Humanitarian Resource Management System for Disaster Response. \label{fg:system}}
	{}
\end{figure}
\vspace{-0.4cm}

The goal of resource management for disaster response is to timely and effectively meet the demands for disaster relief resources from disaster-affected people   \citep{fiedrich_optimized_2000,noyan_stochastic_2016}.
To this end, a humanitarian resource management system for disaster response is a multi-stakeholder and multi-objective system \citep{abbasi_don_2019,abbasi_call_2021}.
Figure~\ref{fg:system} depicts a typical humanitarian resource management system, which consists of three types of stakeholders: a central agency (e.g., Federal Emergency Management Agency or FEMA in the U.S.), a number of local agencies (e.g., local emergency resource distribution centers), and disaster-affected people. 
A local agency collects demands from its local community, dispatches available resources to meet the demands through its distribution points, and requests more resources from the central agency  \citep{vanajakumari_integrated_2016}.
Based on the requests from local agencies, the central agency allocates disaster-relief resources to each of them. 
All stakeholders agree on the common objective of humanitarian resource management, i.e., minimizing the cost of delays in satisfying the demands of disaster-affected people; meanwhile, each type of stakeholders also could have its own objective \citep{abbasi_don_2019,abbasi_call_2021}. 
As illustrated in Figure~\ref{fg:system}, a local agency pays attention to its fill rate, i.e., ability of fulfilling resource requests from its local community \citep{noyan_stochastic_2016} 
whereas the central agency is concerned about fair allocation, which aims to ensure no local agency is systematically disadvantaged in the allocation of disaster-relief resources \citep{bertsimas_efficiency_2012,huang_modeling_2015}. 

Existing resource management studies for disaster response focus on problems such as how to allocate and transport resources from the central agency to local agencies and how to locate distribution points in a disaster-affected area \citep{fiedrich_optimized_2000, tzeng_multi-objective_2007,noyan_stochastic_2016}.
These studies commonly assume the types and quantities of disaster relief resources requested by a local agency as given and overlook the problem of the local agency deciding the optimal quantity of requested resources for each resource type. A straightforward solution to this problem is to set quantities of requested resources as quantities of currently unfulfilled demands \citep{huang_modeling_2015}.
For example, if demands for 3 units of food and 2 units of shelters are not fulfilled, a local agency would request 3 units of food and 2 units of shelters from the central agency.
Such solution is a reactive solution (i.e., in reaction to what has already happened). This solution neglects future demands arrived between resource request time and resource arrival time,
which inevitably leads to significant time delays of satisfying these future demands. The cost incurred during disaster response is an increasing function of time delay \citep{holguin-veras_appropriate_2013};
thus, significant time delays result in high costs and even loss of lives. For example, \citet{fiedrich_optimized_2000} show that the survival rate of disaster victims decreases in time, and \citet{petrovic_dynamic_2012} find that suppression resource for wildfire is effective only if it reaches the fire in time. Therefore, it is important to decide types and quantities of requested resources based on not only currently unfulfilled demands but also future demands arrived between resource request time and resource arrival time.

In response, we introduce a new resource management problem for disaster response, namely proactive resource request problem. It aims to decide types and quantities of requested resources by considering both currently unfulfilled demands and future demands, with the objective of minimizing the cost incurred by delays in satisfying these demands.
To solve the problem, we must tackle two methodological challenges: how to predict future demands and how to decide an optimal resource request plan based on currently unfulfilled demands and predicted future demands. To address these challenges, we take salient characteristics of the problem into consideration and develop a novel deep learning method for future demand prediction. We then formulate the problem as a stochastic optimization model, analyze key properties of the model, and propose an effective solution method to the problem.
Extensive empirical analyses with real world and simulated data demonstrate that our method significantly outperforms prevalent existing methods in reducing the cost of delays in satisfying demands. Moreover, through simulations, we also show superior performance of our method over these benchmarks in a multi-stakeholder and multi-objective setting as depicted in Figure~\ref{fg:system}.

\section{Related Work} \label{sec:rw}

\subsection{Resource Management for Disaster Response}\label{sec:rw:rm}

Existing studies on resource management for disaster response focus on four research problems: resource allocation, resource distribution, facility location, and resource procurement \citep{gupta_disaster_2016}.
Resource allocation aims to decide the optimal assignment of resources to disaster-affected areas.
For example, \citet{fiedrich_optimized_2000} propose a dynamic combinatorial optimization model that allocates technical response resources after a major earthquake, with the objective of minimizing the number of fatalities during the search and rescue period of disaster response. 
Unlike resource allocation, resource distribution plans routes to transport resources from distribution centers to affected people \citep{tzeng_multi-objective_2007}.
To this end, \citet{tzeng_multi-objective_2007} adopt a multi-objective programming method to design a relief delivery system with the goal of minimizing the transportation cost and time and maximizing the  satisfaction of demands simultaneously. 
Facility location decides how to locate distribution points in a disaster-affected region for the provision of disaster relief resources \citep{noyan_stochastic_2016}.
In this vein,
\citet{noyan_stochastic_2016} propose last mile relief networks, which determine the locations and capacities of resource distribution points while considering the uncertainties in demands and transportation network conditions. The goal of relief networks is to maximize the expected total accessibility, which measures the ease of access to resources. Lastly, resource procurement develops policies for procuring disaster relief resources from suppliers. 
An exemplar study by \citet{natarajan_inventory_2014} proposes an optimal resource procurement policy that minimizes the shortages of fulfilling demands for disaster relief resources subject to the constraint of disaster relief funds. 

Recent studies have solved several research problems elaborated above simultaneously.
For example, \citet{rennemo_three-stage_2014} present a three-stage stochastic programming model, each stage of which models the problems of facility location, initial resource allocation, and last mile resource distribution, respectively. To better capture real world disaster response situations, the model treats the availability of vehicles, demands for disaster relief resources, and the condition of transportation infrastructure as stochastic model elements. \citet{ahmadi_humanitarian_2015} propose a location-routing model, which decides locations of central warehouses in the disaster preparedness phase and determines locations of local distribution centers and routes of vehicles in the disaster response phase. The model introduces novel constraints to model the golden time period of disaster response. Like \citet{rennemo_three-stage_2014}, \citet{vanajakumari_integrated_2016} develop an integrated logistics model that solves the facility location, resource allocation, and resource distribution problems simultaneously. They also provide empirical insights that are useful for logistic managers during the disaster response phase. 

\subsection{Inventory Control Models}
\label{sec:rw:ic}
Depending on how future demands are modeled, Inventory Control (IC) models can be classified into classical IC models and data-driven IC models. Classical IC models assume that certain form of information on future demands is known to decision-makers and the optimal IC policy is then derived accordingly. 
For example, the Economic Order Quantity model assumes that the quantities of future demands are known and deterministic and then obtains a closed-form solution of optimal order quantity \citep[p.146]{zipkin_foundations_2000}. To better capture demand uncertainty, more IC models assume that future demand quantities follow a known probability distribution, such as normal distribution \citep{mieghem_newsvendor_2002}
and Poisson distribution \citep{ guijarro_exact_2012}. To develop inventory control policies that minimize the inventory cost over multiple periods, time series models have been used to model period-correlated demands.
For example, \citet{gilbert_arima_2005} characterizes demands using the ARIMA model and theoretically analyzes the bullwhip effect in a multistage supply chain.  In addition, a number of studies consider demand uncertainties caused by the evolution of model environment.
An exemplar study by \citet{hu_s_2016} adopts a Markov-modulated demand model, which assumes that demands are generated from one period to another by an evolving ``state'' factor characterized by a discrete-time Markov chain. Different from the literature reviewed above that focus on demand quantity information solely, a number of IC models assume richer information about future demands: both the quantity and arrival time of a future demand. They often model future demands using continuous stochastic arrival processes, such as a compound Poisson process \citep{zhao_analysis_2009}, Markov modulated Poisson process \citep{arts_repairable_2016}, and time-dependent phase-type process \citep{ nasr_continuous_2018}. For example, \citet{arts_repairable_2016} investigate a repairable stocking system, in which the demand of a repairable item follows a Markov modulated Poisson process and failed parts can be expedited with extra cost to shorten waiting time. 

Classical IC models focus on the structure of optimal IC policies under various distribution or process assumptions of future demands, which may not accurately reflect true future demands. In response, various data-driven IC models are proposed more recently to predict future demand quantities or the distribution of future demand quantities from historical demand data and then decide the optimal IC policy based on the prediction.  Notably, time series methods, such as Croston-like methods and ARIMA, have gained popularity in the data-driven IC literature for predicting future demand quantities \citep{syntetos_forecasting_2009}.
Besides, much literature aims to predict future demand quantities using linear regression. 
A recent study by \citet{ban_big_2019} formulates a data-driven newsvendor problem. They estimate future demand quantities using a linear regression on observed problem features and propose an empirical risk minimization approach to solve the problem. Built on the work by \citet{ban_big_2019}, \citet{oroojlooyjadid_applying_2020} apply a deep learning method (i.e., a multilayer perceptron) to predict future demand quantities.  
IC policies have also been derived based on the estimated distribution of future demand quantities. In case reasonable prior knowledge of demand distribution is available, the Bayesian method can be applied to update the posterior demand distribution as new demands arrive (e.g., \citealp{chen_bounds_2010}).   In addition, some studies focus on demand learning for the joint dynamic pricing and inventory control problem.
For example,  \citet{chen2022dynamic} and \citet{chen2023optimal} treat the expected demand quantity over each planning period as a function of price.  In particular, the former fits the demand-price curve via a parametric function, while the latter uses a non-parametric setting.
Besides demand quantity forecasting, predicting future demand quantiles has drawn attentions since safety stock levels can be computed from demand quantiles directly.  In this vein, \citet{taylor_forecasting_2007} proposes an exponentially weighted quantile regression to predict daily sales with high volatility and skewness. \citet{fricker_applying_2000} apply bootstrapping to estimate future demand distribution and set reorder points based on quantiles derived from the distribution.

\subsection{Demand Forecasting and Temporal Point Process}\label{sec:rw:tpp}

Demand forecasting can be formulated as a regression problem, where training data consists of demand explanatory variables and corresponding demand quantities. Once training data set is constructed, statistical and machine learning methods, such as regression models and tree-based methods, can be applied to it to predict future demands. When demand data are sequentially recorded at equal time intervals (e.g., per day), time series forecasting methods are widely used (e.g., \citealp{ arunraj_hybrid_2015}).
Recently, using deep learning methods to predict demands has become a growing area of research due to their advanced learning and forecasting capabilities. In particular, RNN-based methods, such as LSTM \citep{tan_ultra_2019} and deep sequence-to-sequence method \citep{yi_electric_2021}, have been employed to predict demand quantities over discrete time periods in various domains. Besides, graph convolution networks have been utilized to predict transportation demands, e.g., demands for ride-hailing and bike-sharing \citep{ye_coupled_2021}.

Predicting future demands can also be formulated as a problem of predicting a sequence of future events, each of which corresponds to a demand. 
To solve this problem, temporal point process (TPP) can predict future demand events with continuous arrival times.
Because of its solid mathematical foundation and excellent predictive performance, TPP has become the dominant technique for predicting a sequence of future events \citep{shchur_intensity-free_2020}. TPP is a stochastic process that models the arrival times and marks of a sequence of events \citep{du_recurrent_2016}.
For example, in our study, TPP models the arrival times of demand events, as well as the mark (i.e., types and quantities of requested disaster relief resources) associated with each demand. TPP has wide use cases, including 
intermittent demand forecasting \citep{turkmen_intermittent_2019}, health event prediction \citep{enguehard_neural_2020}, 
and identification of similar sequences \citep{gupta_learning_2022}. TPP can be characterized by its conditional intensity function (CIF), which models the instantaneous occurrence rate of an event conditioning on the history of previous events. Early TPP models usually make restrictive parametric assumptions on CIFs.
For instance, the classic Hawkes process assumes that the arrival of an event temporarily raises the conditional intensity of the process \citep{rizoiu_hawkes_2017}.
However, the assumptions made by early TPP models might not reflect the reality and therefore they suffer from model misspecification errors \citep{du_recurrent_2016}. To overcome these limitations, recent studies develop CIFs based on deep learning models \citep{du_recurrent_2016,mei_transformer_2022}. 

In particular, Recurrent Neural Network (RNN) and its variants have been widely used to encode event history because of their capabilities of capturing nonlinear dependency of an event on its previous events. 
In this vein, \cite{du_recurrent_2016} employ a discrete-time RNN to embed past events as a vector and then design a CIF with its parameters derived from the vector, thereby capturing the dependency of an event on its past events. \cite{mei_neural_2017} construct their CIF based on vectors that summarize past events through a continuous-time LSTM, which enables their proposed TPP to model more complicated event arrival patterns.
\cite{xiao_learning_2018} also use LSTM and propose to train TPP with Wasserstein loss, which measures the distance between an event sequence in the training data and its prediction. 
Built on \citet{xiao_learning_2018}, \cite{yan_improving_2018} add one additional loss that measures the mean square error between the number of events in an event sequence and that in its prediction. 
\cite{deshpande_long_2021} partition a training event sequence into equal time intervals and count the number of events in each time interval. They propose to train TPP by maximizing the likelihood of counts in these time intervals and the likelihood of observing each individual event in the training event sequence.

Recently, a number of studies employ the Transformer architecture
to encode event history \citep{mei_transformer_2022}. For example, \cite{enguehard_neural_2020} and \cite{mei_transformer_2022} utilize Transformer blocks to represent a sequence of historical events as an embedding matrix, and then evaluate the CIF at any future timestamp by using the timestamp as the query to summarize the embedding matrix as an intensity score. Similar idea is adopted by \cite{zhang_self-attentive_2020} and \cite{zuo_transformer_2020}, but they assume a simpler dependency of the CIF on a future timestamp. Another recent trend is to characterize TPP using a conditional probability density function,
instead of a CIF. For example, \cite{shchur_intensity-free_2020} model the conditional probability density of event interarrival times with a mixture of log-normal distributions, of which the parameters are derived from the vector representation of past events.  
\cite{gupta_modeling_2022} investigate the problem of training TPP with incomplete observations of historical events. They propose to treat missing events as latent variables, and model the dynamics of observed events and that of missing events via two coupled TPP models. 

Our literature review suggests several research gaps. First, existing resource management studies for disaster response overlook the problem of deciding optimal quantities of resources requested by a local agency. To address this gap, we propose a new resource management problem that proactively decides the optimal quantity of requested resources for each resource type based on both currently unfulfilled demands and future demands. Second, even though our proactive resource request problem is conceptually related to IC models, adapting an existing IC model to our problem is not an effective approach. 
More specifically, classical IC models make assumptions on the distribution or process of future demands, which may not reflect true future demands in reality; data-driven IC models focus on predicting future demand quantities over discrete time periods.
We model future demand quantities and arrival times as a  stochastic arrival process in a continuous timeline and learn the process from historical demand data.
Third, to solve the proactive resource request problem, it is necessary to predict future demands. Most demand forecasting methods focus on forecasting future demand quantities or predicting future demand quantities over discrete time periods (e.g., \citealp{arunraj_hybrid_2015,yi_electric_2021}). However, to effectively solve the problem, both quantities and continuous arrival times of future demands need to be modeled
and predicted. Moreover, the distribution of demands is non-stationary during the outbreak of a disaster, which further complicates the task of predicting quantities and arrival times of future demands. To this end, a TPP model is applicable for modeling non-stationary demand arrival processes. However, existing TPP models are not well-suited for our problem because of its following characteristics. (1) The problem requires to model both the type and quantity of requested resources. However, existing TPP mark embedding functions are designed to model either the type or the quantity of an event. (2) It is essential to incorporate the importance scores of different types of resources into the prediction of future demands, because different types of resources are not equally important for disaster relief. Existing TPP models overlook the heterogeneous importance of disaster relief resources in demand prediction. (3) The quantities of different resource types requested in a demand could be correlated. Existing TPP models fail to capture this correlation. 
In response, we develop a novel TPP model that features three methodological novelties, each of which addresses one of the salient characteristics of the problem. 
Fourth, our proposed resource request problem requires a novel model and solution method. Accordingly, we develop a stochastic optimization model for the problem and identify its key properties. Armed with these properties, we propose an effective solution method to the problem.

\section{Problem Formulation} \label{sec:pf}

Consider a disaster (e.g., a flood or earthquake) that occurs at time $T_{-}$. In response to it, a local agency collects demands for disaster relief resources, dispatches available resources to meet the demands, and requests more resources from a central emergency management agency. Each demand is described by its arrival time as well as types (e.g., food, shelter) and quantities of resources requested. 
Let $T$ denote the time that the local agency requests resources from the central agency. At this time, the local agency is aware of remaining resources and unfulfilled demands and decides the types and quantities of resources that need to be requested from the central agency. 
Let $R_k$ denote the quantity of remaining type $k$ resources and $U_k$ be the quantity of unfulfilled demands for type $k$ resources at time $T$, where $k=1,2,\dots,K$ and $K$ is the number of resource types. It is clear that either $R_k$ or $U_k$ is zero, i.e., $R_k\times U_k=0$, $k=1,2,\dots,K$. We use a simple example to illustrate the calculation of $R_k$ and $U_k$.
 
\begin{example}
\label{exp:exist_demand}
Consider a disaster that causes the following demands by time $T$.
\vspace{-0.2cm}
\begin{table}[H]
\centering
\resizebox{\textwidth}{!}{
\begin{tabular}{|c|c|c|c|}
\hline
Demand ID & Demand Time & Type of Resources Requested & Quantity of Resources Requested \\ \hline
1         & 2021/07/18 12:04:33 & Shelter          & 4                              \\ \hline
2         & 2021/07/18 16:38:26 & Shelter          & 3                              \\ \hline
2         & 2021/07/18 16:38:26 &Food                       & 6                            \\\hline
\end{tabular}}
\end{table}
\vspace{-0.7cm}
\noindent A local agency with resources of $5$ units of shelters and $10$ units of food 
dispatches 4 units of shelters
to fulfill demand 1. Subsequently, it distributes 6 units of food and the remaining 1 unit of shelter to meet demand 2. Clearly, 2 units shelters requested in demand 2 can not be fulfilled by available resources and there are 4 units of food left over. We thus have $U_{shelter}=2$,  $R_{shelter}=0$,  $U_{food}=0$, and  $R_{food}=4$ at time $T$.
\label{ex:r_u_compute}
\end{example} 

At time $T$, for each resource type $k$ to be requested from the central agency, the local agency needs to decide its requested quantity $x_k$, $k=1,2,\dots,K$. A simple way is to decide $x_k$ in a reactive manner and set it according to unfulfilled demands $U_k$, for $k=1,2,\dots,K$ \citep{huang_modeling_2015}.
However, resources requested at time $T$ will arrive at a later time $T_{+}$, and there will be demands occurring during the period of $T$ to $T_{+}$. Reactive decisions at time $T$ only consider currently unfulfilled demands $U_k$ but neglect future demands in the period of $T$ to $T_{+}$, which result in significant time delays of satisfying these future demands. These time delays in turn lead to high costs and even loss of lives as the cost incurred during disaster response is an increasing function of time delay \citep{holguin-veras_appropriate_2013}.
Therefore, it is important to develop a proactive method that decides $x_k$ based on not only currently unfulfilled demands but also future demands in the period of $T$ to $T_{+}$.
In addition, the transportation capacity of moving resources from the central agency to the local agency is not unlimited and it is constrained due to damaged transportation infrastructure and the shortage of transportation equipment \citep{gossler_how_2019}.
Consequently, transportation capacity $W$ constraints quantities of resources that can be received (and hence requested) by the local agency. We are now ready to define the proactive resource request problem.

\begin{definition} [\bf  \underline{P}roactive  \underline{R}esource  \underline{R}equest (PRR) Problem] A local agency requests disaster relief resources from a central agency at time $T$. Given the quantity $R_k$ of remaining resources and the quantity $U_k$ of unfulfilled demands for each resource type $k$, $k=1,2,\dots,K$, as well as the  transportation capacity $W$ at time $T$, the local agency needs to decide the quantity $x_k$ of requested resources for each resource type $k$ such that the cost incurred by delays in satisfying unfulfilled demands by time $T$ and future demands arrived in the period of $T$ to $T_{+}$ is minimized while satisfying the constraint of the transportation capacity.\label{def:prr}
\end{definition}

Generally speaking, out study belongs to the predictive and prescriptive analytics research in the Information Systems (IS) field. Over the years, IS scholars have developed predictive analytics methods that predict future outcomes or prescriptive analytics methods that make optimal decisions informed by predictions to solve a diverse set of critical business and societal problems 
\citep{abbasi_metafraud_2012,fang_predicting_2013,lin_first_2021,fang_prescriptive_2021,zhu_deep_2021-1}. Our study adds to this stream of IS research with a new research problem and a novel method.

\section{Solution Method} \label{sec:method}
\begin{figure}[t]
	\FIGURE
	{\includegraphics[scale=0.17]{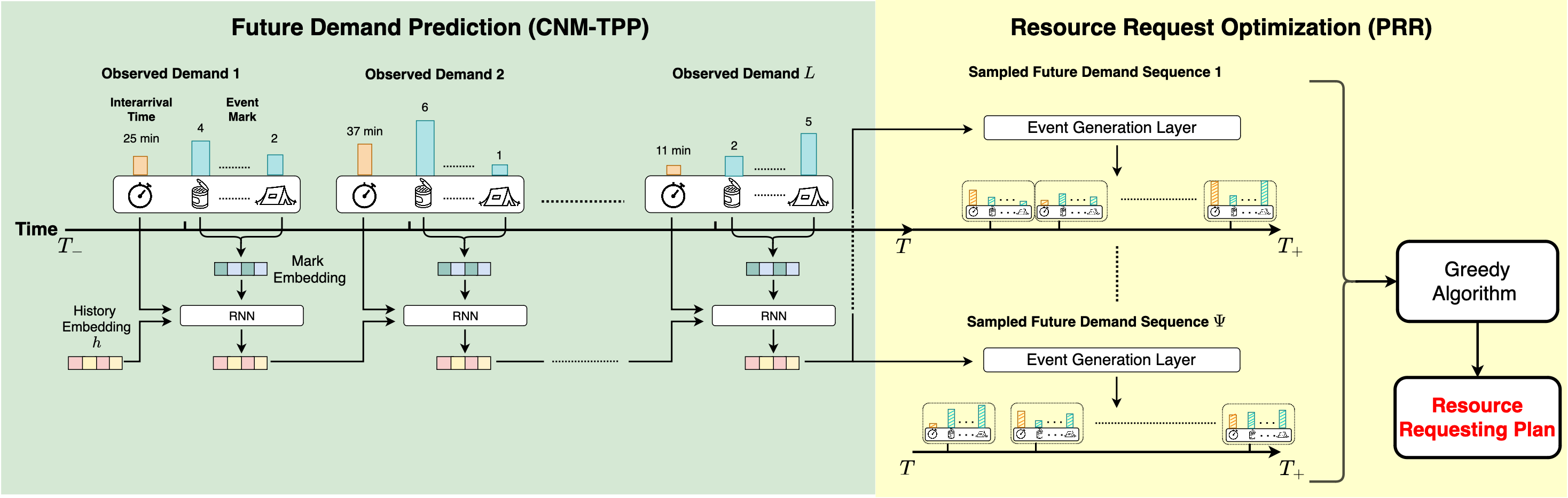}}
	{Overall Architecture of the Proposed Solution Method. \label{fg:model}}	{}
\end{figure}

To solve the PRR problem, we need to tackle its two subproblems: (1) how to predict future demands in the period of $(T, T_{+}]$ and (2) how to decide an optimal resource requesting plan in consideration of both currently unfulfilled demands and predicted future demands. 
Figure~\ref{fg:model} illustrates the overall architecture of the proposed solution method. As shown, the method consists of two components, each of which solves its corresponding subproblem.
The CNM-TPP component takes the observed (historical) demands in the period of $[T_-,T]$ as inputs and summarizes these demands as a dense embedding vector. 
The event generation layer is designed to sample future demands given the embedding vector of historical demands.
Next, the PRR component iteratively invokes the event generation layer to sample sequences of future demands. These sampled future demands, together with currently unfulfilled demands, are taken as inputs by our proposed greedy algorithm to decide the resource requesting plan. We detail each component in Sections \ref{sec:method:ctpp} and \ref{sec:method:prrm}, respectively. 

\subsection{Future Demand Prediction} \label{sec:method:ctpp}

We propose the \underline{C}ost-aware \underline{N}eural \underline{M}arked \underline{T}emporal \underline{P}oint \underline{P}rocess (CNM-TPP), which learns a model from observed demands in the period of $[T_{-}, T]$ to predict future demands in the period of $(T, T_{+}]$. CNM-TPP treats each demand as an event. Accordingly, we can represent observed demands in the period of $[T_{-}, T]$ as an event sequence $S=<e_1, e_2, ...,e_L>$, where $L$ denotes the number of events (i.e., demands) in the period and the $i$-th event $e_i=(t_i, m_i)$ is described by its occurrence time $t_i$ (i.e., demand time) as well as its mark $m_i$ indicating the types and quantities of the resources requested by the demand. To learn a model from $S$, it is critical to compute the likelihood $p(S)$ of observing $S$ \citep{ shchur_neural_2021}.
Specifically, we can factorize $p(S)$ as
\begin{equation}
\begin{aligned}
\label{eq:p_S}
p(S) &= P(\tau_{L+1}>T-t_L) \prod_{i=1}^{L} p_{\tau}\big( \tau_i | \mathcal{H}_{i-1} \big) p_{m}\big( m_i | \mathcal{H}_{i-1} \big),
\end{aligned}
\end{equation}
where $\mathcal{H}_{i-1}=<e_1, e_2, ...,e_{i-1}>$ denotes the sequence of past events before the $i$-th event $e_i$, $i=1,2,\dots,L$, and $\mathcal{H}_0=\emptyset$. We define interarrival time $\tau_i=t_i - t_{i-1}$ and $p_{\tau}\big( \tau_i | \mathcal{H}_{i-1} \big)$ is the probability density that the interarrival time between the $i$-th event and the $(i-1)$-th event is $\tau_i$ conditioning on past events $\mathcal{H}_{i-1}$; $p_{m}\big( m_i | \mathcal{H}_{i-1} \big)$ represents the probability mass that the mark of the $i$-th event is $m_i$ conditioning on $\mathcal{H}_{i-1}$. 
Since interarrival time $\tau_{L+1}=t_{L+1} - t_L$, we have $P(\tau_{L+1}>T-t_L)=P(t_{L+1}>T)$, which is the probability that the $(L+1)$-th event occurs after time $T$. The derivation of Equation \eqref{eq:p_S} is given in Appendix \ref{apd:add_detail:eq1}.

Based on the structure of Equation \eqref{eq:p_S}, we design CNM-TPP as a deep neural network with three building blocks: (1) a history embedding layer that summarizes the information contained in $\mathcal{H}_{i-1}$, (2) an event generation layer that specifies probability functions $p_{\tau}()$ and $p_{m}()$, and (3) the learning objective of CNM-TPP. In what follows, we elaborate each building block in turn.

\subsubsection{History Embedding} \label{sec:method:ctpp:he}  
The objective of this layer is to embed the information contained in $\mathcal{H}_{i-1}$ as a numeric vector $h_{i-1} \in \real^{n_e}$, where hyperparameter $n_e$ is the embedding size. Given the sequential nature of the events in $\mathcal{H}_{i-1}$, it is natural to embed it with a recurrent neural network (RNN) \citep{du_recurrent_2016, shchur_intensity-free_2020}. Therefore, we have:
\begin{equation}
    \label{eq:hist_emb}
    \begin{aligned}
    h_{i-1} &= \text{RNN}(h_{i-2}, e_{i-1}) \\
    &= 
    \max\{ W^{(h)} h_{i-2} + w^{(t)} \tau_{i-1} + W^{(m)}f_m(e_{i-1}) + b^{(h)}, \ 0 \},
    \end{aligned}
\end{equation}
where activation function $\max\{v_1, v_2\}$ compares vectors $v_1$ and $v_2$ element-wisely and returns the larger one on each dimension, $h_{i-2}$ is the embedding vector of $\mathcal{H}_{i-2}$, $\tau_{i-1}$ is the interarrival time between events $e_{i-2}$ and $e_{i-1}$, and $(W^{(h)} \in \real^{n_e \times n_e}, w^{(t)} \in \real^{n_e}, W^{(m)} \in \real^{n_e \times n_e}, b^{(h)}\in \real^{n_e})$ are learnable parameters of the RNN model. Mark embedding function $f_m()$ represents the mark of $e_{i-1}$ as a numeric vector. 

Existing TPPs model either the type or the quantity of an event \citep{turkmen_intermittent_2019, enguehard_neural_2020,  shchur_neural_2021}. 
However, an event mark in the PRR problem contains not only the type(s) of requested resources but also the quantity of each requested resource type.
To accommodate this characteristic of the PRR problem, we
represent an event mark using a vector $a$ with $K$ entries and its $k$-th entry $a_k$ denotes the requested units of type $k$ resources, $k=1,2,\dots,K$.

 \begin{example}
 \label{ex:mark}
Consider a simple scenario of three resource types: shelter, medication, and food, i.e., $K=3$. 
In this scenario, we can represent an event mark requesting 3 units of shelters and 6 units of food with a vector $a=(3,0,6)^T$, where $a_1$, $a_2$, and $a_3$ denote the requested units of shelter, medication, and food respectively. 
\end{example}

To summarize the information in an event mark, we need to embed $a_k$ for $k=1,2,\dots,K$. To that end, we define an embedding function $f_{\text{q}}()$ adapted from \cite{vaswani_attention_2017}. 
The function embeds $a_k$ (i.e., the quantity of requested type $k$ resources) as a numeric vector of length $n_e$, and each entry $f_{\text{q}}(a_k, x)$ of the vector is defined as 
\begin{equation}
\label{eq:pos_encode}
    f_{\text{q}}(a_k, x) = \sin \Big( a_k /10000^{x/n_e} \Big)
\end{equation}
where $\sin()$ is the sine function and $x=1,2,\dots,n_e$. Function $f_{\text{q}}()$ has two desirable properties. First, by Equation \eqref{eq:pos_encode}, it embeds $a_k$ as a zero vector if $a_k=0$ (i.e., requesting 0 units of type $k$ resources). Second, similar values of $a_k$ yield embedding vectors carrying similar information. 

With $f_{\text{q}}()$ defined, we propose the mark embedding function $f_m()$ for the PRR problem as:
\begin{equation}
  \label{eq:me_detail}
    f_m(e_{i-1}) = \sum_{k=1}^{K} W^{(r)}_{:,k} \odot f_{\text{q}}(a^{(i-1)}_k)
\end{equation}
where $W^{(r)} \in \real^{n_e \times K}$ is the learnable embedding matrix for resource types, its $k$-th column $W^{(r)}_{:,k}$ represents resource type $k$, $k=1,2,\dots,K$, $a^{(i-1)}_k$ is the quantity of type $k$ resources requested in event $e_{i-1}$, and $\odot$ denotes element-wise multiplication. According to Equation \eqref{eq:me_detail}, the embedding of an event mark, $f_m(e_{i-1})$, is the aggregation of $K$ embeddings, each of which summarizes the information about a resource type and its requested quantity in the event as $W^{(r)}_{:,k} \odot f_{\text{q}}(a^{(i-1)}_k)$. Taken together, the parameter set of the history embedding layer is given by
\begin{equation}
    \label{eq:param_he}
    \Theta_H = \{W^{(h)}, w^{(t)}, W^{(m)}, b^{(h)}, W^{(r)} \}.
\end{equation} 

Our mark embedding function differs from existing ones, which are designed to represent either the type or the quantity of an event \citep{turkmen_intermittent_2019, enguehard_neural_2020,  shchur_neural_2021}. However, an event mark in the PRR problem contains not only the type of requested resources but also the quantity of requested resources. Therefore, the novelty of our mark embedding function is the introduction of $W^{(r)}_{:,k} \odot f_{\text{q}}(a^{(i-1)}_k)$ in Equation \eqref{eq:me_detail}. Specifically, function $f_{\text{q}}()$ is designed to capture the quantity of requested resources. It is then integrated with $W^{(r)}_{:,k}$ through element-wise multiplication to represent both the quantity and type of requested resources.

\subsubsection{Event Generation} \label{sec:method:ctpp:eg} 

Following \cite{shchur_intensity-free_2020}, we model the conditional probability density $p_{\tau}()$ of event interarrival times  as a mixture of $n_z$ log-normal distributions, where $n_z$ is a hyperparameter of our method. Specifically, we have
\begin{equation}
    \label{eq:cpdf_tau}
    p_{\tau}\big( \tau_i | \mathcal{H}_{i-1} \big) = \sum_{z=1}^{n_z} \alpha_{i,z} \frac{1}{\tau_i \sigma_{i,z} \sqrt{2\pi}} \exp\Big(- \frac{(\log \tau_i - \mu_{i,z})^2}{2 \sigma_{i,z}^2} \Big)
\end{equation}
where parameters $\alpha_{i,z}$, $\mu_{i,z}$, and $\sigma_{i,z}$ denote the weight, mean, and standard deviation of the $z$-th log-normal distribution, respectively. To model the dependency of $\tau_i$ on past events $\mathcal{H}_{i-1}$, we derive the parameters of Equation \eqref{eq:cpdf_tau} from $h_{i-1}$, the history embedding vector of $\mathcal{H}_{i-1}$.  
In particular, by following \cite{shchur_intensity-free_2020}, we have
\begin{equation}
    \label{eq:compute_gm_param}
    \begin{aligned}
    \alpha_i &= \text{Softmax} \Big(\text{MLP}_{1}(h_{i-1})\Big) \\
    \mu_i &= \text{MLP}_{2}(h_{i-1}) \\
    \sigma_i &= \exp\Big( \text{MLP}_{3}(h_{i-1}) \Big)
    \end{aligned}
\end{equation}
where vectors $\alpha_i=(\alpha_{i,1}, \alpha_{i,2},\dots,\alpha_{i,n_z})^T$, $\mu_i=(\mu_{i,1}, \mu_{i,2},\dots,\mu_{i,n_z})^T$, and $\sigma_i=(\sigma_{i,1}, \sigma_{i,2},\dots,\sigma_{i,n_z})^T$. In Equation \eqref{eq:compute_gm_param}, $\text{MLP}_{1}$, $\text{MLP}_{2}$, and $\text{MLP}_{3}$ are multi-layer perceptrons (MLP), each of which takes $h_{i-1}$ as input and outputs a vector of length $n_z$, $\text{Softmax}(\cdot)$ is the softmax function, and $\exp(\cdot)$ denotes the exponential function applied to its input element-wisely.

To specify probability mass function $p_m()$, we formulate the generation of an event mark as the generation of $K$ non-negative integers (conditioning on past events $\mathcal{H}_{i-1}$), where each integer denotes the units of a resource type requested in the event and  is modeled by a Poisson distribution. Accordingly, the probability $p_{m}\big( m_i | \mathcal{H}_{i-1} \big)$ that the mark of event $e_i$ is $m_i$ conditioning on $\mathcal{H}_{i-1}$ is given by
\begin{equation}
  \label{eq:cpmf_m}
  \begin{aligned}
    p_m\big( m_i |\mathcal{H}_{i-1} \big) 
    &= p_m\big( a^{(i)}_{1}, a^{(i)}_{2}, \dots, a^{(i)}_{K} |\mathcal{H}_{i-1} \big)  \\
    &= P\big( a^{(i)}_{1} | \mathcal{H}_{i-1} \big) P\big( a^{(i)}_{2} | \{a^{(i)}_{1}\}, \mathcal{H}_{i-1} \big) \dots P\big( a^{(i)}_{K} | \{a^{(i)}_{1},a^{(i)}_{2},\dots,a^{(i)}_{K-1}\}, \mathcal{H}_{i-1} \big) \\
    &= \prod_{k=1}^{K} P\big( a^{(i)}_{k} | a_{1:k-1}^{(i)}, \mathcal{H}_{i-1} \big)
  \end{aligned}
\end{equation}
where $a^{(i)}_{k}$ is the units of type $k$ resources requested in event $e_i$, and $a_{1:k-1}^{(i)}= \{a^{(i)}_{1},a^{(i)}_{2},\dots,a^{(i)}_{k-1}\}$ with $a_{1:0}^{(i)}=\emptyset$ for convenience. The second step of Equation \eqref{eq:cpmf_m} follows from the chain rule in probability theory.
Equation \eqref{eq:cpmf_m} captures correlations among the units of different resource types requested in a demand. For example, in a demand, the requested units of food and water might be positively correlated. In this equation, $P\big( a^{(i)}_{k} | a_{1:k-1}^{(i)}, \mathcal{H}_{i-1}\big)$ is specified as
\begin{equation}
  \label{eq:p_m_k}
  P\big( a^{(i)}_{k} | a_{1:k-1}^{(i)}, \mathcal{H}_{i-1} \big) = \text{Poisson}(a^{(i)}_{k} ; \lambda_{i,k}),
\end{equation}
where $\lambda_{i,k}$ is the mean of the Poisson distribution for modeling $a^{(i)}_{k}$. To model the dependency of $a^{(i)}_{k}$ on $\mathcal{H}_{i-1}$ and $a_{1:k-1}^{(i)}$, we derive $\lambda_{i,k}$ as 
\begin{subequations}
 \label{eq:poisson_mean}
 \begin{align}
 h_{i,k}^{(\lambda)} &= \text{RNN}_{\lambda}\Big( h_{i,k-1}^{(\lambda)}, \ W^{(r)}_{:,k-1} \odot f_{\text{q}}(a^{(i)}_{k-1}) \Big) \label{eq:poisson_mean_hist} \\
 \lambda_{i,k} &= \exp \Big( {W^{(r)}_{:,k}}^T h_{i,k}^{(\lambda)} ) \Big) \label{eq:poisson_mean_lambda}
 \end{align}
\end{subequations}
where $h_{i,1}^{(\lambda)}=h_{i-1}$ with $h_{i-1}$ being the embedding vector of $\mathcal{H}_{i-1}$. Equation \eqref{eq:poisson_mean_hist} uses a RNN layer to summarize the information in $\mathcal{H}_{i-1}$ and $a_{1:k-1}^{(i)}$ into vector $h_{i,k}^{(\lambda)}$ of length $n_e$. In this equation, we use the same idea behind Equation \eqref{eq:me_detail} to embed
the quantity $a_{k-1}^{(i)}$ of type $k-1$ resource requested in event $e_{i}$
as $W^{(r)}_{:,k-1} \odot f_{\text{q}}(a^{(i)}_{k-1})$.
Equation \eqref{eq:poisson_mean_lambda} derives $\lambda_{i,k}$ from $h_{i,k}^{(\lambda)}$ via an exponential transformation to ensure $\lambda_{i,k}>0$.\footnote{There are two implementation considerations for the probability mass function specified in Equations \eqref{eq:cpmf_m}--\eqref{eq:poisson_mean}. First, it assigns a non-zero probability to an event mark requesting no resources (i.e., $a^{(i)}_{k}=0 \text{ for }  k=1,2,\dots,K$). To remedy this issue, we can set $p_m\big( m_i |\mathcal{H}_{i-1} \big) =0$ for the case of $a^{(i)}_{k}=0 \text{ for } k=1,2,\dots,K$. We then normalize the probabilities for all other cases 
such that these probabilities sum up to $1$. Second, we can simplify the function for the situation where $a_k$ only takes one of the two values $0$ or $1$, which is discussed in Appendix \ref{apd:add_detail:mark}.} 
The parameter set for the event generation layer is given by
\begin{equation}
    \label{eq:param_eg}
    \Theta_{E} = \{ \Theta(\text{MLP}_1), \Theta(\text{MLP}_2), \Theta(\text{MLP}_3), \Theta(\text{RNN}_{\lambda}) \}
\end{equation}
where $\Theta(\text{MLP}_i)$ is the parameter set of  $\text{MLP}_i$ for $i=1,2,3$, and $\Theta(\text{RNN}_{\lambda})$ is the parameter set of $\text{RNN}_{\lambda}$.

\subsubsection{Learning Objective} \label{sec:method:ctpp:learn_obj} 

Sections \ref{sec:method:ctpp:he} and \ref{sec:method:ctpp:eg} specify a generative model of event sequences parameterized by $\Theta=\{ \Theta_H, \Theta_E \}$. 
To learn these parameters from a training event sequence $S$ observed in the period of $[T_-,T]$, a widely used strategy in the TPP literature is to minimize the negative log likelihood (NLL) of $S$ \citep{shchur_neural_2021}, which is formally defined as
\begin{equation}
    \label{eq:obj_nll}
    \text{NLL}(S|\Theta) = -\log p(S) = - \log P(\tau_{L+1} > T-t_L ) -\sum_{i=1}^{L} \log p_{\tau}\big( \tau_i | \mathcal{H}_{i-1} \big) - \sum_{i=1}^{L} \log p_{m}\big( m_i | \mathcal{H}_{i-1} \big) 
\end{equation}
where the likelihood $p(S)$ of $S$ is given by Equation \eqref{eq:p_S}, $p_{\tau}$ and $p_m$ are respectively specified by Equations \eqref{eq:cpdf_tau} and \eqref{eq:cpmf_m}, and $P(\tau_{L+1} > T-t_L )=1-\int_{0}^{T-t_L} p_{\tau}(\tau_{L+1}|\mathcal{H}_{L}) d\tau_{L+1}$. However, learning the parameters using the NLL objective only is ineffective for the PRR problem because the NLL objective fails to cover two important features of the problem. First, the PRR problem requires to predict (infer) a sequence of events occurring in the period of $(T, T_{+}]$, while the NLL objective trains a model to predict next event $e_i=(t_i, m_i)$ conditioning on the true history $\mathcal{H}_{i-1}$. Such discrepancy between model training and model inference (i.e., next event prediction at model training versus next event sequence prediction at model inference) causes the model trained solely with the NLL objective not well-suited to predict an event sequence in the period of $(T, T_{+}]$. Second, different types of resources are not equally important for disaster relief \citep{perez-rodriguez_inventory-allocation_2016}.
For example, the time delay of satisfying demands for lifesaving resources results in more severe outcome and hence incurs higher cost than the time delay of meeting demands for regular disaster relief resources. Therefore, it is necessary to incorporate the importance scores of different types of resources into the prediction of future demands (events) in the period of $(T, T_{+}]$.

To capture these two features, we introduce a learning objective in addition to the NLL objective. The introduced objective trains a model to predict next sequence of events by minimizing the expected cost-aware distance between an observed sequence and its prediction. Consider an observed sequence $S^{(i,l)}$ of $l$ events starting from event $e_i$, i.e., $S^{(i,l)}=<e_{i}, e_{i+1}, \dots, e_{i+l-1}>$, which is part of the training sequence $S$. Using the model given in Sections \ref{sec:method:ctpp:he} and \ref{sec:method:ctpp:eg}, we can sample a prediction $\tilde{S}^{(i,l)}$ of $S^{(i,l)}$ with Algorithm \ref{alg:sample_p_i_star}, where $\tilde{S}^{(i,l)}=<\tilde{e}_{i}, \tilde{e}_{i+1}, \dots, \tilde{e}_{i+l-1}>$. As shown, the algorithm takes the embedding vector $h_{i-1}$ of past events $\mathcal{H}_{i-1}$ as an input, assigns it to the history embedding vector $\hat{h}$, and initializes $\tilde{S}^{(i,l)}$ as a empty sequence (line 1). It then iteratively generates the time $\hat{t}$ (lines 3--5) and mark $\hat{m}$ (lines 6--\ref{alg:sample_p_i_star:set_e_hat}) of an event, conditioning on $\hat{h}$, updates $\hat{h}$ with the newly generated event (line \ref{alg:sample_p_i_star:compute_h}), and adds the generated event to $\tilde{S}^{(i,l)}$ (line \ref{alg:sample_p_i_star:set_e_tilde}). 
For implementation details of sampling from the mixture log-normal distribution (line \ref{alg:sample_p_i_star:draw_tau}) and from the Poisson distribution (line \ref{alg:sample_p_i_star:draw_m}), please refer to Appendix \ref{apd:add_detail:repar_trick}.

\begin{algorithm}[h]
  \caption{Sampling $\tilde{S}^{(i,l)}$}
  \label{alg:sample_p_i_star}
    \textbf{Input:} $h_{i-1}$: embedding vector of past events $\mathcal{H}_{i-1}$, $l$: sampling sequence length \\
    \textbf{Output:} $\tilde{S}^{(i,l)}$
	\begin{algorithmic}[1]
    \State Set $\hat{h}=h_{i-1}$, $\hat{t}=t_{i-1}$, and $\tilde{S}^{(i,l)}=<>$
    \For {$\hat{l} =0,1,\dots, l-1$}
    \State  Compute $(\alpha, \mu, \sigma)$ from $\hat{h}$ via Equation \eqref{eq:compute_gm_param}
    \State Draw interarrival time $\hat{\tau}$ from $\text{LogNormalMixture}(\alpha, \mu, \sigma)$ \label{alg:sample_p_i_star:draw_tau} specified by Equation \eqref{eq:cpdf_tau} 
    \State Set event time $\hat{t}=\hat{t}+\hat{\tau}$ \label{alg:sample_p_i_star:set_t}
    \For{$k=1,2,\dots,K$}
    \State  Compute mean of the Poisson distribution for resource type $k$ from $\hat{h}$ and $\hat{a}_{1:k-1}$ \\
    \hspace{1.2cm} via Equation \eqref{eq:poisson_mean}
    \State Sample $\hat{a}_k$ from the Poisson distribution specified by Equation \eqref{eq:p_m_k}  \label{alg:sample_p_i_star:draw_m}
    \EndFor
    \State Set $\hat{m}=(\hat{a}_1, \hat{a}_2, \dots, \hat{a}_K)$, and $\hat{e}=(\hat{t}, \hat{m})$ \label{alg:sample_p_i_star:set_e_hat}
    \State Compute $\hat{h}=\text{RNN}(\hat{h}, \hat{e})$ via Equation \eqref{eq:hist_emb} \label{alg:sample_p_i_star:compute_h}
    \State Set $\tilde{e}_{i+\hat{l}}=\hat{e}$,  $\tilde{S}^{(i,l)}=\tilde{S}^{(i,l)}\oplus<\tilde{e}_{i+\hat{l}}>$ \quad \quad \quad //$\oplus$: operator concatenating two sequences \label{alg:sample_p_i_star:set_e_tilde}
    \EndFor
    \State \Return $\tilde{S}^{(i,l)}$
	\end{algorithmic} 
\end{algorithm}

The cost-aware distance between an observed sequence $S^{(i,l)}$ and its prediction $\tilde{S}^{(i,l)}$ is defined as 
\begin{equation}
\label{eq:d_s}
  D_s(S^{(i,l)}, \tilde{S}^{(i,l)}) = \sum_{j=0}^{l-1} D (e_{i+j}, \tilde{e}_{i+j}),
\end{equation}
where $D (e_{i+j}, \tilde{e}_{i+j})$ denotes the cost-aware distance between an observed event $e_{i+j}$ in $S^{(i,l)}$ and its corresponding predicted event $\tilde{e}_{i+j}$ in $\tilde{S}^{(i,l)}$. In general, the cost-aware distance $D(e, \tilde{e})$ between an observed event $e$ and its prediction $\tilde{e}$ is defined as 
\begin{equation}
\label{eq:d_e}
\begin{aligned}
  D (e, \tilde{e}) &= - \sum_{k=1}^K \log \Big( (\frac{1}{c_k})^{(t - \tilde{t})^2} (\frac{1}{c_k})^{(a_k - \tilde{a}_k)^2} \Big) \\
 &=  \sum_{k=1}^K (t-\tilde{t})^2 \log c_k + \sum_{k=1}^K (a_k - \tilde{a}_k)^2 \log c_k, \\
\end{aligned}
\end{equation}
where $t$ and $\tilde{t}$ respectively denote the time of event $e$ and its prediction, $a_k$ and $\tilde{a}_k$ respectively represent the units of type $k$ resources requested by the event and its prediction, and $c_k > 1$ is the importance score of type $k$ resources and higher score means more important for disaster relief, $k=1,2,\dots,K$. 
Our design of Equation \eqref{eq:d_e} is based on the intuition that the cost-aware distance between an event and its prediction is the aggregation of their discrepancies across all resource types and the discrepancy in each resource type is assessed from two perspectives: the cost-aware time difference as measured by $(t-\tilde{t})^2 \log c_k$ and the cost-aware quantity difference as computed by $(a_k - \tilde{a}_k)^2 \log c_k$. 
Importance score $c_k$ models the degree of importance of satisfying demands for type $k$ resources. According to Equation \eqref{eq:d_e}, prediction errors on time and quantities (i.e., $(t-\tilde{t})^2$ and $(a_k - \tilde{a}_k)^2$) lead to larger $D(e, \tilde{e})$ for events requesting higher importance resources (i.e., larger $c_k$). Consequently, a model trained to minimize $D_s(S^{(i,l)}, \tilde{S}^{(i,l)})$ places more focus on reducing prediction errors for events requesting higher importance resources,
which in turn reduces the overall cost for disaster relief because time delays of satisfying requests for higher importance resources result in higher costs. In this sense, distances defined in Equations \eqref{eq:d_s} and \eqref{eq:d_e} are cost-aware distances. In addition, $D_s(S^{(i,l)}, \tilde{S}^{(i,l)})$ measures the cost-aware distance between an observed sequence and its prediction; hence, a model trained on it can effectively predict a sequence of future events. In short, our design of $D_s(S^{(i,l)}, \tilde{S}^{(i,l)})$ captures both features of the PRR problem discussed at the beginning of this subsection. 

It is more robust to sample many predictions of an observed sequence and train a model on the expected cost-aware distance between an observed sequence and its prediction. Accordingly, for an observed sequence $S^{(i,l)}$ starting from event $e_i$ with length $l$, the expected cost-aware distance between the sequence and its prediction is given by 
\begin{equation}
\label{eq:e_d_s}
    E_{\tilde{S}^{(i,l)}} \big( D_s(S^{(i,l)}, \tilde{S}^{(i,l)}) \big) = \int D_s(S^{(i,l)}, \tilde{S}^{(i,l)}) p(\tilde{S}^{(i,l)} | \mathcal{H}_{i-1} \big) d \tilde{S}^{(i,l)},
\end{equation}
where $D_s(S^{(i,l)}, \tilde{S}^{(i,l)})$ can be computed using Equation \eqref{eq:d_s} and $p(\tilde{S}^{(i,l)} | \mathcal{H}_{i-1} \big)$ denotes the density of 
$\tilde{S}^{(i,l)}$ conditioning on the observed history $\mathcal{H}_{i-1}$. The expected distance computed using Equation \eqref{eq:e_d_s} is for a given pair of starting event $e_i$ and sequence length $l$. We can further randomly pick an event in the training sequence $S$ as $e_i$ and sample a sequence length $l$.\footnote{Because our objective is to predict future events (demands) in the period of $(T, T_+]$, we empirically estimate the distribution of sequence lengths that are likely to be observed within a time interval of length $T_{+}-T$. Specifically, we count the number of events within a time interval of length $T_{+}-T$ in the training sequence $S$ and empirically compute the probability for each distinct count.}  Accordingly, we can define our proposed learning objective,  \underline{C}ost-aware \underline{S}equence \underline{D}istance (CSD), as the expected cost-aware distance between an observed sequence and its prediction, expected on $e_i$, $l$, and $\tilde{S}^{(i,l)}$:
\begin{equation} 
\label{eq:obj_acd}
  \text{CSD} (S | \Theta) = E_{e_i} E_{l} E_{\tilde{S}^{(i,l)}} \big( D_s(S^{(i,l)}, \tilde{S}^{(i,l)}) \big).
\end{equation}  
The CSD objective has no closed form solution and can be computed using Monte Carlo method
\citep{bishop_pattern_2006},
which repeatedly samples $e_i$, $l$, and ${S}^{(i,l)}$, generates a prediction $\tilde{S}^{(i,l)}$ of ${S}^{(i,l)}$ with Algorithm \ref{alg:sample_p_i_star}, and computes the cost-aware distance between ${S}^{(i,l)}$ and $\tilde{S}^{(i,l)}$ using Equation \eqref{eq:d_s} until convergence. 
Lastly, the learning object of CNM-TPP is given by
\begin{equation}
  \label{eq:obj_full}
    \mathcal{L}(S|\Theta) = \text{CSD} (S | \Theta) + \gamma \text{NLL}(S | \Theta)
\end{equation}
where $\text{NLL}(S | \Theta)$ is specified by Equation \eqref{eq:obj_nll}, and hyperparameter $\gamma > 0$ controls the relative contribution of its two component objectives.

Our proposed model, CNM-TPP, is trained
with the observed sequence $S$ of events (demands) in the period of $[T_{-}, T]$, where $S=<e_1, e_2, ...,e_L>$. Specifically, the objective $\text{NLL}(S | \Theta)$ is derived with Equation \eqref{eq:obj_nll} and the objective $\text{CSD} (S | \Theta)$ is computed according to Equation \eqref{eq:obj_acd} with Monte Carlo method. The model parameters $\Theta$
of CNM-TPP are then learned by optimizing its learning objective specified by Equation \eqref{eq:obj_full} through gradient
descent. Once trained, CNM-TPP can infer a sequence of future events (demands) in the period of $(T, T_+]$ using a procedure similar to Algorithm \ref{alg:sample_p_i_star} with two modifications. First, it takes the embedding vector $h_{L}$ of past events $\mathcal{H}_{L}$ as an input, where ${H}_{L}$ is the training sequence $S$ and $h_{L}$ can be computed using Equation \eqref{eq:hist_emb}. Second, the inferred sequence of events must be within the time interval of $(T, T_+]$. The inference procedure is given in Appendix \ref{apd:add_detail:inf}.

In comparison to existing TPPs, our proposed CNM-TPP features three methodological novelties. First, in consideration of the two critical features of the PRR problem, i.e., heterogeneous importance of disaster relief resources and prediction of future sequence of demands, we develop a novel learning objective to train our CNM-TPP model. The CSD learning objective is instantiated through our proposed Equations \eqref{eq:d_s}, \eqref{eq:d_e}, and \eqref{eq:obj_acd} as well as Algorithm \ref{alg:sample_p_i_star}. Second, we define a new mark embedding function in Equation \eqref{eq:me_detail}, which models an event mark containing both types and quantities of requested resources. Third, the event generation layer of our TPP effectively captures the correlations among different types of resources requested in a demand through Equations \eqref{eq:cpmf_m}--\eqref{eq:poisson_mean}.

\subsection{Resource Request Optimization and Solution} \label{sec:method:prrm}

The CNM-TPP model in Section~\ref{sec:method:ctpp} allows us to predict future demands between $T$ and $T_+$.  
In this section, we propose a stochastic optimization model to determine the requested quantity $x_k$ for each resource type $k$, $k=1,2,\dots, K$, based on the predicted future demands.

Let $Q_k= <(t^k_j,q^k_j), \, j=1, 2, \dots, n^k>$ be the stochastic sequence of future demands for type $k$ resources between $T$ and $T_{+}$, where $t^k_j$ is the arrival time of the $j$-th demand, $q^k_j$ is the corresponding demand quantity, and $n^k$ is the number of demand arrivals during period $(T,T_+]$. Denote $|Q_k|= \sum_{j=1}^{n^k} q^k_j$ to be the total amount of type $k$ resources requested by victims during the period.
Clearly, the cost due to time delays of meeting the demands for type $k$ resources by time $T_+$ depends on the quantity of unfulfilled demands $U_k$ or the quantity of remaining resources $R_k$ available at time $T$, future demands $Q_k$ occurring between $T$ and $T_+$, and the quantity of requested resources $x_k$. 
Denote the corresponding cost as $f_k(x_k,Q_k,U_k, R_k)$.
Let $W$ be the transportation capacity and $w_k$ be the capacity consumed while shipping one unit of resource $k$, $k=1,2,\dots, K$.
Then the proactive resource request problem can be formulated as the following stochastic optimization problem:
\begin{equation}
\begin{aligned}
\label{pb:basic}
\min_{\M} \quad & E_{\Q}(\sum_{k=1}^Kf_k(x_k,Q_k,U_k, R_k))\\
\textrm{s.t.} \quad & \sum_{k=1}^K w_kx_k\le W\\
\quad & x_k\in \intg^+,    \quad k=1,2,\dots, K,
\end{aligned}
\end{equation}
where the expectation is taken over the stochastic future  demand sequence $\Q=(Q_1,Q_2,\dots,Q_K)$ of all resource types, the decision variables $\M=(x_1,x_2,\dots,x_K)$ represent the requested quantities for resources of all types, and $\intg^+$ consists of positive integers and zero.

We take three steps to solve the above stochastic optimization problem. 
First, we define the cost function and convert the cost minimization problem to a cost reduction maximization problem. 
In the second step, since $\Q$ is a highly non-stationary stochastic process,
we propose to solve the stochastic optimization problem using demand arrivals generated by the CNM-TPP model. We show that the optimal solution based on demand generation converges uniformly to that of the stochastic optimization, as the sample size increases.
Finally, since the optimization problem with the generated demands is an integer program and by itself an NP-hard problem, we design an efficient greedy heuristic algorithm to obtain an approximate solution. To ensure the solution quality, we show that the objective function is concave and piece-wise linear w.r.t. the decision variables (resource request allocation $\M$), which allows us to provide a performance guarantee for the approximate solution.  

\subsubsection{Conversion and Simplification of Optimization Problem \eqref{pb:basic}.}
\label{sec:method:prrm:conversion}
We model the cost $f_k$ in the stochastic optimization problem \eqref{pb:basic} as the deprivation cost, which measures the economic value of human suffering because of the deprivation of vital resources \citep{holguin-veras_appropriate_2013,perez-rodriguez_inventory-allocation_2016}.  
In this study, we adopt the exponential deprivation cost widely used in post-disaster resource management models \citep{holguin-veras_appropriate_2013}:
\begin{equation}
\label{eq:depri}
	\gamma(\delta):= e^{\phi+b c\delta}-e^{\phi},
\end{equation}
where $\delta$ is the amount of time delay it takes to meet a demand, $c$ measures the importance of the demanded resource, and $\phi$ and $b$ are deprivation parameters originally defined in \citet{holguin-veras_appropriate_2013}.
Now consider a demand of one unit of type $k$ resources that arrives at time $t \in (T,T_+]$. If at time $T$, we have requested sufficient type $k$ resources, which will arrive at time $T_+$, to meet this demand, the deprivation cost would be 
$e^{\phi+bc_k(T_+-t)}-e^{\phi}$, where $c_k$ is the importance score of type $k$ resources introduced in Section~\ref{sec:method:ctpp}. 
Otherwise, the demand will not be met until some future time $\xi_k$ ($\xi_k>T_+$), such as the next delivery of resource $k$. 
In this case, the deprivation cost increases to $e^{\phi+bc_k(\xi_k-t)}-e^{\phi}$. 
Consequently, the reduction in deprivation cost, due to the unit of type $k$ resources requested at $T$, is given by:
\begin{equation}
\begin{aligned}
\label{eq:reduction}
	B^k(t)
	&=(e^{\phi+bc_k(\xi_k-t)}-e^{\phi})-(e^{\phi+bc_k(T_+-t)}-e^{\phi})\\
	&=e^{\phi-bc_kt}(e^{bc_k\xi_k}-e^{bc_kT_+}).
\end{aligned}
\end{equation}
Clearly, the deprivation cost reduction $B^k(t)$ is a monotone convex decreasing function of $t$. 
Therefore, fulfilling an earlier demand leads to a more significant cost reduction than fulling a later one, and a local agency should dispatch resources on a first come, first serve basis. 

We next construct the sequence of net demands $\tilde{Q}_k$, $k=1,2,\dots, K$, waiting to be fulfilled by time $T_+$. 
It depends on the arrivals of future demands $Q_k$, and unfulfilled demands or remaining resources at time $T$. If the quantity of unfulfilled demands $U_k \neq 0$, then $|\tilde{Q}_k|=|Q_k|+U_k$.
We obtain $\tilde{Q}_k$ by merging unfulfilled demands with future demands $Q_k$.
If the quantity of remaining resources $R_k \neq 0$, then $|\tilde{Q}_k|=|Q_k|-R_k$. We form $\tilde{Q}_k$ by removing the first $R_k$ units of demands from $Q_k$. 
We denote the sequence of net demands as 
 $\tilde{Q}_k= <(\tilde{t}^k_j,\tilde{q}^k_j), \, j=1, 2, \dots, \tilde{n}^k>$, where $\tilde{n}^k$ is the number of demands in $\tilde{Q}_k$.
The following example illustrates the construction of $\tilde{Q}_k$.

\begin{example}
\label{exp:future_demand}
Continue with Example~\ref{exp:exist_demand}. Consider the following future demands between $[T,T_{+}]$.
\vspace{-0.25cm}
\begin{table}[H]
\centering
{
\begin{tabular}{|c|c|c|c|}
\hline
Demand ID & Demand Time & Type of Resources Requested & Quantity of Resources Requested\\ \hline
3         & 2021/07/18 18:08:12 & Shelter          & 1                              \\ \hline
3         & 2021/07/18 18:08:12 & Food          & 5                              \\ \hline
4         & 2021/07/18 19:14:29 & Food            & 3                            \\\hline
\end{tabular}}
\end{table}
\vspace{-0.45cm}
\noindent According to Example~\ref{exp:exist_demand}, we have $U_{shelter}=2$,  $R_{shelter}=0$,  $U_{food}=0$, and  $R_{food}=4$ at time $T$. 
Since there are 2 units of unfulfilled demand for shelter at time $T$ and 1 unit new request between $T$ and $T_+$,
we form $\tilde{Q}_{shelter} =  <(\tilde{t}^{shelter}_1,\tilde{q}^{shelter}_1), (\tilde{t}^{shelter}_2,\tilde{q}^{shelter}_2)>$,
where $\tilde{t}^{shelter}_1 = \text{`2021/07/18 16:38:26'}$, $\tilde{q}^{shelter}_1 =2$, 
$\tilde{t}^{shelter}_2 =\text{`2021/07/18 18:08:12'}$,  $\tilde{q}^{shelter}_2 =1$, $\tilde{n}^{shelter} = 2$, and $|\tilde{Q}_{shelter}|=3$. 
The agency uses 4 units of food remaining at time $T$ to partially fulfill the food request in Demand 3. 
We have
$\tilde{Q}_{food} =  <(\tilde{t}^{food}_1,\tilde{q}^{food}_1), (\tilde{t}^{food}_2,\tilde{q}^{food}_2)>$,
where $\tilde{t}^{food}_1 = \text{`2021/07/18 18:08:12'}$, $\tilde{q}^{food}_1 =1$, 
$\tilde{t}^{food}_2 =\text{`2021/07/18 19:14:29'}$,  $\tilde{q}^{food}_2 =3$, $\tilde{n}^{food} = 2$, and $|\tilde{Q}_{food}|=4$. 
\end{example}

We are now ready to figure out the cost reduction $g_k(x_k,\tilde{Q}_k)$ if $x_k$ units of resources, requested at time $T$, will be available at time $T_+$ 
to meet the sequence of net demands $\tilde{Q}_k$, $k = 1, 2,\dots, K$.
Denote $B^k_j=B^k(\tilde{t}^k_j)$, where $\tilde{t}_j^k$ is arrival time of the $j$-th demand in $\tilde{Q}_k$ and $B^k(\tilde{t}^k_j)$ is given by Equation \eqref{eq:reduction}, $j = 1, 2,\dots, \tilde{n}^k$. Since $\tilde{t}_1^k\le \tilde{t}_2^k\le\dots\le \tilde{t}^k_{\tilde{n}^k}$, by Equation \eqref{eq:reduction}, we have
\vspace{-0.1cm}
\begin{equation}
\label{eq:b_order}
    B_1^k\ge B_2^k\ge\dots\ge B_{\tilde{n}^k}^k.
\end{equation}
Consequently, $x_k$ units of resources should be allocated to fulfill the demands according to their arrival times in  $\tilde{Q}_k$, first come first serve.
Let $J^k$ be the {\em largest} sequence index $J$ such that 
$
\sum_{j=1}^{J} \tilde{q}^k_j \leq x_k.
$
Then, the total cost reduction due to $x_k$ units of requested resources is given by:
\begin{equation}
\label{eq:g_k_explicit}
	g_k(x_k,\tilde{Q}_k)=\begin{cases}
		\sum_{j=1}^{J^k } \tilde{q}^k_jB^k_{j} +(x_k-\sum_{j=1}^{J^k} \tilde{q}^k_j)B^k_{J^k +1} 
		& \text{if } x_k < |\tilde{Q}_k|,\\\\
		\sum_{j=1}^{ \tilde{n}^k} \tilde{q}^k_jB^k_{j} & \text{if } x_k \geq |\tilde{Q}_k|.
	\end{cases}
\end{equation}
Since cost minimization is equivalent to cost reduction maximization, 
with the above preparation, we are able to simplify the original stochastic optimization problem \eqref{pb:basic}
as follows: 
\begin{equation}
\begin{aligned}
\label{pb:new}
\max_{\M}  \quad & \G(\M) := E_{\tilde{\Q}}(\sum^K_{k=1}g_k(x_k,\tilde{Q}_k))\\
\textrm{s.t.} \quad & \sum_k w_k x_k\le W,\\
\quad & x_k\in \intg^+,\quad k=1,2,\dots, K,
\end{aligned}
\end{equation}
where the expectation is taken over $\tilde{\Q}=(\tilde{Q}_1, \tilde{Q}_2, \dots, \tilde{Q}_K)$.

\subsubsection{SAA Approximation of Stochastic Optimization.}
\label{sec:method:prrm:saa}
It is difficult to precisely evaluate the expected deprivation cost reduction 
$E_{\tilde{\Q}}(\sum_kg_k(x_k,\tilde{Q}_k))$ in problem~\eqref{pb:new}, since the underlying stochastic process $\tilde{\Q}$ is highly non-stationary. 
Instead, we follow the Sample Average Approximation (SAA) method \citep{kim_guide_2015} 
to obtain its approximation by generating sufficient number of samples of $\tilde{\Q}$. To generate a sample of $\tilde{\Q}$, we use the inference procedure of the CNM-TPP model given in Appendix~\ref{apd:add_detail:inf} to predict future demands in $(T, T_+]$ and then construct a sample of $\tilde{\Q}$ with the predicted future demands (e.g., see Example 3). Let $\tilde{\Q}^{(\psi)}=(\tilde{Q}_1^{(\psi)},\tilde{Q}_2^{(\psi)},\dots, \tilde{Q}_K^{(\psi)})$ be a sample of $\tilde{\Q}$, $\psi = 1, 2,\dots, \Psi$ and $\Psi$ is the number of samples. 
We have the following approximation for the expected deprivation cost reduction:
\begin{equation}
E_{\tilde{\Q}}(\sum_{k=1}^{K}g(x_k,\tilde{Q}_k)) \approx    \frac{1}{\Psi}\sum_{\psi=1}^{\Psi}\sum_{k=1}^{K}g_k(x_k,\tilde{Q}_k^{(\psi)}).
\end{equation}
As a result, we obtain the following SAA approximation of the stochastic optimization problem \eqref{pb:new}: 
\begin{equation}
\begin{aligned}
\label{pb:SAA}
\max_{\M} \quad &  \G^\Psi(\M):= \frac{1}{\Psi}\sum_{\psi=1}^{\Psi}\sum_{k=1}^{K}g_k(x_k,\tilde{Q}_k^{(\psi)}) \\
\textrm{s.t.} \quad & \sum_k w_kx_k\le W,\\
\quad & x_k\in \intg^+,\quad k=1,2,\dots, K.
\end{aligned}
\end{equation}
Notice that the SAA approximation converts a stochastic optimization problem to a deterministic optimization problem,
since the objective function $\G^\Psi(\M)$ is now a deterministic function of $\M$. Even though problem \eqref{pb:SAA} remains a difficult problem to solve, the conversion allows us to explore the special structure of $\G^\Psi(\M)$ and design an efficient and effective approximation algorithm, to be discussed in the next subsection.

Next, we provide a theoretical justification that the proposed SAA approximation is a valid and accurate solution approach.
More specifically, 
we show that the optimal objective function value of the deterministic optimization problem \eqref{pb:SAA} converges to that of the original stochastic optimization problem \eqref{pb:new}, as the sample size increases:
\begin{theorem}\rm
\label{thm:consis}
Let  $\pi^*$ and $\Pi^*_\Psi$ be the optimal objective function values of problems \eqref{pb:new} and \eqref{pb:SAA}, respectively. We have
$$
\lim_{\Psi \to \infty} \sup |\Pi^*_\Psi - \pi^*| = 0.
$$
\proof{$\rm Proof.$}  See Appendix~\ref{apd:thm:consis}. \endproof
\end{theorem}
By Theorem~\ref{thm:consis}, $\Pi^*_\Psi$ is a consistent estimator of $\pi^*$ as $\Psi\rightarrow \infty$, which provides a theoretical support for solving the complicated stochastic optimization problem~\eqref{pb:new} via the SAA approximation problem~\eqref{pb:SAA}. 

\subsubsection{A Greedy Algorithm for Problem \eqref{pb:SAA}.}
\label{sec:method:prrm:greedy}
The optimization problem \eqref{pb:SAA} resulted from the SAA approximation, however,
remains a difficult problem to be solved efficiently. 
As shown in the following theorem, it is a nonlinear Knapsack problem and belongs to a class of NP-hard problems: 
\begin{theorem}\rm
\label{thm:np}
The optimization problem \eqref{pb:SAA} is NP-hard.
\proof{$\rm Proof.$} See Appendix~\ref{apd:thm:np}. \endproof
\end{theorem}

Given the hardness of the problem, we design a heuristic solution procedure by careful exploring the special structure of the problem. We start by analyzing the properties of the objective function $\G^\Psi(\M)$. 
Denote
\begin{equation} \label{eq:gpsik}
\G^\Psi_k(x_k) = \frac{1}{\Psi}\sum_{\psi=1}^{\Psi}g_k(x_k,\tilde{Q}_k^{(\psi)}), \quad k = 1, 2, \dots, K.
\end{equation}
We thus have $\G^\Psi(\M) = \sum_k \G^\Psi_k(x_k)$ and problem~\eqref{pb:SAA}
can be rewritten as:
\begin{equation}
\begin{aligned}
\label{pb:SAA1}
\max_{\M} \quad &  \sum_{k=1}^K \G^\Psi_k(x_k) \\
\textrm{s.t.} \quad & \sum_k w_kx_k\le W,\\
\quad & x_k\in \intg^+,\quad k=1,2,\dots, K.
\end{aligned}
\end{equation}
It turns out that function $\G^\Psi_k(\cdot)$ has some nice structural properties, which allow us to design an efficient greedy heuristic to solve problem \eqref{pb:SAA}.
\begin{proposition}\rm
\label{prop:calg_k}
Function $\G^\Psi_k(\cdot)$  is (i) continuous piece-wise linear, (ii) monotonically non-decreasing, and (iii) concave, $k = 1, 2,\dots, K$.
\proof{$\rm Proof.$} Notice that $\G^\Psi_k(x_k)$ is a linear combination (average) of $g_k(x_k,\tilde{Q}_k)$. Since  $g_k(x_k,\tilde{Q}_k)$ has the above three properties, as shown by Proposition EC.1 in Appendix~\ref{apd:thm:consis:prop}, so does $\G^\Psi_k(x_k)$. \endproof
\end{proposition}

To design a greedy algorithm for problem \eqref{pb:SAA1}, we need to identify all non-differential points (i.e., kinks) of $\G^\Psi_k(x_k)$, $k=1, 2,\dots, K$, as well as the left and right derivatives (slopes) around each kink.
Since $\G^\Psi_k(x_k)$ is a linear combination of $g_k(x_k,\tilde{Q}_k^{(\psi)})$,
any kink of the later function is also a kink of the former function.  
Notice also that for each function $g_k(x_k,\tilde{Q}_k^{(\psi)})$, a kink occurs as a demand in $\tilde{Q}_k^{(\psi)}$ occurs.
Now consider a kink of $\G^\Psi_k(x_k)$ associated with the $i$-th demand in $\tilde{Q}_k^{(\psi)}$ that occurs at time $\tilde{t}_i^{k,(\psi)}$. 
The corresponding location of the kink in $x_k$-axis (quantity of requested resources $x_k$) is given by the following quantity: 
\begin{equation} \label{eq:kink}
s^{k,(\psi)}_i = \sum_{j=1}^{i} \tilde{q}_j^{k,(\psi)}, \quad \psi \in \{1,2,\dots, \Psi\}, 
\quad i \in \{1, 2,\dots, \tilde{n}^{k,(\psi)}\}, \quad k = 1, 2,\dots, K.
\end{equation}
Next we show how to calculate the left and right derivatives of $\G^\Psi_k(x_k)$ at $x_k=s^{k,(\psi)}_i$.
Let $J(k,\psi,\varphi,i)$ be the first demand in $\tilde{Q}_k^{(\varphi)}$ whose request for type $k$ resource cannot be fully met by given $s_{i}^{k,(\psi)}$ units of type $k$ resource, $\varphi = 1, 2,\dots, \Psi$. That is,
$$
J(k,\psi,\varphi,i) = \min j \in \{1,\dots, \tilde{n}^{k,(\varphi)}\}| s_j^{k,(\varphi)} \ge s_i^{k,(\psi)}.
$$
Then the left derivative of $\G^\Psi_k(x_k)$ at $s_i^{k,(\psi)}$ (the slope to the left of the kink) is given by:
\begin{equation} \label{eq:leftslope}
{\G^\Psi_k}^{\prime}(s_{i}^{k,(\psi)-}) = \frac{1}{\Psi}\sum_{\varphi=1}^{\Psi} \big(B^k(\tilde{t}^{k,(\varphi)}_{J(k,\psi,\varphi,i)})\times\mathds{1}(s_{i}^{k,(\psi)}\le|\tilde{Q}_k^{(\varphi)}|)\big)
\end{equation}
and the right derivative (slope to the right of the kink) is given by:
\begin{equation} \label{eq:rightslope}
{\G^\Psi_k}^{\prime}(s_{i}^{k,(\psi)+}) = {\G^\Psi_k}^{\prime}(s_{i}^{k,(\psi)-})
-\frac{1}{\Psi}( B^k(\tilde{t}^{k,(\psi)}_{i}) - B^k(\tilde{t}^{k,(\psi)}_{i+1})).
\end{equation}
Since the second term in the right derivative is negative, the right slope is smaller than the left slope, which implies that $\G^\Psi_k(x_k)$ is concave w.r.t $x_k$.

Now we are in a position to present the greedy algorithm. 
To facilitate the description, we relabel all kinks and slopes of $\G^\Psi_k(x_k)$, by sorting them according to the arrival times of their associated demands.
For simplicity, we assume $\Psi$ to be a fixed constant in the remaining exposition.
Let $s^k_i$, $i = 1, 2,\dots, M^k$ be all kinks of function $\G^\Psi_k(x_k)$, as calculated by Equation \eqref{eq:kink}, such that $0 \leq s^k_1 \leq, \dots, \leq s^k_{M^k}$, where $M^k=\sum_{\psi=1}^\Psi\tilde{n}^{k,(\psi)}$.
Let $\bar{B}^k_i$ be the $i$-th slope of function $\G^\Psi_k(x_k)$ for all $x_k \in (s^k_{i-1}, s^k_i)$. Suppose $s^k_i$ is the kink associated with $\tilde{i}$-th demand in $\tilde{Q}_k^{(\psi)}$, i.e., $s^k_i = s^{k,(\psi)}_{\tilde{i}}$. Accordingly, $\bar{B}^k_i$ can be calculated by
\begin{equation}
\label{eq:slope}
    \bar{B}^k_i= {\G^\Psi_k}^{\prime}(s_{i}^{k-})= {\G^\Psi_k}^{\prime}(s_{\tilde{i}}^{k,(\psi)-})= \frac{1}{\Psi}\sum_{\varphi=1}^{\Psi} \big(B^k(\tilde{t}^{k,(\varphi)}_{J(k,\psi,\varphi,\tilde{i})})\times\mathds{1}(s_{\tilde{i}}^{k,(\psi)}\le|\tilde{Q}_k^{(\varphi)}|)\big).
\end{equation}
In view of Proposition~\ref{prop:calg_k} and the above discussion, we have 
$\bar{B}^k_1  \geq \bar{B}^k_2\geq,\dots, \geq \bar{B}^k_{M^k} \geq 0.$
We start by ranking all slopes in descending order according to the ratio 
$\bar{B}^k_i/w_k$ among all resource types $k = 1, 2,\dots, K$ and slopes $i=1, 2,\dots,  M^k$.
Suppose the resulting sequence is $\PP$.
Our proposed greedy algorithm decides quantities of requested resources by following sequence $\PP$ since the ratio reflects the current effectiveness of type $k$ resources, in terms of reducing the deprivation cost, relative to 
its capacity consumption. 
Let $\kappa(\rho)$ be the type of requested resource associated with the $\rho$-th slope in sequence $\PP$ and $I_k(\rho)$ be the total number of slopes associated with type $k$ resource request among top $\rho$ ranked slopes in sequence $\PP$. The greedy algorithm then follows sequence $\PP$ to iteratively decide the requested quantity of type $k$ resources as follows:
\begin{equation}
\label{eq:xkj}
x_k(\rho) = 
\begin{cases}
		 s^k_{I_k(\rho)} & \text{if } I_k(\rho) \neq 0,\\
		0 & \text{otherwise,}
\end{cases}
\quad k =1,2,\dots, K.
\end{equation}
The greedy algorithm stops when the shipping capacity constraint is violated for the first time.
The corresponding sequence index is given by:
\begin{equation} \label{eq:jw+}
\rho_W = \min \rho \in \{1,2,\dots,|\PP|\} | \sum_{k=1}^K w_k x_k(\rho) \geq W.
\end{equation}
If the inequality in \eqref{eq:jw+} holds as equality, then we obtain an optimal solution to problem~\eqref{pb:SAA1}: ${\M}^* = (x_1^*, \dots, x_K^*)$, where $x_k^* = x_k(\rho_W)$, $k=1,2, \dots, K$.
Otherwise, we keep $x^*_k$ for all resource types except type $\kappa(\rho_W)$, which is currently a less effective resource type relative to others ranked higher in sequence $\PP$.  
Denote this critical type $\kappa(\rho_W)$ as $k^{\circ}$. We reduce the requested quantity for resource type $k^{\circ}$ until the total capacity is satisfied,
which results in the following maximum fractional requested quantity for resource type $k^{\circ}$:
\begin{equation}
\label{eq:xfrac}
    \tilde{x}_{k^{\circ}} = \frac{1}{w_{k^{\circ}}} (W-\sum_{k \neq k^{\circ}} w_k x_k(\rho_W)).
\end{equation}
Let $x_{k^{\circ}} = \lfloor \tilde{x}_{k^{\circ}} \rfloor$ be the integer portion of  $\tilde{x}_{k^{\circ}}$. 
Then our greedy algorithm generates the following feasible solution to problem~\eqref{pb:SAA1}: 
\begin{equation}
\label{eq:xkg}
{\M}^g = (x_1^g, \dots, x_K^g), \text{ where }
x_k^g = 
\begin{cases}
		x_k(\rho_W)  & \text{if } k \neq k^{\circ},\\
		x_{k^{\circ}}  & \text{if } k = k^{\circ},
		\end{cases}
\quad k =1,2,\dots, K.
\end{equation}

To evaluate the performance of the greedy algorithm, we compare the cost reduction $\Pi^g_\Psi$ by the greedy algorithm
v.s. the optimal cost reduction $\Pi^{*}_\Psi$ of problem~\eqref{pb:SAA1}, where $\Pi^g_\Psi = \G^\Psi({\M}^g)$ and ${\M}^g$ is the solution to the problem by the greedy algorithm.
We have the following performance guarantee for the greedy algorithm. 
\begin{theorem}\rm
\label{thm:gap}
$\Pi^g_\Psi \geq \Pi^{*}_\Psi - \bar{B}^{k^\circ}_{I^\circ}$, where $k^\circ = k(\rho_W)$ and $I^\circ = I_{k^\circ}(\rho_W)$
\proof{$\rm Proof.$}
Consider the following continuous solution to  problem~\eqref{pb:SAA1}, which differs from  $\M^g$ on 
only the requested quantity for resource type $k^{\circ}$:
\begin{equation*}
\label{eq:xkc}
\M^c = (x_1^c, \dots, x_K^c), \text{ where }
x_k^c = 
\begin{cases}
		x_k(\rho_W)  & \text{if } k \neq k^{\circ},\\
		\tilde{x}_{k^{\circ}}  & \text{if } k = k^{\circ},
		\end{cases}
\quad k =1,2,\dots, K.
\end{equation*}
By Proposition~\ref{prop:calg_k}, the objective function is continuous and concave. 
Consequently, $\M^c$ is an optimal solution for problem~\eqref{pb:SAA1} without the integer constraint 
and $\G^\Psi(\M^c) \geq \Pi^{*}_\Psi$.
We have 
$$
\begin{aligned}
\Pi^{*}_\Psi - \Pi^g_\Psi & \leq   \G^\Psi(\M^c) - \Pi^g_\Psi \\
& = (\tilde{x}_{k^{\circ}} - \lfloor \tilde{x}_{k^{\circ}} \rfloor)\bar{B}^{k^\circ}_{I^\circ}\\
& \leq \bar{B}^{k^\circ}_{I^\circ}
\end{aligned}
$$
and the theorem follows immediately. \Halmos
\endproof
\end{theorem}

\begin{algorithm}[H]
	\caption{A Greedy Algorithm to Solve Problem~\eqref{pb:SAA}}
  \label{alg:greedy}
    \textbf{Input:} $W$, $\{w_k|k=1,2,\dots,K\}$, $\{\tilde{\Q}^{(\psi)}|\psi=1,2,\dots,\Psi\}$ \\
    \textbf{Output:} ${\M}^g = (x_1^g, \dots, x_K^g)$
	\begin{algorithmic}[1]
    \State Calculate slopes $\bar{B}^k_{i}$ via Equation~\eqref{eq:slope}, $i = 1,2,\dots,M^k$, $k=1,2,\dots,K$
    \State Obtain slope sequence $\PP$ by ranking all slopes in descending order according to the ratio $\bar{B}^k_{i}/w_k$\label{alg:greedy:ln:ranking}
    \State $\rho=1$ 
    \While {$\sum_{k=1}^K w_k x_k(\rho) < W$}
        \State $\rho=\rho+1$
        \State Calculate $x_k(\rho)$ via Equation~\eqref{eq:xkj}, $k=1,2,\dots,K$
    \EndWhile
    \State  $\rho_W=\rho$
        \If{$\sum_{k=1}^K w_k x_k(\rho_W) = W$}
            \State $x_k^g=x_k(\rho_W)$, $k=1,2,\dots,K$
        \Else
            \State $k^\circ=\kappa(\rho_{W})$
            \State Obtain maximum fractional requested quantity for resource type $k^\circ$, $\tilde{x}_{k^\circ}$, via Equation~\eqref{eq:xfrac}
            \State $x_k^g = 
\begin{cases}
		\lfloor \tilde{x}_{k^\circ} \rfloor & \text{if } k = k^\circ,
		\\
		x_k(\rho_W)  & \text{if } k \neq k^\circ,\\
\end{cases}
\quad k =1,2,\dots, K$
        \EndIf
    \State \Return ${\M}^g = (x_1^g, \dots, x_K^g)$
	\end{algorithmic} 
\end{algorithm}

Note that problem \eqref{pb:SAA1} is equivalent to problem \eqref{pb:SAA}.
The above theorem guarantees that the maximum loss from using the efficient greedy algorithm we propose, rather than an optimal but time consuming algorithm, to solve the NP-hard problem \eqref{pb:SAA}, is no more than the deprivation cost of missing a demand for one unit of certain type resources. We formally present the proposed greedy algorithm in Algorithm~\ref{alg:greedy}. The computational time of Algorithm~\ref{alg:greedy} is dominated by the sorting operation in line~\ref{alg:greedy:ln:ranking} of the algorithm, which is known to run efficiently in log-linear time.

\section{Empirical Evaluation}\label{sec:eval}

\subsection{Data and Evaluation Procedure}\label{sec:eval:data}

\begin{figure}[H]
	\FIGURE
	{\includegraphics[scale=0.45]{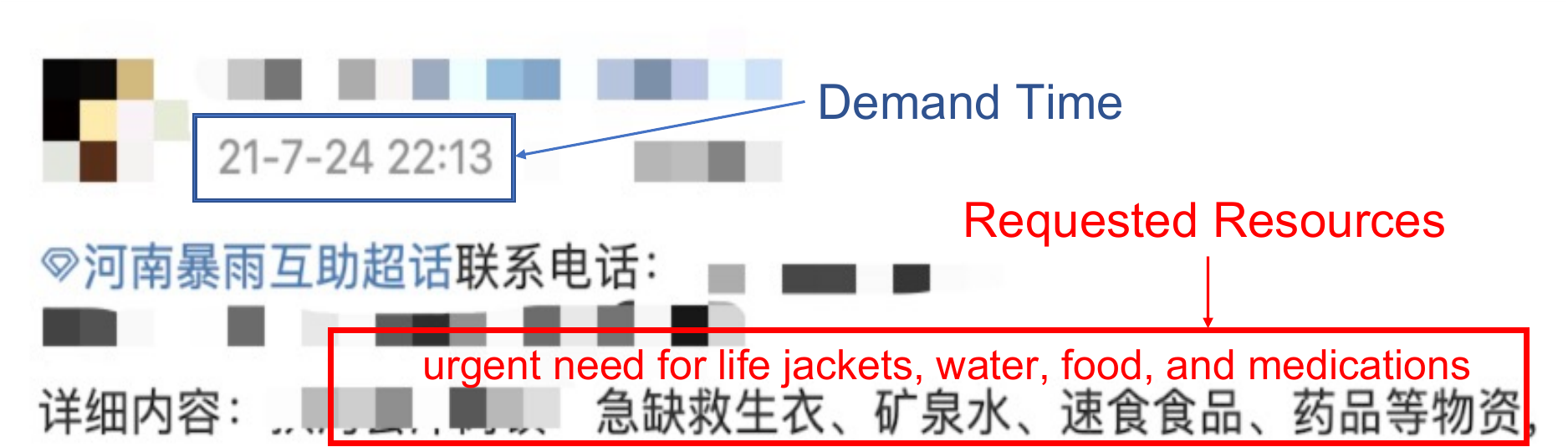}}
	{An Emergency Demand Post on Weibo in Response to the 2021 Henan Floods. \label{fg:weibo}}
	{}
\end{figure}

In July 2021, severe floods struck China's Henan province, causing 398 deaths and \$12.7 billion in property damages.\footnote{See \href{https://en.wikipedia.org/wiki/2021_Henan_floods}{https://en.wikipedia.org/wiki/2021\_Henan\_floods} (last accessed on March 23, 2023)} During the response phase of this disaster, Weibo, the largest social media platform in China, became an important tool for disaster-affected people to request disaster relief resources.\footnote{See \href{https://www.whatsonweibo.com/how-social-media-is-speeding-up-zhengzhou-flooding-rescue-efforts/}{https://www.whatsonweibo.com/how-social-media-is-speeding-up-zhengzhou-flooding-rescue-efforts/}  (last accessed on March 23, 2023)} 
Figure~\ref{fg:weibo} shows a Weibo post in response to the disaster, with demand time and requested resources highlighted. Therefore, we evaluate the performance of our and benchmark methods with data collected from Weibo posts concerning emergency demands in response to the 2021 Henan floods.  

We employed a public dataset extracted from 3496 Weibo posts concerning the 2021 Henan floods during the period of July 21, 2021 to July 27, 2021.\footnote{The dataset can be accessed at \href{https://github.com/GiveHenanAHand/henan-rescue-viz-website}{https://github.com/GiveHenanAHand/henan-rescue-viz-website}.} Among these Weibo posts, we manually identified 860 demands (posts) that requested disaster relief resources. In real world practices, 
disaster relief resources serving a common objective are packaged as a kit to enable rapid response and fast delivery \citep{vanajakumari_integrated_2016}. 
For example, resources with a common objective of saving lives, such as food, shelter, and medication, can be grouped into a kit. Therefore, we categorized resources requested in the demands into three kits: onsite support (including resources such as flashlight and inflatable boat), lifesaving (including resources such as medication and food), as well as damage repair (including resources such as flood barrier and pump), according to the \textit{Catalog of Emergency Resources} issued by the National Development and Reform Commission of China.\footnote{The \textit{Catalog of Emergency Resources} broadly categorizes disaster-relief resources into three groups: onsite support, lifesaving, and damage repair, with a list of resources in each group. This catalog can be accessed at  \href{https://www.ndrc.gov.cn/fzggw/jgsj/yxj/sjdt/201504/W020190906509018532631.pdf}{https://www.ndrc.gov.cn/fzggw/jgsj/yxj/sjdt/201504/W020190906509018532631.pdf.}}
As a result, there are three types of resources in our evaluation, each of which corresponds to a kit. Among the 860 demands, type 1 resources (i.e., onsite support kit) are requested in 328 demands, type 2 resources (i.e., lifesaving kit) are requested in 666 demands, and type 3 resources (i.e., damage repair kit) are requested in 194 demands. The Weibo dataset does not include quantities of requested resources. To overcome this limitation, we set quantities to 1 unit for requested resources in the dataset. 

\begin{figure}[H]
	\FIGURE
	{\includegraphics[scale=0.25]{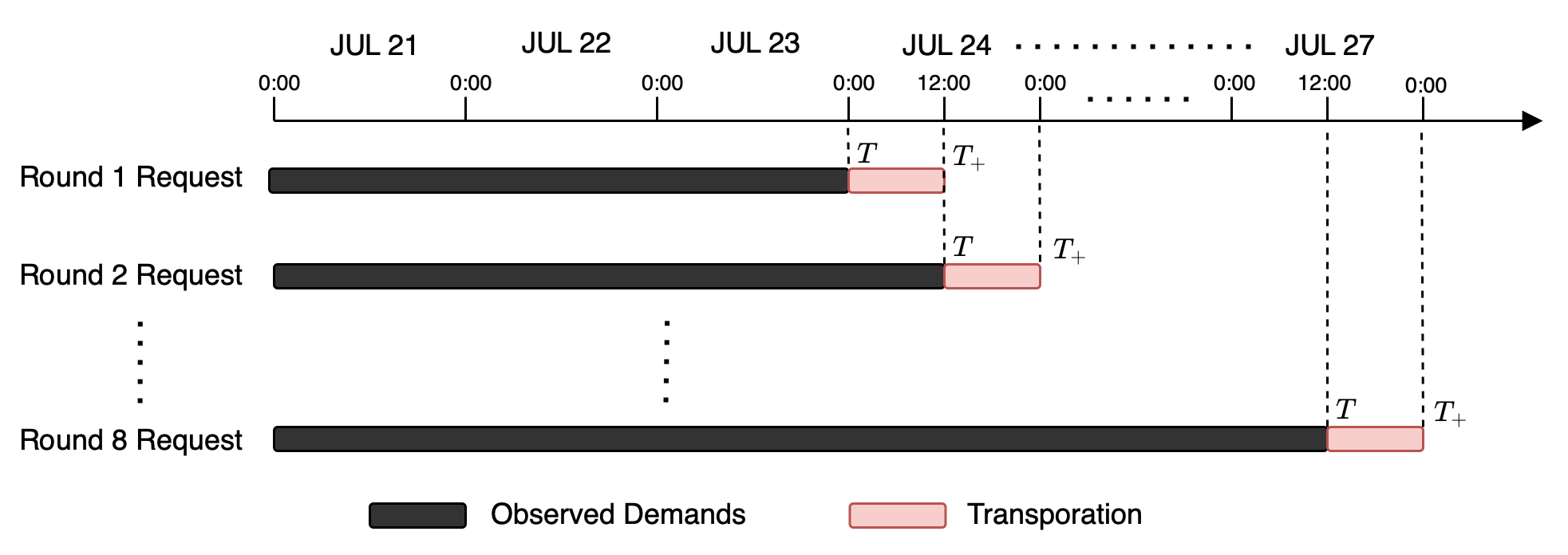}}
	{Timeline of Resource Requests in the Evaluation. \label{fg:timeline}}
	{}
\end{figure}

Next, we detail the evaluation procedure. As shown in Figure~\ref{fg:timeline}, we used the demands occurred in July 21, 22, and 23 to train an investigated method. Resources were then requested at hour 00:00 of July 24 (i.e., $T$), with their quantities decided by the trained method. We set the transportation time to 12 hours. Thus, requested resources arrived at hour 12:00 of July 24 (i.e., $T_+$). Once arrived, these resources were distributed to meet demands and another round of resource request started. Hence, we reset resource request time $T$ to hour 12:00 of July 24, trained the method using the demands occurred in the period from hour 00:00 of July 21 to hour 12:00 of July 24, and requested resources according to their quantities decided by the trained method. We repeated the above-mentioned process and the last resource request time was hour 12:00 of July 27. There could be demands not satisfied until the end of July 27. These demands were fulfilled by a transportation started at the end of July 27 and arrived 12 hours later (i.e., hour 12:00 of July 28).

\vspace{-0.3cm}
\vspace{-0.3cm}

The performance of a method was evaluated using the average unit demand deprivation cost incurred by the method, where the average was taken over the demands occurred in the period from July 24 to July 27. Consider a unit demand for type $k$ resources (i.e., a demand for one unit of type $k$ resources), $k=1,2,3$. We can observe its demand time and derive its fulfillment time by running the method according to the evaluation procedure elaborated above. Thus, we can calculate the delay $\delta$ of fulfilling this unit demand as the difference between its fulfillment time and demand time. By Equation~\eqref{eq:depri}, the deprivation cost of fulfilling this unit demand is 
\begin{equation*}
e^{\phi+bc_k\delta}-e^{\phi},
\end{equation*}
where $c_k$ is the importance score of type $k$ resources. Let $N_k$ denote the total units of type $k$ resources requested in the demands occurred in the period from July 24 to July 27, $k=1,2,3$. 
The average unit demand deprivation cost is given by
\begin{equation}
\label{eq:ave_unit_depri}
\frac{1}{N}\sum_{k=1}^3\sum_{i_k=1}^{N_k}(e^{\phi+bc_k\delta_{i_k}}-e^{\phi}),
\end{equation}
where $\delta_{i_k}$ denotes the delay of meeting the $i_k$-th unit demand for type $k$ resources and $N=\sum_{k=1}^3 N_k$ is the total units of requested resources. \citet{holguin-veras_appropriate_2013} estimate the deprivation parameters $\phi=1.5031$ and $b=0.1172$ based on disaster-affected victims' willingness to pay for a deprived resource and  calculate deprivation costs in dollar amount with these parameter values. Subsequent disaster management studies have commonly used these parameter values for the calculation of deprivation costs (e.g., \citealp{rivera-royero_dynamic_2016}).
Therefore, we set $\phi=1.5031$ and $b=0.1172$ in Equation \eqref{eq:ave_unit_depri} by following \citet{holguin-veras_appropriate_2013}.

\subsection{Benchmark Methods}\label{sec:eval:bench}
Our proposed method consists of two components: the CNM-TPP model described in Section \ref{sec:method:ctpp} for future demand prediction and the proactive resource request (PRR) method proposed in Section \ref{sec:method:prrm} that takes future demands predicted by CNM-TPP as an input and decides optimal quantities of requested resources. Therefore, we named our method CNM-PRR. We benchmarked our method against the current practice of resource request, which reactively sets quantities of requested resources as quantities of currently unfulfilled demands \citep{huang_modeling_2015}.
We called this benchmark the reactive resource request (ReR) method. The comparison between our method and ReR not only reveals the practical value of our method but also demonstrates the benefit of proactive resource request, the main novelty of our study.

In addition, as reviewed in Section \ref{sec:rw:tpp}, temporal point process (TPP) is the dominant technique for predicting future events. Hence, we can adapt existing TPP models to predict future demands. In particular, we chose LogNormMix, a state-of-the-art TPP model proposed by \citet{shchur_intensity-free_2020}, which has shown superior performance over other existing TPP models in various event prediction tasks. We also included three recently proposed TPP models as benchmarks: A-NDTT, AttnMC, and CTDRP. Specifically,  A-NDTT by \citet{mei_transformer_2022} is a state-of-the-art TPP model based on the transformer architecture,  AttnMC is a deep learning-based TPP model proposed by \citet{enguehard_neural_2020}, and CTDRP by \citet{turkmen_intermittent_2019} is a novel deep learning-based TPP model designed for supply chain demand forecasting. 
To adapt these TPP methods for future demand prediction, we replaced their mark embedding functions with ours defined in Equation \eqref{eq:me_detail} because their mark embedding function cannot capture both types and quantities of requested resources in a demand.
To solve the disaster response problem, a demand forecasting method needs to be combined with a resource request method, which employs future demands predicted by the method to decide quantities of requested resources. To this end, we combine each of them with our proactive resource request (PRR) method and name the resulted method as LogNormMix-PRR (LogNormMix+PRR), A-NDTT-PRR (A-NDTT+PRR), AttnMC-PRR (AttnMC+PRR), and CTDRP-PRR (CTDRP+PRR).

Moreover, we benchmarked our method against state-of-the-art data-driven inventory control (IC) models because our method is a data-driven method and data-driven IC models represent recent developments in the IC literature. Among IC problems, the newsvendor problem, which determines the optimal stock level based on future demand quantities, is most similar to the PRR problem. We thus applied state-of-the-art data-driven methods for the newsvendor problem, LR-NV (Linear Regression-based data-driven method for the NewsVendor problem, \citealp{ban_big_2019}) and DL-NV (Deep Learning-based data-driven method for the NewsVendor problem, \citealp{oroojlooyjadid_applying_2020}), to solve the PRR problem and compared the performance of our method against these methods. 

Also, we integrated LogNormMix with a simple importance adjusted first-come-first-serve (IFCFS) resource request method. Specifically, the IFCFS method first prioritized and grouped resources in the demands by their importance scores; within each importance group, resources were requested in a first-come-first-serve manner according to their demand times until the transportation capacity was reached. We named this benchmark the LogNormMix-IFCFS method.
Table \ref{tab:Bench} summarizes the methods compared in the evaluation.   

\vspace{-0.4cm}
\begin{table}[H]
 	\TABLE
	{ Summary of Methods Compared in the Evaluation \label{tab:Bench}}
	{
	 		\begin{tabular}{L{95pt} L{310pt}}
 			\hline
 			\textbf{Method} & \textbf{Notes}      \\ \hline
 			CNM-PRR                 & Our Proposed method   \\
 			  ReR & Reactive resource request method  \\ 
 			  LogNormMix-PRR    & The combination of LogNormMix, a state-of-the-art TPP model by \citet{shchur_intensity-free_2020}, and our proactive resource request (PRR) method  \\
  			   A-NDTT-PRR & The combination of A-NDTT \citep{mei_transformer_2022}, a recent TPP model based on the transformer architecture, and our PRR method \\
  			  AttnMC-PRR & The combination of AttnMC, a deep learning-based TPP model by \citet{enguehard_neural_2020}, and our PRR method \\
  			  CTDRP-PRR & The combination of CTDRP, a novel deep learning-based TPP model by \citet{turkmen_intermittent_2019} for supply chain demand forecasting, and our PRR method \\
  			 LR-NV & a data-driven newsvendor method by \citet{ban_big_2019}, predicting future demand quantities using linear regression\\
  			 DL-NV & a data-driven newsvendor method by \citet{oroojlooyjadid_applying_2020}, predicting future demand quantities using a deep learning method \\
  			  LogNormMix-IFCFS   & The combination of LogNormMix and the importance adjusted first-come-first-serve (IFCFS) resource request method  \\
             \hline
 		\end{tabular}
 		}
 		{}
 \end{table}
\vspace{-0.4cm}

Next, we discuss the implementation details of these methods. We implemented our CNM-TPP model with PyTorch, and then trained it with the Adam optimizor
using learning rate $0.001$ for $30$ epochs. To tune the hyperparameters of CNM-TPP and other benchmarks, we reserved a portion of the training demands occurred in July 21, 22, 23 as a validation dataset. 
The hyperparameters of CNM-TPP includes $n_e$ (the embedding size of RNN in Equation \eqref{eq:hist_emb}), $n_z$ (the number of log-normal mixture components in Equation \eqref{eq:cpdf_tau}), and $\gamma$ (the multiplicator in Equation \eqref{eq:obj_full}).
We set $n_e=n_z=64$, and $\gamma=1$. A detailed walk-through of the computation flow of CNM-TPP under this parameter setting can be found in Appendix \ref{apd:para_spc}.
LogNormMix features two key hyperparameters: the embedding size of historical events and the number of log-normal mixture components. We set both as $64$. 
Both A-NDTT and AttnMC are constructed with transformer blocks which can be specified by two hyperparameters: embedding size and number of layers. For both models, we set the former as $64$ and the latter as $1$. 
CTDRP has one hyperparameter, namely, the size of the hidden state of a RNN layer, which is set as $64$. 
We note that our parameter setting is comparable to those reported in \cite{enguehard_neural_2020} and \citet{mei_transformer_2022}.
We defined the unit back-ordering cost in LR-NV and DL-NV as the cost incurred due to the lack of one unit of resources. In addition, to extend them to multiple types of resources, we added a constraint that the total capacity consumed by shipping the requested units of all resource types must not exceed the transportation capacity. 
DL-NV employed a multilayer perceptron with 3 hidden layers of sizes 128, 64, and 16, respectively.

\subsection{Evaluation Results and Analysis}\label{sec:eval:results}

Following the evaluation procedure, we conducted experiments to evaluate the performance of each compared method. Recall that there are three types of resources in the evaluation: type 1 resources (i.e., onsite support kit), type 2 resources (i.e., lifesaving kit), and type 3 resources (i.e., damage repair kit). Since each type is a bundle of various-sized resources, we assume that the three types of resources consume the same transportation capacity. Accordingly, we set the transportation capacity of shipping one unit of any type of resources to 1, i.e., $w_1=w_2=w_3=1$. In general, it is more urgent and important to satisfy demands for lifesaving resources than the other two types. Therefore, the importance score of type 2 resources is higher than that of the other two types. We thus set the importance scores of types 1, 2, and 3 resources to 2, 4, and 2, respectively, i.e., $c_1=2$, $c_2=4$, and $c_3=2$. We set the transportation capacity $W$ to $200$ and then varied its value to examine the performance of the investigated methods under different transportation capacities.

Table~\ref{tab:henan_res} reports the average unit demand deprivation cost of each method. Recall that deprivation cost measures the economic value of human suffering because of the deprivation of vital resources \citep{holguin-veras_appropriate_2013}. If a local agency employs our method to decide types and quantities of requested resources, the average economic value of human suffering due to the delay of one unit of resources is \$25.69, which is substantially lower than that of the reactive resource request (ReR) method. The enormous cost reduction by our method over ReR demonstrates the benefit of proactive resource request and reveals the practical value of our method in alleviating human suffering caused by a disaster.\footnote{The other benchmarks are 
also proactive resource request methods and the significant cost reductions by these methods over ReR further confirm the benefit of proactive resource request.} Moreover, in comparison to A-NDTT-PRR, the best performing benchmark, our method reduces cost by 15.15\%. Since the only difference between A-NDTT-PRR and our method is their respective TPP models, the cost reduction is attributed to the superior performance of our proposed TPP model over A-NDTT, a state-of-the-art existing TPP model, in predicting future demands. Our method also outperforms LogNormMix-IFCFS by 88.63\% in cost reduction, which is due to the superiority of our TPP model over LogNormMix and the performance advantage of our proactive resource request (PRR) method over the importance adjusted first-come-first-serve (IFCFS) method.

\begin{table}[H]
	\TABLE
	{ Performance Comparison on Average Unit Demand Deprivation Cost
	\\($W=200$,  $w_1=1$, $w_2=1$, $w_3=1$, $c_1=2$, $c_2=4$, $c_3=2$ ) \label{tab:henan_res}}
	{\begin{tabular}{C{110pt} C{120pt} C{120pt}}
\hline
\textbf{Method}                                             & \begin{tabular}[c]{@{}c@{}}Average Unit Demand\\ Deprivation Cost    \end{tabular}  & \begin{tabular}[c]{@{}c@{}}Cost Reduction \\by CNM-PRR \end{tabular} \\ \hline
\begin{tabular}[c]{@{}c@{}}CNM-PRR\\ (Our Method)  \end{tabular} & \$25.69 & \\
  ReR  & \$21905.50 & 99.88\% \\
  LogNormMix-PRR  & \$32.32 & 20.51\% \\
  A-NDTT-PRR & \$30.28 & 15.15\% \\  
  AttnMC-PRR   & \$31.26 & 17.82\% \\ 
  CTDRP-PRR   & \$37.35 & 31.22\% \\ 
  LR-NV  & \$70.14 & 63.37\% \\ 
  DL-NV   & \$58.76 & 56.28\% \\ 
LogNormMix-IFCFS  & \$225.95 & 88.63\% \\
\hline
\end{tabular}}
	{}
\end{table}
\vspace{-0.6cm}

\begin{figure}[H]
	\FIGURE
	{\includegraphics[scale=0.25]{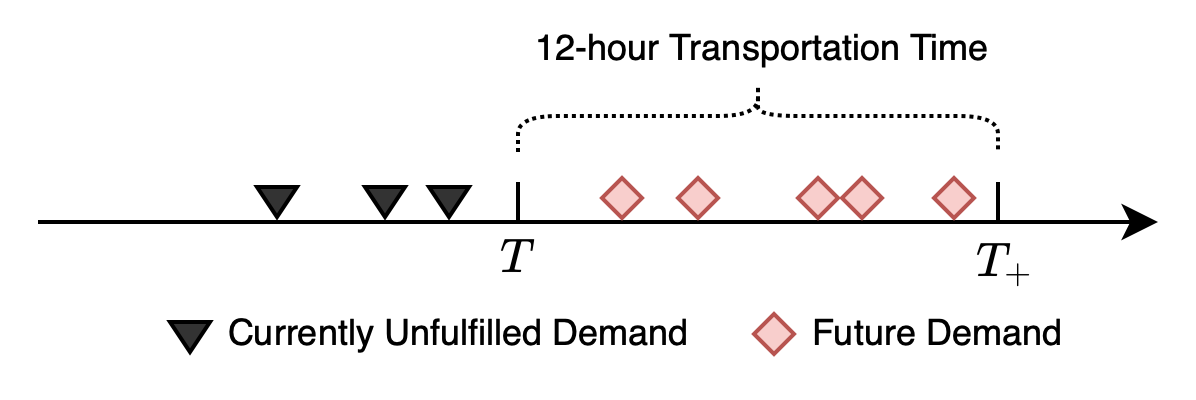}}
	{Demands Could be Considered at Resource Request Time $T$. \label{fg:demand}}
	{}
\end{figure}
\vspace{-0.6cm}

To dig deeper into the superior performance of our method, let us examine two kinds of demands that could be considered at resource request time $T$. As shown in Figure \ref{fg:demand}, currently unfulfilled demands occurred before $T$ and future demands will arrive between $T$ and $T_+$ (i.e., resource arrival time). Because of the 12-hour transportation time, the time delay of satisfying a currently unfulfilled demand is inevitably greater than 12 hours. On the other hand, if a future demand is proactively requested at $T$, the time delay of meeting this demand is less than 12 hours. Benchmark method ReR neglects future demands and requests resources based on currently unfulfilled demands only. Consequently, the time delay of satisfying a demand by this method is long, which in turn leads to high deprivation cost. On the other hand, an effective method that can accurately predict future demands and proactively request resources to meet these predicted demands can significantly reduce the time delay of satisfying a demand to less than 12 hours, thereby incurring much lower deprivation cost. 
Table \ref{tab:henan_time} reports the average time delay of satisfying one unit of resource demand by each method. As reported, the average 
time delay of our method is 7.42 hours, which is substantially shorter than 12 hours and outperforms that of each benchmark method by a range between $9.74\%$ and $62.11\%$. The reactive ReR method, on the other hand, incurs the longest time delay, which is substantially longer than 12 hours.

We further investigate the average percentage of future demands occurring between time $T$ and $T_+$ (see Figure \ref{fg:demand}) that are proactively requested and fulfilled.    
Table \ref{tab:henan_ratio} lists the average percentage for each method. As expected, the average percentage for the ReR method is zero because it neglects future demands. Our method attains the highest average percentage, outperforming that of A-NDTT-PRR (the best performing benchmark) by $8.55\%$.

\vspace{-0.3cm}
\begin{table}[H]
	\TABLE
	{ Performance Comparison on Average Unit Demand Time Delay\\($W=200$,  $w_1=1$, $w_2=1$, $w_3=1$, $c_1=2$, $c_2=4$, $c_3=2$ ) \label{tab:henan_time}}
	{\begin{tabular}{C{110pt} C{120pt} C{120pt}}
\hline
\textbf{Method}                                             & \begin{tabular}[c]{@{}c@{}}Average Unit Demand\\ Time Delay(in Hour)   \end{tabular} & \begin{tabular}[c]{@{}c@{}}Time Delay Reduction \\by CNM-PRR \end{tabular} \\ \hline
\begin{tabular}[c]{@{}c@{}}
CNM-PRR\\ (Our Method)  \end{tabular} 
& 7.42 & \\
  ReR & 19.58 & 62.11\% \\
  LogNormMix-PRR & 8.41  & 11.79\%\\
    A-NDTT-PRR     & 8.22  & 9.74\% \\  
  AttnMC-PRR    & 8.37  & 11.32\% \\  
  CTDRP-PRR   & 8.81  & 15.83\% \\  
  LR-NV    & 10.02 & 26.00\% \\  
  DL-NV  & 9.56 & 22.36\% \\  
LogNormMix- IFCFS  & 13.87 & 46.51\% \\
\hline
\end{tabular}}
	{}
\end{table}

\vspace{-0.5cm}
\begin{table}[H]
	\TABLE
	{ Performance Comparison on Average Percentage of Fulfilled Future Demands \\($W=200$,  $w_1=1$, $w_2=1$, $w_3=1$, $c_1=2$, $c_2=4$, $c_3=2$ ) \label{tab:henan_ratio}}
	{\begin{tabular}{C{110pt} C{120pt} C{120pt}}
\hline
\textbf{Method}                                             & \begin{tabular}[c]{@{}c@{}}Average Percentage of Fulfilled\\ Future Demands \end{tabular} & \begin{tabular}[c]{@{}c@{}}Improvement \\by CNM-PRR \end{tabular} \\ \hline
\begin{tabular}[c]{@{}c@{}}CNM-PRR\\ (Our Method)  \end{tabular} & 0.80 & \\
  ReR                                                        & 0.00 & - \\
  LogNormMix-PRR   & 0.72 & 11.10\%\\
      A-NDTT-PRR      & 0.74 & 8.55\% \\  
  AttnMC-PRR    & 0.73 & 10.06\% \\  
  CTDRP-PRR   & 0.68 & 17.00\% \\  
  LR-NV    & 0.58 & 37.27\% \\  
  DL-NV  & 0.62 & 28.64\% \\  
 LogNormMix- IFCFS  & 0.42 & 90.46\% \\
\hline
\end{tabular}}
	{}
\end{table}
\vspace{-0.3cm}

The evaluation demonstrates the superior performance of our proposed CNM-PRR method over each benchmark. All the performance improvements by our method reported in Tables \ref{tab:henan_res} to \ref{tab:henan_ratio} are statistically significant ($p<0.01$).
The superiority of our method is attributed to its CNM-TPP model that can accurately predict future demands and its proactive resource request (PRR) method that can proactively decide optimal quantities of requested resources. As a result, our method achieves the best performance in future demand fulfillment (as reported in Table \ref{tab:henan_ratio}) and unit demand time delay (as reported in Table \ref{tab:henan_time}) among all the compared methods, thereby attaining the lowest deprivation cost.  

We conducted an ablation analysis to demonstrate the contribution of each novelty of our CNM-TPP to the performance of our method. Specifically, CNM-TPP features three methodological novelties: a new mark embedding
function, a novel CSD learning objective, and the modeling of the
correlations among different types of resources requested in a demand. Since the proposed mark
embedding function is indispensable to solve the PRR problem, our ablation
analysis focuses on the contributions of the other two novelties.
To inspect the contribution of the correlation modeling, we simplified the event generation layer
by assuming conditional independence among different types of resource requested in a demand.
We called the simplified TPP with the conditional independence (CI) assumption as CI-TPP and its integration with the PRR method as CI-PRR. 
To evaluate the contribution of the CSD learning objective, we further dropped it from CI-TPP; and the resulted TPP was only trained with the NLL objective, like many existing TPPs \citep{shchur_neural_2021}. We named the resulted TPP as NLL-CI-TPP, and its integration with the PRR method as NLL-CI-PRR. 
Table \ref{tab:ablation_res_main} compares the performance between our method and CI-PRR, NLL-CI-PRR.

Since CI-PRR is derived by ignoring the correlations among different types of resources requested in a demand, the superiority of CNM-PRR over CI-PRR reveals the contribution of the correlation modeling to the performance of our method. 
Similarly, the outperformance of CI-PRR over NLL-CI-PRR uncovers the contribution of the CSD learning objective. 
As reported in Table \ref{tab:ablation_res_main}, both novelties contribute significantly to the performance of our method. 
Specifically, by modeling the correlations among different types of resources requested in a demand, CNM-PRR outperforms CI-PRR by $9.35\%$; by introducing the CSD learning objective, CI-PRR surpasses NLL-CI-PRR  by $12.31\%$. 

\vspace{-0.3cm}
\begin{table}[h]
\TABLE{Ablation Analysis
\\($W=200$,  $w_1=1$, $w_2=1$, $w_3=1$, $c_1=2$, $c_2=4$, $c_3=2$ )
\label{tab:ablation_res_main}}
{
\begin{tabular}{C{100pt} C{100pt} C{100pt} C{100pt}}
\hline
\textbf{Method} & 
\begin{tabular}[c]{@{}c@{}}Average Unit Demand\\ Deprivation Cost
\end{tabular} & 
\begin{tabular}[c]{@{}c@{}}Cost Reduction \\by CNM-PRR 
\end{tabular} &
\begin{tabular}[c]{@{}c@{}}Cost Reduction \\by CI-PRR 
\end{tabular} 
\\ 
\hline
\begin{tabular}
[c]{@{}c@{}}CNM-PRR\\ (Our Method)  
\end{tabular} 
    & \$25.69 & & \\
CI-PRR & \$28.34 & 9.35\% & \\
NLL-CI-PRR  & \$32.32 & 20.51\% & 12.31\% \\
\hline
\end{tabular}}
{}
\end{table}
\vspace{-0.3cm}

We further demonstrated the superior performance of our method in different contexts. Specifically, in Appendix~\ref{apd:rob}, we evaluated the robustness of our method’s superiority over the benchmarks by varying transportation capacity $W$ and importance scores $c_k$. In Appendix \ref{apd:model_ext}, we extended our method to the scenario, where resource arrival time is stochastic and over-requested resources incur holding costs.  In Appendix~\ref{apd:syn}, we employed simulation to demonstrate the generalizability of our method.

\subsection{Evaluation under a Multi-stakeholder and Multi-objective Setting}\label{sec:eval:multistakeholder}
We conducted simulations to evaluate our and benchmark methods under a multi-stakeholder and multi-objective setting as depicted in Figure~\ref{fg:system}. As shown in the figure, all stakeholders share the common objective of minimizing the cost of delays in demand satisfactions, which can be operationalized using the average unit demand deprivation cost (i.e., Equation \eqref{eq:ave_unit_depri}).
Moreover, each type of stakeholder could have its own specific objective \citep{abbasi_don_2019,abbasi_call_2021}. 
In particular, a local agency also pays close attention to its fill rate. Let $R^f_l$ denote the fill rate of local agency $l = 1, \ldots, L$, where $L$ represents the total number of local agencies. Fill rate $R^f_l$ can be measured as the percentage of resources requested by disaster-affected people being met in the area served by local agency $l$ \citep{noyan_stochastic_2016}. 
The central agency is concerned about the fairness in allocating resources to local agencies and aims to ensure that the allocation decision would not systematically disadvantage any local agency \citep{bertsimas_efficiency_2012,huang_modeling_2015}. 
A common measure for the fairness of allocation is the equality of fill rates \citep{huang_modeling_2015},
which can be operationalized by the standard deviation of fill rates among local agencies:
\vspace{-0.1cm}
\begin{equation}
\label{eq:fairness}
	\big(\frac{1}{ L}\sum_{l=1}^{ L}(R^f_l-\bar{R}^f)^2 \big)^{\frac{1}{2}}, \vspace{-0.1cm}
\end{equation}
where $\bar{R}^f= \sum_{l=1}^{L}R^f_l/ L$ represents the average fill rate across all $ L$ local agencies. Clearly, high standard deviation of fill rates indicates high divergence of fill rates among local agencies, hence low fairness of allocation.

In a simulation, we run the following procedure for our method and each benchmark method. 

\vspace{0.1cm}
\begin{enumerate}
    \vspace{0.1cm}
    \item Execute a method at each local agency to decide the quantity of resources requested from the local agency for each resource type.
    \vspace{0.1cm}
    \item The central agency, after receiving resource requests from all local agencies,  adopts the following commonly used proportional (ration) allocation policy \citep{chen_modeling_2012}:
    for any resource type, if the stockpile at the central agency is sufficient to meet all requests, allocate according to the requests; otherwise, allocate all available stockpiles to local agencies by the amounts proportional to their requests.
    \vspace{0.1cm}
    \item Evaluate the performance of the method in terms of how well each of the three objectives discussed above is satisfied. 
\end{enumerate}

We considered three local agencies in our simulation. Using the simulation mechanism described in Appendix~\ref{apd:syn}, we simulated demands for three types of resources at each local agency. Different local areas might be affected by a disaster differently. As a result, the intensity of demand arrivals at different local agencies could be different. 
To simulate this situation, we set parameters $\nu$ (background trend in the self-correcting process) and $\lambda_0$ (base occurrence rate in the Hawkes process) to be 2 for one local agency; we set these parameters to be 1 and 0.5 for the other two agencies, respectively.
We performed $200$ simulation runs. 
Let $\I_k$ denote the stockpile level of resource type $k$ at the central agency, $k=1,2,3$.
We set $\I_k$ as the median of the actual demands for resource type $k$ across 200 simulation runs, which corresponds to $(\I_1, \I_2, \I_3)=(198, 220, 198)$. In this way, our simulation covers a wide range of degrees of inventory capacity tightness relative to the actual demand. More specifically, the actual demand for type $k$ resources exceeds the inventory capacity $50\%$ of the time and is below the inventory capacity $50\%$ of the time in the simulation. Among 200 simulation runs, the level of under-capacity is as high as 80\%.

Tables~\ref{tab:multi} summarizes the performance of each compared method, averaged across 200 simulation runs. 
As reported, our method significantly outperforms each benchmark in satisfying every objective ($p < 0.01$).\footnote{ 
The percentage improvements by our method over benchmarks LogNormMix-PRR, A-NDTT-PRR, AttnMC-PRR, CTDRP-PRR on fill rate are not as large as other improvements reported in the table. This is partly due to the fact that these benchmarks employ our PRR method, which effectively decides types and quantities of resources requested from the central agency to satisfy local demands.} 
The superior performance of our method is attributed to its CNM-TPP component that predicts future demands more accurately and its PRR component that effectively decides types and quantities of requested resources based on both currently unfulfilled demands and predicted future demands.
Consequently, resource requests generated by our method not only lead to low deprivation cost but also reflect true demands from local agencies more accurately, which in turn increases fill rate and improves resource allocation fairness.

\vspace{-0.3cm}
\begin{table}[H]
	\TABLE
	{ Performance Comparison on Objectives of Different Stakeholders 
	\\($\I_1=198$, $\I_2=220$, $\I_3=198$) \label{tab:multi}}
{\begin{tabular}{C{110pt} C{100pt} C{90pt} C{120pt}}
\hline
\down\textbf{Method}                                             & \begin{tabular}[c]{@{}c@{}}Average Unit Demand\\ Deprivation Cost    \end{tabular} & Average Fill Rate &
\begin{tabular}[c]{@{}c@{}}Average Standard Deviation \\ of  Fill Rates    \end{tabular}\\ \hline
\down\begin{tabular}[c]{@{}c@{}}CNM-PRR\\ (Our Method)  \end{tabular} &  \$28.01 & 0.927 &  0.063 \\
LogNormMix-PRR                                                         & \begin{tabular}[c]{@{}c@{}} \$33.54 \\ (16.49\%) \end{tabular}& \begin{tabular}[c]{@{}c@{}} 0.898 \\ (3.21\%) \end{tabular}& \begin{tabular}[c]{@{}c@{}}0.085 \\ (25.53\%) \end{tabular}\\
  A-NDTT-PRR                                                         & \begin{tabular}[c]{@{}c@{}} \$32.06\\ (12.64\%) \end{tabular} & \begin{tabular}[c]{@{}c@{}} 0.903 \\ (2.56\%) \end{tabular}& \begin{tabular}[c]{@{}c@{}} 0.080 \\ (21.24\%) \end{tabular}\\
  AttnMC-PRR                                                         & \begin{tabular}[c]{@{}c@{}} \$31.22\\ (10.29\%) \end{tabular}& \begin{tabular}[c]{@{}c@{}} 0.910 \\ (1.77\%) \end{tabular}& \begin{tabular}[c]{@{}c@{}} 0.093 \\ (31.79\%) \end{tabular}\\
  CTDRP-PRR                                                         & \begin{tabular}[c]{@{}c@{}} \$35.30 \\ (20.66\%) \end{tabular}& \begin{tabular}[c]{@{}c@{}} 0.906 \\ (2.26\%) \end{tabular}& \begin{tabular}[c]{@{}c@{}} 0.106 \\ (40.22\%) \end{tabular}\\
  LR-NV                                                         &
  \begin{tabular}[c]{@{}c@{}} \$52.98 \\ (47.13\%) \end{tabular}& \begin{tabular}[c]{@{}c@{}} 0.829 \\ (11.78\%) \end{tabular}& \begin{tabular}[c]{@{}c@{}} 0.098 \\ (35.68\%) \end{tabular}\\
  DL-NV                                                         &
  \begin{tabular}[c]{@{}c@{}} \$48.29\\ (42.00\%) \end{tabular}& \begin{tabular}[c]{@{}c@{}} 0.888\\ (4.29\%) \end{tabular}& \begin{tabular}[c]{@{}c@{}} 0.095 \\ (33.89\%) \end{tabular} \\
  LogNormMix-IFCFS                                                         & \begin{tabular}[c]{@{}c@{}} \$288.11\\ (90.28\%) \end{tabular}& \begin{tabular}[c]{@{}c@{}} 0.519 \\ (78.51\%) \end{tabular}& \begin{tabular}[c]{@{}c@{}} 0.125 \\ (49.58\%) \end{tabular}\\
\hline
\end{tabular}
}
	{}
\begin{center}
\vspace{6pt}
\footnotesize{Note: The percentage improvement by our method over a benchmark is listed in parentheses.}
\end{center}
\end{table}

\section{Discussion and Conclusion}
\label{sec:conclusion}
\subsection{Contributions}
\label{sec:conclusion:contr}
Disaster response is critical to save lives and reduce damages in the aftermath of a disaster. Fundamental to disaster response operations is the management of disaster relief resources. Prior resource management research overlooks the problem of deciding optimal quantities of resources requested by a local agency. In response to this research gap, we formulate a new resource management problem, i.e., proactive resource request problem. To solve the problem, we develop a novel TPP model to predict future demands and propose an effective solution method to the problem. We demonstrate the superior performance of our method over prevalent existing methods using both real world and simulated data.

Our study belongs to the computational genre of design science research in Information Systems, which develops computational methods to solve business and societal problems and aims at making
methodological contributions \citep{rai_editors_2017,gupta_traits_2018}.
Accordingly, one contribution of our study is the formulation of a new and important resource management problem for disaster response, which proactively decides optimal quantities of requested resources based on both currently unfulfilled demands and future demands. By solving the problem, our study adds to extant literature with the following methodological contributions. First, we propose a novel TPP model to accommodate salient characteristics of the problem. Specifically, our proposed TPP model differs from existing TPP models in its learning objective, mark embedding function, and event generation layer. Second, we formulate the problem as a stochastic optimization model and analyze its properties in Theorems \ref{thm:consis} and \ref{thm:np}. We then develop an efficient solution method to the problem based on the analyzed properties and show its effectiveness in Theorem \ref{thm:gap}.

\subsection{Implications for Disaster Response Management Practices}
\label{sec:conclusion:implication}
Our study offers several design principles that guide the development of a humanitarian resource management system for disaster response. As demonstrated in our study, a predictive analytics method characterizes uncertainties and predicts future demands. Taking the predicted future demands as inputs, a prescriptive analytics method then searches through the policy space and discovers a resource request policy that optimizes the objective of disaster response.
In addition, we show that, in comparison to benchmark methods, our method that integrates predictive and prescriptive analytics can attain higher fairness in allocating resources among local agencies. Without a prescriptive analytics method, a predictive analytics method alone cannot produce effective resource request decisions for disaster response. On the other hand, a prescriptive analytics method by itself cannot generate a practical resource request policy as it lacks the understanding of real world uncertainties. Accordingly, we have

\vspace{0.2cm}
\begin{itemize}[label={}, leftmargin=5.5mm, rightmargin=5.5mm]
\item \textbf{Design Principle (DP) 1:} A predictive analytics method should go hand in hand with a prescriptive analytics method, in order to timely, effectively, and fairly satisfy demands for disaster relief resources in a highly volatile and uncertain environment, in which the distribution of demands is non-stationary.  
\end{itemize}
\vspace{0.2cm}

We demonstrate substantial benefits of proactive resource request in alleviating sufferings of disaster-affected people. For example, we show that the average unit demand deprivation cost incurred by our method is significantly lower than that incurred by the reactive resource request method. As a result, we have

\vspace{0.2cm}
\begin{itemize}[label={}, leftmargin=5.5mm, rightmargin=5.5mm]
\item \textbf{DP 2:} Practitioners need to reconsider the current practice of reactive resource request and adopt the way of proactive resource request in designing a humanitarian resource management system, which takes both currently unfulfilled demands and future demands into consideration. 
\end{itemize}
\vspace{0.2cm}

Information technology (IT), especially machine learning, enables local disaster relief agencies to predict future demands by analyzing patterns of past demands. We show that a carefully designed TPP model can effectively characterize the dynamics of non-stationary demands arising from a disaster and predict future demands. Such prediction empowers local agencies to act with foresight and avoid myopic decisions, thereby saving more lives and properties. Therefore, our study demonstrates the value of IT in general and machine learning in particular for effective disaster response. Consequently, we have 

\vspace{0.2cm}
\begin{itemize}[label={}, leftmargin=5.5mm, rightmargin=5.5mm]
\item \textbf{DP 3:} A critical precursor to proactive resource request is the prediction of future demands. Therefore, a humanitarian resource management system should make effective use of proper predictive analytics technologies for future demand prediction.
\end{itemize}
\vspace{0.2cm}

Finally, our study shows that it is viable to collect and analyze demand data from social media platforms to decide quantities of requested disaster relief resources. It thus demonstrates the value of social media data for effective disaster response. Considering that social media platforms have become a major means for requesting disaster relief resources, a humanitarian resource management system should have the capability of listening demands of disaster-affected people from social media \citep{abbasi_don_2019} and allocating resources based on the demands. Therefore, we suggest
\vspace{0.2cm}
\begin{itemize}[label={}, leftmargin=5.5mm, rightmargin=5.5mm]
\item \textbf{DP 4:} A humanitarian resource management system should be capable of collecting demand data from social media platforms in real time, discovering demand patterns from huge amounts of collected data, and making intelligent resource request decisions based on the discovered patterns. 
\end{itemize}
\vspace{0.2cm}
Social media data, however, are voluntarily contributed and such voluntary nature of social media data undermines their quality, which in turn creates challenges for designing humanitarian resource management systems. For example, missing values in social media data could impact the accuracy of future demand prediction and the effectiveness of resource request decisions. Moreover, information sharing among central and local agencies is critical for effective disaster response and thus information interoperability among systems located at central and local agencies is another design challenge. To address these challenges, IS research on data quality and management could provided effective solutions \citep{xu2023deep}.          

\subsection{Limitations and Future Research}
\label{sec:conclusion:limit}
Our study has limitations and can be extended in several directions. 
First, our CNM-TPP model is trained with a sequence of observed demands to predict future demands. Thus, like many other machine learning algorithms, CNM-TPP also has the cold-start problem, i.e., how to predict future demands when there are few or even no observed demands. To address this limitation, future work could develop a transfer learning solution, which learns a model using observed demands in other similar disasters and adapts the model to predict future demands for the focal disaster. 
Second, while some disaster relief resources are not reusable (e.g., food), some others are reusable (e.g., pump). Different from non-reusable resources, a reusable resource could satisfy multiple demands. Therefore, another area worthy of future research is to extend our method to decide optimal quantities of requested resources by considering the differences between reusable and non-reusable resources.
Third, our proposed method is a single period method, which is executed in a rolling horizon manner to solve the resource request problem. Future study could extend our work to make resource request decisions for longer time period. 
Given the sequential nature of the problem, a viable solution method could be one based on reinforcement learning \citep{sutton_reinforcement_2018}.
Finally, this study solves one resource management problem for disaster response, i.e., the resource request problem. Future work could integrate our deep learning-based proactive resource request method with solution methods to other important resource management problems (e.g., resources allocation and transportation problems) such that we can have a more holistic resource management method for effective disaster response.


\bibliographystyle{informs2014} 
\bibliography{bibs/Project-PRR,bibs/Project-PRR-hz} 
\clearpage

\renewcommand{\theequation}{\arabic{equation}} 
\renewcommand{\thetable}{\arabic{table}}

\ECSwitch

\ECHead{Supplementary Materials}

\begin{appendices}

\setcounter{algorithm}{0}
\renewcommand\thealgorithm{A.\arabic{algorithm}}

\section{Additional Details of Future Demand Prediction}\label{apd:add_detail}

\subsection{Derivation of Equation \eqref{eq:p_S}}
\label{apd:add_detail:eq1}

We compute $p(S)$ as
\begin{align*}
p(S) &= P(t_{L+1}>T) \prod_{i=1}^{L} p_{e}\big( e_i | \mathcal{H}_{i-1} \big) \\
  &= P(t_{L+1}>T) \prod_{i=1}^{L} p_{t}\big( t_i | \mathcal{H}_{i-1} \big) p_{m}\big( m_i | \mathcal{H}_{i-1} \big) \\
 &= P(\tau_{L+1}>T-t_L) \prod_{i=1}^{L} p_{\tau}\big( \tau_i | \mathcal{H}_{i-1} \big) p_{m}\big( m_i | \mathcal{H}_{i-1} \big),
\end{align*}
where $\mathcal{H}_{i-1}=<e_1, e_2, ...,e_{i-1}>$ denotes the sequence of past events before the $i$-th event $e_i$, $i=1,2,\dots,L$, and $\mathcal{H}_0=\emptyset$. The first step of the derivation follows directly from the chain rule. In this step, $P(t_{L+1}>T)$ is the probability that the $(L+1)$-th event occurs after time $T$ and $p_{e}\big( e_i | \mathcal{H}_{i-1} \big)$ denotes the probability of the $i$-th event conditioning on past events $\mathcal{H}_{i-1}$. Next, given $e_i=(t_i, m_i)$ and the widely adopted conditional independence assumption between time and mark of an event in the TPP literature \citepsec{du_recurrent_2016_a, shchur_intensity-free_2020_a}, $p_{e}\big( e_i | \mathcal{H}_{i-1} \big)$ can be computed as the multiplication of $p_{t}\big( t_i | \mathcal{H}_{i-1} \big)$ and $p_{m}\big( m_i | \mathcal{H}_{i-1} \big)$, where $p_{t}\big( t_i | \mathcal{H}_{i-1} \big)$ denotes the probability density of the $i$-th event occurring at $t_i$ conditioning on past events $\mathcal{H}_{i-1}$ and $p_{m}\big( m_i | \mathcal{H}_{i-1} \big)$ represents the probability mass of the mark of the $i$-th event being $m_i$ conditioning on $\mathcal{H}_{i-1}$. Lastly, following the common practice of modeling interarrival times of TPP \citepsec{du_recurrent_2016_a, shchur_intensity-free_2020_a}, we define interarrival time $\tau_i=t_i - t_{i-1}$ and have the final factorization of $p(S)$, where $p_{\tau}\big( \tau_i | \mathcal{H}_{i-1} \big)$ is the probability density of the interarrival time between the $i$-th event and the $(i-1)$-th event being $\tau_i$ conditioning on $\mathcal{H}_{i-1}$.

\subsection{Simplified Mark Embedding and Generation for the Case of $a_k\in\{0,1\}$}
\label{apd:add_detail:mark}

The general mark embedding and generation proposed in Section \ref{sec:method:ctpp} can be simplified for the special situation where we only know whether a resource type is requested. In this situation, $a_k=0$ denotes that type $k$ resource is not requested in an event whereas $a_k=1$ indicates the resource type is requested, $k=1,2,\dots,K$. The mark embedding function for this special situation can be defined as:
\begin{equation}
  \label{eq:me_detail_deg}
    f_m(e_{i-1}) = \sum_{k=1}^{K} W^{(r)}_{:,k} I(a^{(i-1)}_{k}=1)
\end{equation}
where $W^{(r)}_{:,k}$ is the embedding vector representing resource type $k$, $a^{(i-1)}_{k}=1$ if resource type $k$ is requested in event $e_{i-1}$ and $a^{(i-1)}_{k}=0$ otherwise, and indicator function $I(.)$ returns one if its internal condition is true and zero otherwise. Under this special situation, the generation of an event mark can be formulated as the generation of $K$ binary codes, where each code denotes whether a resource type is requested in the event. Since each code takes the value of $0$ or $1$, it can be modeled by a Bernoulli distribution. Accordingly, the probability $p_{m}\big( m_i | \mathcal{H}_{i-1} \big)$ that the mark of event $e_i$ is $m_i$ conditioning on $\mathcal{H}_{i-1}$ is given by
\begin{equation}
    \label{eq:cpmf_m_deg}
    p_m\big( m_i |\mathcal{H}_{i-1} \big) = \prod_{k=1}^{K} P\big( a^{(i)}_{k} | a^{(i)}_{1:k-1}, \mathcal{H}_{i-1} \big), 
\end{equation}
with
\begin{equation}
  \label{eq:p_m_k_deg}
  P\big( a^{(i)}_{k} | a^{(i)}_{1:k-1}, \mathcal{H}_{i-1} \big) = \begin{cases}
p_{k}^{(i)} & \text{if } a^{(i)}_{k}=1,\\\\
		1-p_{k}^{(i)} & \text{if } a^{(i)}_{k}=0,
	\end{cases}
\end{equation}
where $a^{(i)}_{k}$ denotes whether resource type $k$ is requested in event $e_i$, $k=1,2,\dots,K$, and $p_{k}^{(i)}$ is given by  
\begin{subequations}
 \label{eq:bernoulli_p}
 \begin{align}
 h_{i,k}^{(\lambda)} &= \text{RNN}_{\lambda}\Big( h_{i,k-1}^{(\lambda)}, W^{(r)}_{:,k-1} I(a^{(i)}_{k-1}=1) \Big) \\
p_{k}^{(i)} &= \text{Sigmoid} \Big( {W^{(r)}_{:,k}}^T h_{i,k}^{(\lambda)} \Big) 
 \end{align}
\end{subequations}
where $\text{Sigmoid}(x) = 1/\big( 1+\exp(-x) \big)$ for scalar input $x$. 

\subsection{Differentiable Event Sampling} \label{apd:add_detail:repar_trick}

Training CNM-TPP with gradient descent needs to evaluate the gradient of the objective function defined in Equation \eqref{eq:obj_full}, which in turn requires to compute the gradient $\nabla_{\Theta} {D (e, \tilde{e})}$ of $D (e, \tilde{e})$, because $D (e, \tilde{e})$ is part of the objective function. By Equation \eqref{eq:d_e}, we have
\begin{equation} 
\label{eq:gradient_wrt_Dee}
\begin{aligned}
   \nabla_{\Theta} {D (e, \tilde{e})} &= \nabla_{\Theta} \Big ( \sum_{k=1}^K (t-\tilde{t})^2 \log c_k + \sum_{k=1}^K (a_k - \tilde{a}_k)^2 \log c_k \Big) \\
  &= \sum_{k=1}^K \big( 2(\tilde{t}-t) \log c_k \big) \nabla_{\Theta} \tilde{t} + \sum_{k=1}^K \big( 2(\tilde{a}_k-a_k) \log c_k \big) \nabla_{\Theta} \tilde{a}_k.
\end{aligned}
\end{equation}  
 The second step in Equation \eqref{eq:gradient_wrt_Dee} follows from the fact that both $\tilde{t}$ and $\tilde{a}_k$ are sampled from the underlying TPP parameterized by $\Theta$, and therefore are on their own functions of $\Theta$. To evaluate $\nabla_{\Theta} \tilde{t}$ and $\nabla_{\Theta} \tilde{a}_k$, we design the following two algorithms that respectively enable $\tilde{t}$ and $\tilde{a}_k$ differentiable w.r.t. $\Theta$ based on the reparameterization trick \citepsec{kingma_introduction_2019_a}. The general idea of the trick is to draw samples from a distribution that does not depend on $\Theta$, and then transform the resulted samples with differentiable functions parameterized by $\Theta$. We note that it is relatively more difficult to make $\tilde{a}_k$ differentiable w.r.t. $\Theta$, because there is no well-established reparameterization trick for the Poisson distribution. 

To make $\tilde{t}$ differentiable w.r.t. $\Theta$, it is equivalent to make $\hat{\tau}$ differentiable w.r.t. $\Theta$, because $\tilde{t}$ is computed from $\hat{\tau}$ through lines \ref{alg:sample_p_i_star:set_t}, \ref{alg:sample_p_i_star:set_e_hat} and \ref{alg:sample_p_i_star:set_e_tilde} in Algorithm \ref{alg:sample_p_i_star}. Therefore, we implement line \ref{alg:sample_p_i_star:draw_tau} in Algorithm \ref{alg:sample_p_i_star} as a subroutine specified by Algorithm \ref{alg:sample_gm}. 
\begin{algorithm}[h]
	\caption{The Procedure of Sampling from LogNormalMixture}
  \label{alg:sample_gm}
    \textbf{Input:} The precomputed distribution parameters $(\alpha, \mu, \sigma)$, temperature $\zeta$ \\
    \textbf{Output:} A sampled interarrival time
	\begin{algorithmic}[1]
	\State Draw $u \in R^{n_z}$ from $\text{Uniform}(0,1)$ and compute $g=-\log (-\log u )$
    \State Compute mixture component $\hat{\alpha} \in R^{n_z}$ as $\hat{\alpha}= \text{Softmax}((\log \alpha + g)/\zeta)$ 
    \State Draw $\epsilon \in R^{n_z}$ from $\text{Normal}(0,1)$
    \State Compute $\hat{\tau} = \sigma^T (\hat{\alpha} \odot \epsilon) + \mu^T \hat{\alpha}$
    \State Compute $\hat{\tau} =  \exp (\hat{\tau}) \quad \quad \quad$ //$\exp()$: exponential function
    \State \Return $\hat{\tau}$
	\end{algorithmic} 
\end{algorithm}
The first two lines in Algorithm \ref{alg:sample_gm} follow from the reparameterization trick for the categorical distribution \citepsec{jang_categorical_2017_a}. When the temperature scalar $\zeta$ is close to zero, the trick amounts to create a one-hot vector $\hat{\alpha}$ where the position of the hot entry is randomly selected according to the categorical distribution specified by $\alpha$, which is equivalent to drawing a mixture component according to the categorical distribution. The next two lines in Algorithm \ref{alg:sample_gm} follow from the reparameterization trick for the Gaussian distribution \citepsec{kingma_introduction_2019_a}. 

To make $\tilde{a}_k$ differentiable w.r.t. $\Theta$, it is equivalent to make $\hat{a}_k$ differentiable w.r.t. $\Theta$, because $\tilde{a}_k$ is computed from $\hat{a}_k$ through lines \ref{alg:sample_p_i_star:set_e_hat} and \ref{alg:sample_p_i_star:set_e_tilde} of Algorithm \ref{alg:sample_p_i_star}. In practice, the requested amount of any resource type is finite. Therefore, instead of modeling $\hat{a}_k$ as a Poisson distribution, we model it as a truncated Poisson distribution parameterized by mean $\lambda_k$ (which depends on $\Theta$ through Equation \ref{eq:poisson_mean}) and upper bound $\bar{A}$. By definition, we have
\begin{equation}
\label{eq:truncated_poisson_density}
 P(\hat{a}_k = x | \lambda_k,  \bar{A} ) = \frac{\lambda_k^{x} \exp (-\lambda_k)}{x!} / Z_k
\end{equation}
where 
\begin{equation*}
Z_k = \sum_{x=0}^{\bar{A}} \frac{\lambda_k^{x} \exp (-\lambda_k)}{x!}
\end{equation*}
is the normalization constant. Equipped with Equation \eqref{eq:truncated_poisson_density}, $\hat{a}_k$ can be viewed as following a categorical distribution specified by the probability vector
\begin{equation}
\label{eq:truncated_poisson}
\text{Cat}(\lambda_k, \bar{A}) = [  P(\hat{a}_k = 0 | \lambda_k,  \bar{A} ),   P(\hat{a}_k = 1 | \lambda_k,  \bar{A} ), \dots,  P(\hat{a}_k = \bar{A} | \lambda_k,  \bar{A} ) ]^T,
\end{equation}
which allows us to implement line \ref{alg:sample_p_i_star:draw_m} in Algorithm \ref{alg:sample_p_i_star} via the reparameterization trick for the categorical distribution  \citepsec{jang_categorical_2017_a}. In particular, line \ref{alg:sample_p_i_star:draw_m} in Algorithm \ref{alg:sample_p_i_star} is implemented as a subroutine specified by Algorithm \ref{alg:sample_quantity}. 
\begin{algorithm}[h]
	\caption{The Procedure of Sampling from Truncated Poisson}
  \label{alg:sample_quantity}
    \textbf{Input:} The precomputed mean $\lambda_k$, upper bound $\bar{A}$, temperature $\zeta$ \\
    \textbf{Output:} A sampled quantity
	\begin{algorithmic}[1]
	\State Draw $u \in R^{\bar{A}+1}$ from $\text{Uniform}(0,1)$ and compute $g=-\log (-\log u )$
	\State Compute $\text{Cat}(\lambda_k, \bar{A})$ in accordance with Equations \ref{eq:truncated_poisson_density} and \ref{eq:truncated_poisson}
    \State Compute $b_k = \text{Softmax}((\log \text{Cat}(\lambda_k, \bar{A}) + g)/\zeta)$ 
    \State Compute $\hat{a}_k = \sum_{x=0}^{\bar{A}} b_{k,x} x$ 
    \State \Return $\hat{a}_k$
	\end{algorithmic} 
\end{algorithm}
In line 3 of the algorithm, the computed $b_k$ is a vector of length $\bar{A}+1$, where $b_{k,x}$ is the $x$th entry of the vector, and  $\sum_{x=0}^{\bar{A}} b_{k,x} = 1 $ due to the Softmax transformation. When temperature $\zeta$ is close to zero, $b_{k}$ reduces to a one-hot vector, where the position of its hot entry is randomly selected according to the categorical distribution specified by $\text{Cat}(\lambda_k, \bar{A})$. Consequently, line 4 of the algorithm is in effect drawing $\hat{a}_k$ from the truncated Poisson distribution specified by $\lambda_k$ and $\bar{A}$.

\subsection{Inference Procedure of CNM-TPP}
\label{apd:add_detail:inf}

We present Algorithm \ref{alg:sample_future}, the inference procedure of CNM-TPP. The algorithm takes the embedding vector $h_{L}$ of the training sequence $S$ as an input and predicts a sequence of future events (demands) $\hat{S}$ in the period of $(T, T_{+}]$. As shown, it initially sets $\hat{S}$ as a empty sequence (line 1). It then iteratively generates the time $\hat{t}$ (lines 3--5) and mark $\hat{m}$ (lines 8--12) of a future event, conditioning on $\hat{h}$, updates $\hat{h}$ with the newly generated event (line \ref{alg:sample_future:compute_h}), and adds the generated event to $\hat{S}$ (line \ref{alg:sample_future:set_S}). The inner while loop at line \ref{alg:sample_future:inner_while} ensures that the first generated event occurs after $T$.

\begin{algorithm}[]
	\caption{The Inference Procedure of CNM-TPP}
  \label{alg:sample_future}
    \textbf{Input:} $h_{L}$: embedding vector of training sequence $S$, $(T, T_{+}]$: prediction period \\
    \textbf{Output:} $\hat{S}$: a sequence of events in $(T, T_{+}]$ 
	\begin{algorithmic}[1]
    \State Set $\hat{h}=h_{L}$, $\hat{t}=t_{L}$, and $\hat{S}=<>$
    \While {$\hat{t} \le T_{+} $}  \label{alg:sample_future:outer_while}
    \State  Compute $(\alpha, \mu, \sigma)$ from $\hat{h}$ via Equation \eqref{eq:compute_gm_param}
    \State Draw interarrival time $\hat{\tau}$ from $\text{LogNormalMixture}(\alpha, \mu, \sigma)$ specified by Equation \eqref{eq:cpdf_tau}  \label{alg:sample_future:draw_tau}
    \State Set event time $\hat{t}=\hat{t}+\hat{\tau}$ \label{alg:sample_future:compute_t}
    \While{$\hat{t} \le T $} \label{alg:sample_future:inner_while}
        \State Repeat line \ref{alg:sample_future:draw_tau} and \ref{alg:sample_future:compute_t}
    \EndWhile
    \For{$k=1,2,\dots,K$}
    
    \State Compute mean of the Poisson distribution for resource type $k$ from $\hat{h}$ and $\hat{a}_{1:k-1}$ \\
    \hspace{1.2cm} via Equation \eqref{eq:poisson_mean}
    \State Sample $\hat{a}_k$ from the Poisson distribution specified by Equation \eqref{eq:p_m_k} \label{alg:sample_future:draw_m}
    
    \EndFor
    \State Set $\hat{m}=(\hat{a}_1, \hat{a}_2, \dots, \hat{a}_K)$, and $\hat{e}=(\hat{t}, \hat{m})$ 
    \State Compute $\hat{h}=\text{RNN}(\hat{h}, \hat{e})$ via Equation \eqref{eq:hist_emb} \label{alg:sample_future:compute_h}
    \State Set $\hat{S}=\hat{S}\oplus<\hat{e}>$ \label{alg:sample_future:set_S}
    \EndWhile
    \State \Return $\hat{S}$
	\end{algorithmic} 
\end{algorithm}

\clearpage
\section{CNM-TPP Specifications}
\label{apd:para_spc}

We provide a detailed guide through of the computation flow of our CNM-TPP method and its parameter settings, by following \citetsec{zhu_deep_2021-2_a}.

\subsection{History Embedding Layer Specifications}

\label{apd:para_spc_his}

\begin{figure}[H]
	\FIGURE
	{\includegraphics[scale=0.45]{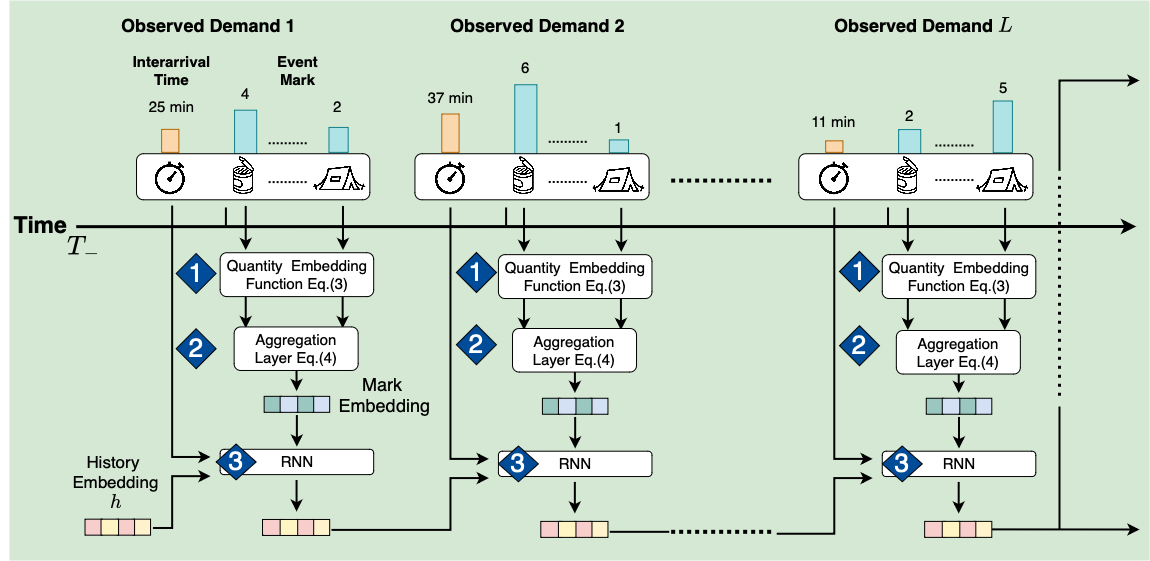}}
	{Architecture of History Embedding Layer in CNM-TPP. \label{fg:model_his}}	{}
\end{figure}

\begin{table}[H]
	\TABLE
	{ History Embedding Layer Specifications\label{tab:model_his}}
	{\begin{tabular}{||C{50pt}|C{200pt}|C{100pt}|C{100pt}||}
\hline
ID & Name & Input Shape & Output Shape \\ \hline
 1  &   Quantity Embedding Function Eq.~\eqref{eq:pos_encode}   & 3  & (3, 64) \\ \hline
 2  &  Aggregation Layer Eq.~\eqref{eq:me_detail}   & (3, 64) & 64 \\ \hline
 3  &  RNN    & 64, 1, 64 & 64 \\ \hline
\end{tabular}}
	{}
\end{table}

Figure \ref{fg:model_his} illustrates the more detailed computation flow of the history embedding layer of CNM-TPP, where we mark key computation nodes in the flow with numbered diamonds. Table \ref{tab:model_his} then reports the input shape and output shape of each computation node under the parameter setting of $K=3$ (the number of resource types) and $n_e=64$ (the hidden size of the RNN layer). We use this parameter setting for all evaluations in Sections \ref{sec:eval:results} and \ref{sec:eval:multistakeholder}, as well as  Appendices~\ref{apd:rob} to~\ref{apd:syn}. 
The two computation nodes annotated by diamond 1 and 2 constitute the mark embedding layer, which takes a vector of length $K=3$ as input and produces a mark embedding vector of length $n_e=64$.
Each RNN node marked by diamond 3 in Figure \ref{fg:model_his} takes three inputs: the first one is the hidden state vector of length $n_e=64$ from the previous time step, the second one is the interarrival time of the current event which is a scalar of length $1$, and the third one is the mark embedding vector of length $n_e=64$ of the current event. These three inputs are combined according to Equation~\eqref{eq:hist_emb} to generate the hidden state vector of length $n_e=64$ at the current time step.

\subsection{Event Generation Layer Specifications}
\label{apd:para_spc_egl}

\begin{figure}[H]
	\FIGURE
	{\includegraphics[scale=0.55]{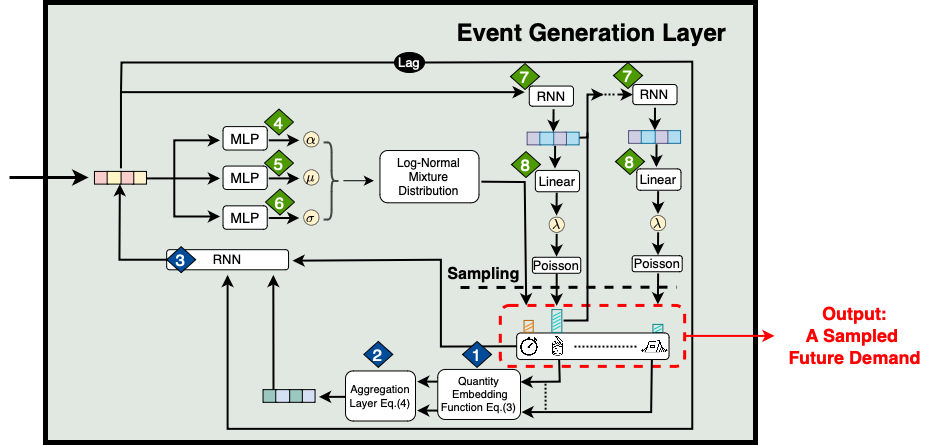}}
	{Architecture of Event Generation Layer in CNM-TPP. \label{fg:model_egl}}	{}
\end{figure}
\vspace{-0.5cm}
\begin{table}[H]
	\TABLE
	{ Event Generation Layer Specifications\label{tab:model_egl}}
	{\begin{tabular}{||C{50pt}|C{200pt}|C{100pt}|C{100pt}||}
\hline
ID & Name & Input Shape & Output Shape \\ \hline
 4  &   MLP    &   64  &  64  \\ \hline
 5  &   MLP     &  64  &    64     \\ \hline
 6  &   MLP    &  64    &   64  \\ \hline
 7  &  RNN    &       64, 64 &         64  \\ \hline
 8  &  Linear Layer    &       64        &        1             \\ \hline
\end{tabular}}
	{}
\end{table}

As described in Section \ref{apd:para_spc_his}, the history embedding layer of CNM-TPP generates a hidden state vector of length $n_e=64$ at each time step. Based on the hidden state vector at a particular time step, Figure \ref{fg:model_egl} illustrates the detailed computation flow of the event generation layer in Figure~\ref{fg:model}, where we mark key computation nodes in the flow with numbered diamonds. 
Table \ref{tab:model_egl} then reports the input shape and output shape of each computation node under the parameter setting of $K=3$ (the number of resource types), $n_e=64$ (the hidden size of the RNN layer) and $n_z=64$ (the number of mixture components). 
 We use this parameter setting for all evaluations in Sections \ref{sec:eval:results} and \ref{sec:eval:multistakeholder}, as well as  Appendices~\ref{apd:rob} to~\ref{apd:syn}. 

The three computation nodes annotated by diamond 4, 5 and 6 generate the parameters characterizing the distribution of the interarrival time of the next demand event. Each of the three computation nodes takes the hidden state vector of length $n_e=64$ as input, and outputs a vector of distribution parameters of length $n_z=64$ via a MLP layer, such that the $z$-th entry of the vector will be used to compute the parameter of the $z$-th mixture component in accordance with Equation~\eqref{eq:compute_gm_param}. We use a single linear layer to instantiate the MLP layer. 
The resulted log-normal mixture distribution is capable of evaluating the density of any given interarrival time in accordance with Equation~\eqref{eq:cpdf_tau}, as well as sampling a batch of possible interarrival times for the next demand event via Algorithm \ref{alg:sample_gm}. The former is necessary for evaluating the NLL objective specified by Equation~\eqref{eq:obj_nll}, while the latter is required by the CSD objective defined with Equation~\eqref{eq:obj_acd}.

The two computation nodes annotated by diamond 7 and 8 generate the parameters characterizing the distribution of the resource quantities requested in the next demand event. The RNN node marked by diamond 7 will be unrolled for $K=3$ times to produce the mean scalars of the $K=3$ Poisson distributions. Each RNN node takes two inputs: one is the hidden state vector of length $n_e=64$ from the previous resource type, and the other is the input vector of length $n_e=64$ embedding the quantity information of the current resource type. The output of each RNN node is a hidden state vector of length $n_e=64$, which is then fed into a linear layer in accordance with Equation~\eqref{eq:poisson_mean} to produce the scalar regarded as the mean of the Poisson distribution for the current resource type. The resulted sequence of Poisson distributions is capable of evaluating the density of any given quantity vector of length $K=3$ in accordance with Equation~\eqref{eq:cpmf_m} and Equation~\eqref{eq:p_m_k}, as well as sampling a batch of possible quantity vectors for the next demand event via Algorithm \ref{alg:sample_quantity}. The former is necessary for evaluating the NLL objective specified by Equation~\eqref{eq:obj_nll},  while the latter is required by the CSD objective defined with Equation~\eqref{eq:obj_acd}.

\clearpage
\section{Proof of Theorem~\ref{thm:consis}}
\label{apd:thm:consis}

To prove Theorem~\ref{thm:consis}, we first introduce Proposition~\ref{prop:g_k} and Lemma~\ref{lm:uniformly}.

\subsection{Proposition~\ref{prop:g_k}}
\label{apd:thm:consis:prop}
\begin{proposition}\rm
\label{prop:g_k}
For any given $\tilde{Q}_k$, function $g_k(x_k,\tilde{Q}_k)$ defined by Equation~\eqref{eq:g_k_explicit} is a (i) continuous piece-wise linear, (ii) monotonically non-decreasing, and (iii) concave function w.r.t $x_k\in  \real^+$, $k= 1,2,\dots,K$.
\end{proposition}

\proof{$\rm Proof.$ }
We first prove $g_k(x_k,\tilde{Q}_k)$ is a continuous piece-wise linear function w.r.t  $x_k\in\real^+$. We rewrite $g_k(x_k,\tilde{Q}_k)$ as:
\begin{equation}
	g_k(x_k,\tilde{Q}_k)=\begin{cases}
 x_kB^k_{1} &\text{if } 0\leq x_k<\tilde{q}^k_1,\\
 \sum_{j=1}^{1 } \tilde{q}^k_jB^k_{j} +(x_k-\sum_{j=1}^{1} \tilde{q}^k_j)B^k_{2} 
		& \text{if } \tilde{q}^k_1\leq x_k < \sum_{j=1}^{2 }\tilde{q}^k_j\\
\sum_{j=1}^{2} \tilde{q}^k_jB^k_{j} +(x_k-\sum_{j=1}^{2} \tilde{q}^k_j)B^k_{2} 
		& \text{if } \sum_{j=1}^{2 }\tilde{q}^k_j\leq x_k < \sum_{j=1}^{3 }\tilde{q}^k_j\\
		\cdots\\
 \sum_{j=1}^{\tilde{n}^k}\tilde{q}^k_jB^k_{j} & \text{if } x_k\geq\sum_{j=1}^{\tilde{n}^k }\tilde{q}^k_j.
	\end{cases}
\end{equation}
It is clear that $g_k(\cdot,\tilde{Q}_k)$ is a (i) piece-wise linear function. To prove its continuity, we thus only need to show $g_k(\cdot,\tilde{Q}_k)$ is continuous at its kinks $\{s^0_1,s^0_2,\dots,s^0_{\tilde{n}^k}\}$, where $s^0_J=\sum_{j=1}^J q_j^k$, $J =1,2,\dots,\tilde{n}^k$. For a kink $s^0_J$, we have
\begin{equation}
\label{eq:left_lim}
\begin{aligned}
g_k(s^{0-}_{J},\tilde{Q}_k)&=\lim_{x_k\rightarrow s^{0-}_{J}} \sum_{j=1}^{J-1}\tilde{q}^k_jB^k_j+(x_k-\sum_{j=1}^{J-1}\tilde{q}^k_j)B^k_{J}\\
& =\sum_{j=1}^{J-1}\tilde{q}^k_jB^k_j+\lim_{x_k\rightarrow (\sum_{j=1}^J q_j^k)^-}(x_k-\sum_{j=1}^{J-1}\tilde{q}^k_j)B^k_{J}\\
&=\sum_{j=1}^{J} \tilde{q}^k_jB^k_{j}
\end{aligned}
\end{equation}
and 
\begin{equation}
\label{eq:right_lim}
\begin{aligned}
g_k(s^{0+}_{J},\tilde{Q}_k)&=\lim_{x_k\rightarrow s^{0+}_{J}} \sum_{j=1}^{J}\tilde{q}^k_jB^k_j+(x_k-\sum_{j=1}^{J}\tilde{q}^k_j)B^k_{J}\\
& =\sum_{j=1}^{J}\tilde{q}^k_jB^k_j+\lim_{x_k\rightarrow (\sum_{j=1}^J q_j^k)^+}(x_k-\sum_{j=1}^{J}\tilde{q}^k_j)B^k_{J}\\
&=\sum_{j=1}^{J} \tilde{q}^k_jB^k_{j}\\
&=g_k(s^{0-}_{J},\tilde{Q}_k).
\end{aligned}
\end{equation}
 Therefore, piece-wise linear function $g_k(\cdot,\tilde{Q}_k)$ is continuous at its kinks. Therefore, $g_k(\cdot,\tilde{Q}_k)$ is continuous on $\real^+$. Moreover, since its continuity and $ B^k_{j}$ is positive $j=1,2,\dots,\tilde{n}^k$, $k=1,2,\dots,K$, $g_k(x_k,\tilde{Q}_k)$ is thus (ii) monotonically non-decreasing w.r.t $x_k$ and bounded by $\sum_{j=1}^{\tilde{n}^k} q_j^kB^k_{j}$.

To prove the concavity of $g_k(\dot,\tilde{Q}_k)$, we define linear functions
\begin{equation}
    h^J_k(x_k,\tilde{Q}_k)=\sum_{j=1}^{J} \tilde{q}^k_jB^k_{j} +(x_k-\sum_{j=1}^{J} \tilde{q}^k_j)B^k_{J+1}, \quad J=0,1,\dots, \tilde{n}^k-1.
\end{equation} Subsequently, we have
\begin{equation}
	g_k(x_k,\tilde{Q}_k)=\begin{cases}
  h^0_k(x_k,\tilde{Q}_k) &0\leq x_k<\tilde{q}^k_1,\\
 h^1_k(x_k,\tilde{Q}_k)
		& \text{if } \tilde{q}^k_1\leq x_k < \sum_{j=1}^{2 }\tilde{q}^k_j\\
h^2_k(x_k,\tilde{Q}_k)
		& \text{if } \sum_{j=1}^{2 }\tilde{q}^k_j\leq x_k < \sum_{j=1}^{3 }\tilde{q}^k_j\\
		\cdots\\
 \sum_{j=1}^{\tilde{n}^k}\tilde{q}^k_jB^k_{j} & \text{if } x_k\geq\sum_{j=1}^{\tilde{n}^k }\tilde{q}^k_j.
	\end{cases}
\end{equation}
Consider a $x_k\in \real^+$, where $ \sum_{j=1}^{J_k}\tilde{q}^k_j\le x_k< \sum_{j=1}^{J_k+1}\tilde{q}^k_j$. We denote sets $\J_-=\{J:J<J_k| J\in \{0,1,\dots, \tilde{n}^k-1\}\}$ and $\J_+=\{J:J>J_k|J\in \{0,1,\dots, \tilde{n}^k-1\}\}$, respectively. Recall that $B_1^k\ge B_2^k\ge\dots\ge B_{\tilde{n}^k}^k$. If $\J_-$ is nonempty, for any $J_-\in \J_-$, we have
\begin{equation}
\begin{aligned}
h^{J_-}_k(x_k,\tilde{Q}_k) &= \sum_{j=1}^{J_-} \tilde{q}^k_jB^k_{j} +(x_k-\sum_{j=1}^{J_-} \tilde{q}^k_j)B^k_{J_-+1}\\
&=\sum_{j=1}^{J_-} \tilde{q}^k_jB^k_{j}+ (\sum_{j=1}^{J_k} \tilde{q}^k_j-\sum_{j=1}^{J_-} \tilde{q}^k_j)B^k_{J_-+1}+(x_k-\sum_{j=1}^{J_k} \tilde{q}^k_j)B^k_{J_-+1}\\
&=\sum_{j=1}^{J_-} \tilde{q}^k_jB^k_{j}+ \sum_{j=J_-+1}^{J_k} \tilde{q}^k_jB^k_{J_-+1}+(x_k-\sum_{j=1}^{J_k} \tilde{q}^k_j)B^k_{J_-+1}\\
&> \sum_{j=1}^{J_k} \tilde{q}^k_jB^k_{j} +(x_k-\sum_{j=1}^{J_k} \tilde{q}^k_j)B^k_{J_k+1}\\
&= h^{J_k}_k(a,\tilde{Q}_k) = g_k(x_k,\tilde{Q}_k). 
\end{aligned}
\end{equation}
Similarly, if $\J_+$ is nonempty, for any $J_+\in \J_+$, we have
\begin{equation}
\begin{aligned}
h^{J_+}_k(x_k,\tilde{Q}_k) &= \sum_{j=1}^{J_+} \tilde{q}^k_jB^k_{j} +(x_k-\sum_{j=1}^{J_+} \tilde{q}^k_j)B^k_{J_++1}\\
&= \sum_{j=1}^{J_k} \tilde{q}^k_jB^k_{j} +\sum_{j=J_k+1}^{J_+} \tilde{q}^k_jB^k_{j}  +(\sum_{j=1}^{J_k} \tilde{q}^k_j-\sum_{j=1}^{J_+} \tilde{q}^k_j)B^k_{J_++1}+(x_k-\sum_{j=1}^{J_k} \tilde{q}^k_j)B^k_{J_++1}\\
&=\sum_{j=1}^{J_k} \tilde{q}^k_jB^k_{j}+(x_k-\sum_{j=1}^{J_k} \tilde{q}^k_j)B^k_{J_++1}+\sum_{j=J_k+1}^{J_+} \tilde{q}^k_jB^k_{j}-\sum_{j=J_k+1}^{J_+} \tilde{q}^k_j B^k_{J_++1}\\
&>\sum_{j=1}^{J_k} \tilde{q}^k_jB^k_{j}+(x_k-\sum_{j=1}^{J_k} \tilde{q}^k_j)B^k_{J_++1}\\
&= h^{J_k}_k(x_k,\tilde{Q}_k) = g_k(x_k,\tilde{Q}_k). 
\end{aligned}
\end{equation}
Thus, we have
\begin{equation}
\begin{aligned}
\label{eq:min_h}
    g_k(x_k,\tilde{Q}_k)&=h^{J_k}_k(x_k,\tilde{Q}_k)\\
    &=\min\{h^0_k(x_k,\tilde{Q}_k), h^1_k(a,\tilde{Q}_k), \dots, h^{\tilde{n}^k-1}_k(x_k,\tilde{Q}_k),V\},
\end{aligned}
\end{equation}
where $V=\sum_{j=1}^{\tilde{n}^k}\tilde{q}^k_jB^k_{j}$. Since $h^{J}_k(\cdot)$ is linear (concave) function, $J=0,1,\dots, \tilde{n}^k-1$,  Equation~\eqref{eq:min_h} immediately implies that $g_k(x_k,\tilde{Q}_k)$ is a concave function w.r.t $x_k$.

This completes the proof of Proposition~\ref{prop:g_k}. \Halmos

\subsection{Lemma~\ref{lm:uniformly}}
\label{apd:thm:consis:lm}
\begin{lemma}\rm
\label{lm:uniformly}
Let $C$ be $[0,\frac{W}{w_1}]\times\dots\times[0,\frac{W}{w_K}]$, which is an nonempty compact set. $\G(\M)$ is finite valued and continuous on $C$, and $\G^\Psi(\M)$ converges to $\G(\M)$ uniformly on $C$ as $\Psi\rightarrow\infty$.
\end{lemma}

\proof{$\rm Proof.$}

Let $g(\M, \tilde{\Q})=\sum_{k=1}^K g_k(x_k,\tilde{Q}_k)$. As shown in Proposition~\ref{prop:g_k}, $g_k(x_k,\tilde{Q}_k)$ is continuous w.r.t $x_k $ on $\real^+$, so $g(\M, \tilde{\Q})$ is continuous w.r.t $\M$. Moveover, since $g_k(x_k,\tilde{Q}_k)$ is non-decreasing w.r.t $x_k$, we have
\begin{equation*}
g_k(x_k,\tilde{Q}_k)\le g_k(\frac{W}{w_k},\tilde{Q}_k), 
\end{equation*}
when  $x_k\in[0,\frac{W}{w_k}]$. Recall that $B_1^k\ge B_2^k\ge\dots\ge B_{\tilde{n}^k}^k$. Consequently, we have
$g_k(\frac{W}{w_k},\tilde{Q}_k)\le \frac{W}{w_k}B_1^k$. Because $g_k(x_k,\tilde{Q}_k)$ is non-negative when $x_k\in[0,\frac{W}{w_k}]$, $ k=1,2,\dots,K$, it implies 
\begin{align*}
|g(\M, \tilde{\Q})|&=|\sum_{k=1}^K g_k(x_k,\tilde{Q}_k)|=\sum_{k=1}^K g_k(x_k,\tilde{Q}_k)\le \sum_{k=1}^K g_k(\frac{W}{w_k},\tilde{Q}_k)\le \sum_{k=1}^K \frac{W}{w_k}B_1^k.
\end{align*}
$g(\M, \tilde{\Q})$ is thus dominated by a constant when $\M\in C$. Therefore, we have
\begin{align*}
|\G(\M)|&=|E_{\tilde{\Q}}(g(\M, \tilde{\Q}))|\le|\sum_{k=1}^K \frac{W}{w_k}B_1^k|< +\infty,
\end{align*}
which means $\G(\M)$ is finite valued.

We next prove the continuity of $\G(\M)$ on $C$. Consider a point $\M^0\in C$. For any sequence $(\M^0_l)$ of points on $C$ which converges to $\M^0$, we have 
\begin{equation*}
    \lim_{l\rightarrow\infty}\G(\M^0_l)=\lim_{l\rightarrow\infty}E_{\tilde{\Q}}(g(\M^0_l,\tilde{\Q}))=E_{\tilde{\Q}}(\lim_{l\rightarrow\infty}g(\M^0_l,\tilde{\Q})),
\end{equation*}
where the interchange in the equation is justified by the Lebesgue Dominated Convergence Theorem. Since $g(\M, \tilde{\Q})$ is continuous w.r.t. $\M$ on $C$, $\lim_{l\rightarrow\infty}g(\M^0_l,\tilde{\Q})=g(\M^0,\tilde{\Q})$, which immediately implies that
\begin{equation*}
    \lim_{l\rightarrow\infty}\G(\M^0_l)=\G(\M^0).
\end{equation*}
Hence, $\G(\M)$ is continuous on $C$.
Then, Lemma~\ref{lm:uniformly} follows from Proposition 7 in \citetsec{shapiro_monte_2003_a} immediately. \Halmos

\subsection{Proof of Theorem~\ref{thm:consis}}
Notice that feasible regions of problem~\eqref{pb:new} and its approximated problem~\eqref{pb:SAA} are both closed subsets of $C$. Since $\G(\M)$ and $\G^\Psi(\M)$ are finite valued on $C$, the optimal solutions of  problem~\eqref{pb:new} and its approximated problem~\eqref{pb:SAA} are thus nonempty and contained in $C$. Given Lemma~\ref{lm:uniformly}, Theorem~\ref{thm:consis} follows from a direct
application of Proposition 6 in \citetsec{shapiro_monte_2003_a} immediately.

\clearpage

\section{Proof of Theorem~\ref{thm:np}}
\label{apd:thm:np}
 SAA approximated problem~\eqref{pb:SAA} is formulated as 
\begin{equation}
\begin{aligned}
\max_{\M} \quad & \frac{1}{\Psi}\sum_{k=1}^K\sum_{\psi=1}^\Psi g_k(x_k,\tilde{Q}^{(\psi)}_k)\\
\textrm{s.t.} \quad & \sum_k w_kx_k\le W,\\
\quad & x_k\in \intg^+,
\end{aligned}
\end{equation}
where $g_k(x_k,\tilde{Q}^{(\psi)}_k)$ is given by Equation~\eqref{eq:g_k_explicit} and $\Psi$ denotes the number of samples. It suffices to show that SAA approximated problem~\eqref{pb:SAA} is NP-hard if we can show that the problem is NP-hard for $\Psi=1$. We thus focus on proving the NP-hardness of the scenario that $\Psi=1$. Given $\Psi=1$, we drops the superscript $(\psi)$ from the the problem formulation and rewrite the problem as:
\begin{equation}
\begin{aligned}
\label{pb:psi1}
\max_{\M} \quad & \sum_{k=1}^K g_k(x_k,\tilde{Q}_k)\\
\textrm{s.t.} \quad & \sum_k w_kx_k\le W,\\
\quad & x_k\in \intg^+.
\end{aligned}
\end{equation}

Consider the knapsack problem, which is a well-known NP-hard problem. Given a set of items with specific weights and values, the knapsack problem determines how to pack items in the knapsack so that the total value is maximized while satisfying the constraint of weight. To prove Theorem~\ref{thm:np} , we show that the knapsack problem is a special case of SAA approximated problem~\eqref{pb:SAA}. Let $\tilde{n}^k=1$, $\tilde{t}^k_1=\tau_k$ and $|\tilde{Q}_k|=\tilde{q}^k_1>W/w_k$, $k=1,2,\dots,K$. We have constant
\begin{equation}
    B^k_1=e^{\phi-bc_k\tau_k}(e^{bc_k\xi_k}-e^{bc_kT_+})
\end{equation}
by Equation~\eqref{eq:reduction}. Consequently, in this case, problem~\eqref{pb:psi1} is equivalent to 
\begin{equation}
\begin{aligned}
\max_{\M} \quad & \sum_{k=1}^K  B^k_1x_k\\
\textrm{s.t.} \quad & \sum_k w_kx_k\le W,\\
\quad & x_k\in \intg^+,
\end{aligned}
\end{equation}
which is a knapsack problem where $w_k$ and $ B^k_1$ are specific weight and value for item $k$, respectively.

This completes the proof of Theorem~\ref{thm:np}.

\clearpage

\section{Robustness Analysis}\label{apd:rob}

For robustness check, we varied the transportation capacity $W$ from $180$ to $230$, with an increment of 10.\footnote{Varying $W$ has a similar effect as varying $w_k$. Thus, we omit the robustness check with $w_k$.} As reported in Table~\ref{tab:rob_cap}, our method consistently outperforms each benchmark across all investigated transportation capacities.\footnote{It is expected that the deprivation cost incurred by a method decreases as the transportation capacity increases, consistent with what we report in Table~\ref{tab:rob_cap}.} 

\begin{table}[H]
	\TABLE
	{Prformance Comparison on Average Unit Demand Deprivation Cost: Varying $W$\\
	($w_1=1$, $w_2=1$, $w_3=1$, $c_1=2$, $c_2=4$, $c_3=2$ ) \label{tab:rob_cap}}
{%
\resizebox{\textwidth}{!}
{\begin{tabular}{C{110pt} C{70pt} C{70pt} C{70pt} C{70pt}C{70pt}C{70pt}}
\hline
\textbf{Method}                                             & $W=180$ & $W=190$ & $W=200$ & $W=210$ & $W=220$ & $W=230$\\ \hline
\begin{tabular}[c]{@{}c@{}}CNM-PRR\\ (Our Method)  \end{tabular} &  \$57.45 & \$32.46 & \$25.69 & \$22.64 & \$21.49 & \$20.69 \\
  ReR                                                         & \begin{tabular}[c]{@{}c@{}} \$35014.77\\ (99.84\%) \end{tabular}&\begin{tabular}[c]{@{}c@{}} \$26719.96\\ (99.88\%) \end{tabular} & \begin{tabular}[c]{@{}c@{}} \$21905.50\\ (99.88\%)\end{tabular}& \begin{tabular}[c]{@{}c@{}} \$17531.21\\ (99.87\%) \end{tabular}
  & \begin{tabular}[c]{@{}c@{}} \$15350.77\\ (99.86\%) \end{tabular}
  & \begin{tabular}[c]{@{}c@{}} \$13195.61\\ (99.84\%) \end{tabular}\\
  LogNormMix-PRR                                                         & \begin{tabular}[c]{@{}c@{}} \$66.52\\ (13.64\%) \end{tabular}& \begin{tabular}[c]{@{}c@{}} \$41.28\\ (21.38\%) \end{tabular}& \begin{tabular}[c]{@{}c@{}} \$32.32\\ (20.51\%) \end{tabular}& \begin{tabular}[c]{@{}c@{}} \$27.60\\ (17.96\%) \end{tabular}& \begin{tabular}[c]{@{}c@{}} \$24.14\\ (10.96\%) \end{tabular}& \begin{tabular}[c]{@{}c@{}} \$21.79\\ (5.04\%) \end{tabular}\\
  A-NDTT-PRR                                                         & \begin{tabular}[c]{@{}c@{}} \$64.18 \\ (10.50\%) \end{tabular}& \begin{tabular}[c]{@{}c@{}} \$32.18 \\ (14.87\%) \end{tabular}& \begin{tabular}[c]{@{}c@{}} \$30.28 \\ (15.15\%) \end{tabular}& \begin{tabular}[c]{@{}c@{}} \$25.94 \\ (12.72\%) \end{tabular}& \begin{tabular}[c]{@{}c@{}} \$22.97 \\ (6.43\%) \end{tabular}& \begin{tabular}[c]{@{}c@{}} \$21.46 \\ (3.56\%) \end{tabular}\\
  AttnMC-PRR                                                         & \begin{tabular}[c]{@{}c@{}} \$65.32 \\ (12.05\%) \end{tabular}& \begin{tabular}[c]{@{}c@{}} \$39.26 \\ (17.32\%) \end{tabular}& \begin{tabular}[c]{@{}c@{}} \$31.26 \\ (17.82\%) \end{tabular}& \begin{tabular}[c]{@{}c@{}} \$26.35 \\ (14.08\%) \end{tabular}& \begin{tabular}[c]{@{}c@{}} \$23.55 \\ (8.71\%) \end{tabular}& \begin{tabular}[c]{@{}c@{}} \$21.63 \\ (4.33\%) \end{tabular}\\
  CTDRP-PRR                                                         & \begin{tabular}[c]{@{}c@{}} \$79.26 \\ (27.53\%) \end{tabular}& \begin{tabular}[c]{@{}c@{}} \$48.25 \\ (32.74\%) \end{tabular}& \begin{tabular}[c]{@{}c@{}} \$37.35 \\ (31.22\%) \end{tabular}& \begin{tabular}[c]{@{}c@{}} \$31.09 \\ (27.18\%) \end{tabular}& \begin{tabular}[c]{@{}c@{}} \$27.48 \\ (21.79\%) \end{tabular}& \begin{tabular}[c]{@{}c@{}} \$25.63 \\ (19.26\%) \end{tabular}\\
  LR-NV                                                         &
  \begin{tabular}[c]{@{}c@{}} \$86.08 \\ (33.27\%) \end{tabular}& \begin{tabular}[c]{@{}c@{}} \$77.45 \\ (58.09\%) \end{tabular}& \begin{tabular}[c]{@{}c@{}} \$70.14 \\ (63.37\%) \end{tabular}& \begin{tabular}[c]{@{}c@{}} \$63.99 \\ (64.61\%) \end{tabular}& \begin{tabular}[c]{@{}c@{}} \$58.87 \\ (63.49\%) \end{tabular}& \begin{tabular}[c]{@{}c@{}} \$54.26 \\ (61.86\%) \end{tabular}\\
  DL-NV                                                         &
  \begin{tabular}[c]{@{}c@{}} \$71.13 \\ (19.24\%) \end{tabular}& \begin{tabular}[c]{@{}c@{}} \$64.46 \\ (49.65\%) \end{tabular}& \begin{tabular}[c]{@{}c@{}} \$58.76 \\ (56.28\%) \end{tabular}& \begin{tabular}[c]{@{}c@{}} \$53.86 \\ (57.96\%) \end{tabular}& \begin{tabular}[c]{@{}c@{}} \$49.98 \\ (56.99\%) \end{tabular}& \begin{tabular}[c]{@{}c@{}} \$47.06 \\ (56.03\%) \end{tabular}\\
  LogNormMix-IFCFS                                                         & \begin{tabular}[c]{@{}c@{}} \$347.02\\ (83.45\%) \end{tabular}& \begin{tabular}[c]{@{}c@{}} \$273.35\\ (88.13\%) \end{tabular}& \begin{tabular}[c]{@{}c@{}} \$225.95\\ (88.63\%) \end{tabular}& \begin{tabular}[c]{@{}c@{}} \$193.36\\ (88.29\%) \end{tabular}& \begin{tabular}[c]{@{}c@{}} \$171.79\\ (87.49\%) \end{tabular}& \begin{tabular}[c]{@{}c@{}} \$154.54\\ (86.61\%) \end{tabular}\\
\hline
\end{tabular}
}
}
	{}
\begin{center}
\vspace{6pt}
\footnotesize{Note: The percentage improvement by our method over a benchmark is listed in parentheses.}
\end{center}
\end{table}

In addition, we varied $c_k$ and analyzed the performance of the methods in two additional scenarios of importance scores. First, we reduced the importance score $c_2$ of type 2 resources (i.e., lifesaving kit) to $3$ while keeping the other two importance scores the same as before. In this scenario, type 2 resources still have the highest importance score but the difference between the importance score of type 2 resources and that of the other two types is not as significant as before. As reported in Tables \ref{tab:rob_c2}, our method significantly outperforms each benchmark in this scenario ($p<0.01$). By comparing the experimental results in Table~\ref{tab:henan_res} and those in Table \ref{tab:rob_c2}, we notice that the cost reduction by our method decreases as the value of $c_2$ is reduced from 4 to 3. In comparison to the importance scores used in Table~\ref{tab:henan_res}, the importance scores employed in Table \ref{tab:rob_c2} are more homogeneous and hence the benefit of the CSD learning objective diminishes, which in turn leads to the reduced advantage of our method over the benchmarks. If all resource types are equally important for disaster relief (e.g., $c_1=c_2=c_3=2$), the boundary condition is met and the benefit of the CSD learning objective vanishes. However, in reality, different resource types are not equally important for disaster relief and the boundary condition is rarely met.

Second, we increased the importance score of type 3 resources (i.e., damage repair kit) to $3$ while keeping the other two importance score intact. Different from the original setting of $c_k$, all three types of resources have different importance scores under this scenario
As reported in Table \ref{tab:rob_c3}, our method still significantly outperforms each benchmark method ($p<0.01$).

\begin{table}[H]
	\TABLE
	{ Performance Comparison on Average Unit Demand Deprivation Cost: Varying $c_2$
	\\($W=200$,  $w_1=1$, $w_2=1$, $w_3=1$, $c_1=2$, $c_2=3$, $c_3=2$ ) \label{tab:rob_c2}}
	{\begin{tabular}{C{110pt} C{120pt} C{120pt}}
\hline
\textbf{Method}                                             & \begin{tabular}[c]{@{}c@{}}Average Unit Demand\\ Deprivation Cost   \end{tabular} & \begin{tabular}[c]{@{}c@{}}Cost Reduction \\by CNM-PRR \end{tabular} \\ \hline
\begin{tabular}[c]{@{}c@{}}CNM-PRR\\ (Our Method)  \end{tabular} & \$15.05 & \\
  ReR                                                        & \$1207.16 & 98.75\% \\
  LogNormMix-PRR                                                          & \$16.50 & 8.78\% \\
  A-NDTT-PRR & \$15.85 & 5.06\% \\  
  AttnMC-PRR   & \$16.14 & 6.72\% \\ 
  CTDRP-PRR   & \$24.93 & 39.62\% \\ 
  LR-NV  & \$35.73 & 57.87\% \\ 
  DL-NV   & \$36.81 & 59.11\% \\
  LogNormMix-IFCFS  & \$136.80 & 89.00\% \\
\hline
\end{tabular}}
	{}
\end{table}

\begin{table}[H]
	\TABLE
	{ Performance Comparison on Average Unit Demand Deprivation Cost: Varying $c_3$
	\\($W=200$,  $w_1=1$, $w_2=1$, $w_3=1$, $c_1=2$, $c_2=4$, $c_3=3$ ) \label{tab:rob_c3}}
	{\begin{tabular}{C{110pt} C{120pt} C{120pt}}
\hline
\textbf{Method}                                             & \begin{tabular}[c]{@{}c@{}}Average Unit Demand\\ Deprivation Cost   \end{tabular} & \begin{tabular}[c]{@{}c@{}}Cost Reduction \\by CNM-PRR \end{tabular} \\ \hline
\begin{tabular}[c]{@{}c@{}}CNM-PRR\\ (Our Method)  \end{tabular} & \$44.08 & \\
  ReR                                                        & \$22842.42 & 99.81\% \\
  LogNormMix-PRR                                                          & \$55.59 & 20.70\% \\
  A-NDTT-PRR & \$50.57 & 12.82\% \\  
  AttnMC-PRR   & \$52.35 & 15.79\% \\ 
  CTDRP-PRR   & \$99.55 & 55.71\% \\ 
  LR-NV  & \$73.13 & 39.72\% \\ 
  DL-NV   & \$90.72 &51.41\% \\
  LogNormMix-IFCFS  & \$369.63 & 88.07\% \\
\hline
\end{tabular}}
	{}
\end{table}

\clearpage

\section{Model Extension}\label{apd:model_ext}

In this appendix, we extend our model from two perspectives: (a) the resource arrival time (lead time) $T_+$ is assumed to be stochastic rather than deterministic, and (b) holding costs associated with the over-requested resources are added to the model. Both extensions are technical in nature and will not cause structural changes in our main results and insights. More specifically, extension (a) will add an additional layer of expectation operations w.r.t. the stochastic lead time $T_+$ on the objective function of optimization problem~\eqref{pb:new}. Extension (b) will introduce an extra term to the cost reduction function~\eqref{eq:g_k_explicit}, which leads to some adjustments in the greedy algorithm.

Let requested resource arrival time $T_+$ be a random variable with the support $[T,\widetilde{T_+}]$, where $\widetilde{T_+}$ represents its upper limit. We further assume that $\widetilde{T_+} < \xi_k$ for all $k$ to ensure the delivery from the central agency arrives before all demands are eventually met. For a unit demand of type $k$ resource that arrives at time $t$, its deprivation cost reduction, due to the unit of type $k$ resource requested at $T$ and arrives at $T_+$, is given by:
\begin{equation}
\label{eq:ext_reduction}
	B^k(T_+,t)
	=e^{\phi-bc_kt}(e^{bc_k\xi_k}-e^{bc_kT_+}).
\end{equation}
Notice that the deprivation cost reduction is now also a function of delivery arrival time $T_+$, whereas the formulation in Equation~\eqref{eq:reduction} treats $T_+$ as a fixed constant. Following notations in Section~\ref{sec:method:prrm}, $\tilde{Q}_k$ represents the sequence of net demands waiting to be fulfilled by time $T_+$, $k=1,2,\dots, K$. The length of $\tilde{Q}_k$ is $n_k$. We similarly denote $B^k(T_+,\tilde{t}^k_j)$ by $B^k_j(T_+)$, where $\tilde{t}_j^k$ is arrival time of the $j$-th demand in $\tilde{Q}_k$ and $B^k(T_+,\tilde{t}^k_j)$ is given by Equation \eqref{eq:ext_reduction}, $j = 1, 2,\dots, \tilde{n}^k$. Similar to Equation~\eqref{eq:b_order}, we have 
\begin{equation}
\label{eq:ext_b_order}
    B_1^k(T_+)\ge B_2^k(T_+)\ge\dots\ge B_{\tilde{n}^k}^k(T_+).
\end{equation}

Besides deprivation cost reduction, we explicitly model holding cost of over-requested resources in the extended model. Let $h_k^c$ be the holding cost associated with one unit of over-requested type $k$ resource. We are now ready to define the cost reduction $g^E_k(x_k,T_+,\tilde{Q}_k)$ if $x_k$ units of resources, requested at time $T$, will be available at a stochastic time $T_+$ to meet the sequence of net demands $\tilde{Q}_k$ by arrival time $T_+$, $k = 1, 2,\dots, K$.
\begin{equation}
\label{eq:ext_g_k_explicit}
	g^E_k(x_k,T_+,\tilde{Q}_k)=\begin{cases}
		\sum_{j=1}^{J^k } \tilde{q}^k_jB^k_{j}(T_+) +(x_k-\sum_{j=1}^{J^k} \tilde{q}^k_j)B^k_{J^k +1}(T_+) 
		& \text{if } x_k < |\tilde{Q}_k|,\\\\
		\sum_{j=1}^{ \tilde{n}^k} \tilde{q}^k_jB^k_{j}(T_+)-h_k^c(x_k-|\tilde{Q}_k|) & \text{if } x_k \geq |\tilde{Q}_k|,
	\end{cases}
\end{equation}
where $|\tilde{Q}_k|$ represents the total requested amount in $\tilde{Q}_k$ and $J^k$ is the largest sequence index $J$ such that $\sum_{j=1}^{J} \tilde{q}^k_j \leq x_k$. 
Compared to cost reduction $g_k(x_k,\tilde{Q}_k)$ in original problem~\eqref{pb:new}, new cost reduction $g^E_k(x_k,T_+,\tilde{Q}_k)$ defined in Equation~\eqref{eq:ext_g_k_explicit} takes arrival time $T_+$ as an additional variable rather than a fixed parameter and penalizes over-requested resources by the holding costs incurred. Consequently, we are able to define the extended stochastic optimization problem as follows: 
\begin{equation}
\begin{aligned}
\label{pb:ext_new}
\max_{\M}  \quad & \G^E(\M) := E_{T_+,\tilde{\Q}}(\sum^K_{k=1}g^E_k(x_k,T_+,\tilde{Q}_k))\\
\textrm{s.t.} \quad & \sum_k w_k x_k\le W,\\
\quad & x_k\in \intg^+,\quad k=1,2,\dots, K,
\end{aligned}
\end{equation}
where the expectation is taken over $\tilde{\Q}=(\tilde{Q}_1, \tilde{Q}_2, \dots, \tilde{Q}_K)$ and stochastic arrival time $T_+$.

To solve this difficult stochastic optimization problem~\eqref{pb:ext_new}, we employ the Sample Average Approximation (SAA) method as discussed in Section~\ref{sec:method:prrm:saa}. More specifically, we first generate a sufficient number of samples of arrival time $T_+$: $T_+^{(\upsilon)}$, $\upsilon = 1, 2,\dots, \Upsilon$, where $\Upsilon$ is the number of samples. Next, for each sampled arrival time $T_+^{(\upsilon)}$, we generate samples of  $\tilde{\Q}$, denoted as $\tilde{\Q}^{(\upsilon,\psi)}$, $\psi = 1, 2,\dots$, according to Section~\ref{sec:method:prrm:saa}, where $\Psi$ is the number of samples generated. Then we have the following approximation for the expected cost function:
\begin{equation}
E_{T_+,\tilde{\Q}}(\sum^K_{k=1}g^E_k(x_k,T_+,\tilde{Q}_k)) \approx   \frac{1}{\Upsilon\Psi}\sum_{\upsilon=1}^{\Upsilon}\sum_{\psi=1}^{\Psi}\sum_{k=1}^{K}g^E_k(x_k,T_+^{(\upsilon)},\tilde{Q}_k^{(\upsilon,\psi)}).
\end{equation}
We thus obtain the following SAA approximation of the extended stochastic optimization problem \eqref{pb:ext_new}: 
\begin{equation}
\begin{aligned}
\label{pb:ext_SAA}
\max_{\M} \quad &  \G^{\Upsilon,\Psi}(\M):= \frac{1}{\Upsilon\Psi}\sum_{\upsilon=1}^{\Upsilon}\sum_{\psi=1}^{\Psi}\sum_{k=1}^{K}g^E_k(x_k,T_+^{(\upsilon)},\tilde{Q}_k^{(\upsilon,\psi)}) \\
\textrm{s.t.} \quad & \sum_k w_kx_k\le W,\\
\quad & x_k\in \intg^+,\quad k=1,2,\dots, K.
\end{aligned}
\end{equation}
We further prove that the optimal objective function value of deterministic optimization problem \eqref{pb:ext_SAA} converges to that of original stochastic optimization problem \eqref{pb:ext_new}, as the sample size increases. Thus, the proposed SAA approximation method is also valid and accurate for the extended problem. 
\begin{theorem}\rm
\label{ext_thm:consis}
Let  $\pi_E^*$ and $\Pi^*_{E,\Psi}$ be the optimal objective function values of problems \eqref{pb:ext_new} and \eqref{pb:ext_SAA}, respectively. We have
$$
\lim_{\Upsilon,\Psi \to \infty} \sup |\Pi^*_{E,\Psi} - \pi_E^*| = 0.
$$
\proof{$\rm Proof.$}  See Appendix~\ref{apd:model_ext:thm}. \endproof
\end{theorem}

Clearly, problem~\eqref{pb:ext_SAA} is NP-hard since it includes the NP-hard problem \eqref{pb:SAA} as its special case. We next propose an efficient greedy heuristic to solve the problem. Denote 
\begin{equation} \label{eq:ext_gpsik}
\G^{\upsilon,\Psi}_k(x_k) = \frac{1}{\Psi}\sum_{\psi=1}^{\Psi}g^E_k(x_k,T_+^{(\upsilon)},\tilde{Q}_k^{(\upsilon,\psi)}), \quad k = 1, 2, \dots, K,  \quad \upsilon = 1, 2, \dots, \Upsilon.
\end{equation}
Note that $\G^{\upsilon,\Psi}_k(x_k)$ inherits all nice structural properties of $G^{\Psi}_k(x_k)$ defined in Equation~\eqref{eq:gpsik}, except the monotonicity property. 
As a result, the left derivative of $\G^{\upsilon,\Psi}_k(x_k)$ at $s_i^{k,(\upsilon,\psi)}$, which denotes the corresponding location of the kink of $\G^{\upsilon,\Psi}_k(x_k)$ associated with the $i$-th demand in $\tilde{Q}_k^{(\upsilon,\psi)}$ in $x_k$-axis, becomes:
\begin{align}
\begin{aligned}
 \label{eq:ext_leftslope}
{\G^{\upsilon,\Psi}_k}^{\prime}(s_{i}^{k,(\upsilon,\psi)-}) &= \frac{1}{\Psi}\sum_{\varphi=1}^{\Psi} \big(B^k(T_+^{(\upsilon)},\tilde{t}^{k,(\upsilon,\varphi)}_{J^{(\upsilon)}(k,\psi,\varphi,i)})\times\mathds{1}(s_{i}^{k,(\upsilon,\psi)}\le|\tilde{Q}_k^{(\upsilon,\varphi)}|)- h^c_k\times\mathds{1}(|\tilde{Q}_k^{(\upsilon,\varphi)}|< s_{i}^{k,(\upsilon,\psi)}))\big)\\
&= \frac{1}{\Psi}\sum_{\varphi=1}^{\Psi} \big((B^k(T_+^{(\upsilon)},\tilde{t}^{k,(\upsilon,\varphi)}_{J^{(\upsilon)}(k,\psi,\varphi,i)})+h^c_k)\times\mathds{1}(s_{i}^{k,(\upsilon,\psi)}\le|\tilde{Q}_k^{(\upsilon,\varphi)}|)- h^c_k)\big),
\end{aligned}
\end{align}
where 
$$
J^{(\upsilon)}(k,\psi,\varphi,i) = \min j \in \{ 1,\dots, \tilde{n}^{k,(\upsilon,\varphi)}\}| s_j^{k,(\upsilon,\varphi)} \ge s_i^{k,(\upsilon,\psi)}.
$$
We are now ready to describe the details of the modified greedy algorithm for the extended model. We still relabel all kinks and slopes of $\G^{\upsilon,\Psi}_k(x_k)$, by sorting them according to the arrival times of their associated demands. Let $s^{k,(\upsilon)}_i$, $i = 1, 2,\dots, M^{k,(\upsilon)}$ be all kinks of function $\G^{\upsilon,\Psi}_k(x_k)$, such that $0 \leq s^{k,(\upsilon)}_1 \leq, \dots, \leq s^{k,(\upsilon)}_{M^{k,(\upsilon)}}$, where $M^{k,(\upsilon)}=\sum_{\psi=1}^\Psi\tilde{n}^{k,(\upsilon,\psi)}$.
Let $\bar{B}^{k,(\upsilon)}_i$ be the $i$-th slope of function $\G^{\upsilon,\Psi}_k(x_k)$ for all $x_k \in (s^{k,(\upsilon)}_{i-1}, s^{k,(\upsilon)}_i)$. Suppose $s^{k,(\upsilon)}_i$ is the kink associated with $\tilde{i}$-th demand in $\tilde{Q}_k^{(\upsilon,\psi)}$, i.e., $s^{k,(\upsilon)}_i = s^{k,(\upsilon,\psi)}_{\tilde{i}}$. Accordingly, $\bar{B}^{k,(\upsilon)}_i$ can be calculated by
\begin{equation}
\label{eq:ext_slope}
    \bar{B}^{k,(\upsilon)}_i={\G^{\upsilon,\Psi}_k}^{\prime}(s_{i}^{k,(\upsilon)-})= {\G^{\upsilon,\Psi}_k}^{\prime}(s^{k,(\upsilon,\psi)-}_{\tilde{i}})= \frac{1}{\Psi}\sum_{\varphi=1}^{\Psi} \big((B^k(T_+^{(\upsilon)},\tilde{t}^{k,(\upsilon,\varphi)}_{J^{(\upsilon)}(k,\psi,\varphi,\tilde{i})})+h^c_k)\times\mathds{1}(s_{\tilde{i}}^{k,(\upsilon,\psi)}\le|\tilde{Q}_k^{(\upsilon,\varphi)}|)- h^c_k)\big).
\end{equation}
Clearly, we have $\bar{B}^{k,(\upsilon)}_1  \geq \bar{B}^{k,(\upsilon)}_2\geq,\dots, \geq \bar{B}^{k,(\upsilon)}_{M^{k,(\upsilon)}}$. Similar to Algorithm~\ref{alg:greedy}, we rank all slopes for a sample of $T_+^{(\upsilon)}$ in descending order according to the ratio $\bar{B}^{k,(\upsilon)}_i/w_k$ among all resource types $k = 1, 2,\dots, K$ and slopes $i=1, 2,\dots,  M^{k,(\upsilon)}$.
Suppose the resulting sequence is $\PP^{(\upsilon)}$. For simplicity, we reuse notations defined in Section~\ref{sec:method:prrm}: $\kappa^{(\upsilon)}(\rho)$ is the type of requested resource associated with the $\rho$-th slope in sequence $\PP^{(\upsilon)}$ and $I_k^{(\upsilon)}(\rho)$ is the total number of slopes associated with type $k$ resource request among top $\rho$ ranked slopes in sequence $\PP^{(\upsilon)}$.  The greedy algorithm then follows sequence $\PP^{(\upsilon)}$ to iteratively calculated the requested quantity of type $k$ resources for sample $T_+^{(\upsilon)}$ of arrival time:
\begin{equation}
\label{eq:ext_xkj}
x_k^{(\upsilon)}(\rho) = 
\begin{cases}
		 s^{k,(\upsilon)}_{I^{(\upsilon)}_k(\rho)} & \text{if } I^{(\upsilon)}_k(\rho) \neq 0,\\
		0 & \text{otherwise,}
\end{cases}
\quad k =1,2,\dots, K.
\end{equation}
In contrast to Algorithm~\ref{alg:greedy}, slope $\bar{B}^{k,(\upsilon)}_i$ can be negative for large $i$. As a consequence, there are two stop criteria for the adapted greedy algorithm: (1) when the shipping capacity constraint is violated for the first time, and (2) when the slope becomes non-positive for the first time. Denote the $\rho$-th slope in ranked $\PP^{(\upsilon)}$ as $\bar{B}^{(\upsilon)}_\rho$. The corresponding sequence index is given by:
\begin{equation} \label{eq:ext_jw+}
\rho_S^{(\upsilon)} = \min \rho \in \{1,2,\dots,|\PP^{(\upsilon)}|\} |\quad \sum_{k=1}^K w_k x^{(\upsilon)}_k(\rho) \geq W  \lor  \bar{B}^{(\upsilon)}_\rho \leq 0.
\end{equation}
If $\bar{B}^{(\upsilon)}_{\rho_S^{(\upsilon)}} \leq 0$, we then obtain the optimal request quantity when arrival time is $T_+^{(\upsilon)}$: ${\M}^{(\upsilon)*} = (x_1^{(\upsilon)*}, \dots, x_K^{(\upsilon)*})$, where $x_k^{(\upsilon)*} = x_k^{(\upsilon)}(\rho_S^{(\upsilon)}-1)$, $k=1, 2,\dots, K$. If $\bar{B}^{k,(\upsilon)}_{\rho_S^{(\upsilon)}} > 0$ and $\sum_{k=1}^K w_k x^{(\upsilon)}_k(\rho_S^{(\upsilon)}) = W$, the optimal request quantity for type $k$ resource $x_k^{(\upsilon)*}$ becomes $x_k^{(\upsilon)}(\rho_S^{(\upsilon)})$, $k=1, 2,\dots, K$. Otherwise, we follows the same procedure as described in Section~\ref{sec:method:prrm:greedy} to generate a feasible solution via Equations~\eqref{eq:xfrac} and~\eqref{eq:xkg}. With solutions for each sampled arrival time $T_+^{(\upsilon)}$ available, the greedy algorithm takes average over all solutions and generates a feasible solution ${\M}^g$ to problem~\eqref{pb:ext_SAA}.  We formally present the proposed greedy algorithm for the extended model in Algorithm~\ref{alg:ext_greedy}.

Using the data set, evaluation procedure, and evaluation metric described in Section \ref{sec:eval:data}, we evaluated the performance of our proposed method for the extended model and compared it with those of benchmarks. In the evaluation, we assumed that the lead time of requested resources (i.e., $T_+-T$) followed a truncated gamma distribution with right truncation point at $\widetilde{T_+}-T$, consistent with the common assumption used in the inventory control literature \citepsec{dunsmuir_control_1989_a, gallego_inventory_2007_a, bischak_analysis_2014_a}. The scale parameter of truncated gamma distribution was set as 1, and its shape parameter was adjusted so that the expectation of the truncated gamma distribution was close to the lead time in Section~\ref{sec:eval:data} (i.e., 12 hours). Besides, we set the unit holding cost as \$1 for all resource types. Other experimental settings followed the empirical studies in Section~\ref{sec:eval:results}. Table~\ref{tab:ext_henan_res} reports the average cost (including deprivation cost and holding cost) of satisfying unit demand for each method. As shown in the table, 
our method outperforms the benchmarks by a range between $17.57\%$ and $99.76\%$. All the performance improvements by our method reported in Table~\ref{tab:ext_henan_res} are statistically significant ($p<0.01$).

\begin{table}[H]
	\TABLE
	{ Performance Comparison on Average Cost of Satisfying Unit Demand
	\label{tab:ext_henan_res}}
	{\begin{tabular}{C{110pt} C{120pt} C{120pt}}
\hline
\textbf{Method}                                             & \begin{tabular}[c]{@{}c@{}}Average Cost \\ of Satisfying Unit Demand    \end{tabular} & \begin{tabular}[c]{@{}c@{}}Cost Reduction \\by CNM-PRR \end{tabular} \\ \hline
\begin{tabular}[c]{@{}c@{}}CNM-PRR\\ (Our Method)  \end{tabular} & \$53.02 & \\
  ReR                                                        & \$21905.50 & 99.76\% \\
  LogNormMix-PRR                                                          & \$77.65 & 31.72\% \\
  A-NDTT-PRR                                                          & \$64.32 & 17.57\% \\  
  AttnMC-PRR                                                          & \$72.66 & 27.02\% \\ 
  CTDRP-PRR                                                          & \$82.22 & 35.51\% \\ 
  LR-NV                                                          & \$191.80 & 72.36\% \\ 
  DL-NV                                                          & \$162.76 & 67.42\% \\ 
LogNormMix-IFCFS  & \$265.44 & 80.02\% \\
\hline
\end{tabular}}
	{}
\end{table}

\begin{algorithm}[]
	\caption{A Greedy Algorithm to Solve Problem~\eqref{pb:ext_SAA}}
  \label{alg:ext_greedy}
    \textbf{Input:} $W$, $\{w_k|k=1,2,\dots,K\}$, $\{T_+^{(\upsilon)}|\upsilon=1,2,\dots,\Upsilon\}$, $\{\tilde{\Q}^{(\upsilon,\psi)}|\upsilon=1,2,\dots,\Upsilon, \psi=1,2,\dots,\Psi\}$ \\
    \textbf{Output:} ${\M}^g = (x_1^g, \dots, x_K^g)$
	\begin{algorithmic}[1]
	\For {$\upsilon\in\{1,2,\dots,\Upsilon\}$} 
    \State Calculate slopes $\bar{B}^{k,(\upsilon)}_{i}$ via Equation~\eqref{eq:ext_slope}, $i = 1,2,\dots,M^{k,(\upsilon)}$, $k=1,2,\dots,K$
    \State Obtain slope sequence $\PP^\upsilon$ by ranking all slopes in descending order according to the ratio $\bar{B}^{k,(\upsilon)}_{i}/w_k$
    \State $\rho=1$
    \While {$\sum_{k=1}^K w_k x^{(\upsilon)}_k(\rho) < W \land \bar{B}^{(\upsilon)}_{\rho}>0$}
        \State $\rho=\rho+1$
        \State Calculate $x^{(\upsilon)}_k(\rho)$ via Equation~\eqref{eq:ext_xkj}, $k=1,2,\dots,K$
    \EndWhile
    \State  $\rho_S^{(\upsilon)}=\rho$
        \If{$\bar{B}^{(\upsilon)}_{\rho} \le 0$}
            \State $x_k^{g,(\upsilon)}=x_k^{(\upsilon)}(\rho_S^{(\upsilon)}-1)$, $k=1,2,\dots,K$
       	\ElsIf{$\sum_{k=1}^K w_k x^{(\upsilon)}_k(\rho) = W$}
    		\State $x_k^{g,(\upsilon)}=x^{(\upsilon)}_k(\rho_S^{(\upsilon)})$, $k=1,2,\dots,K$
        \Else
            \State $k^\circ=\kappa(\rho_S^{(\upsilon)})$
            \State Obtain maximum fractional requested quantity for resource type $k^\circ$, $\tilde{x}^{(\upsilon)}_{k^\circ}$, via Equation~\eqref{eq:xfrac}
            \State $x_k^{g,(\upsilon)} = 
\begin{cases}
		\lfloor \tilde{x}^{(\upsilon)}_{k^\circ} \rfloor & \text{if } k = k^\circ,
		\\
		x^{(\upsilon)}_k(\rho_S^{(\upsilon)})  & \text{if } k \neq k^\circ,\\
\end{cases}
\quad k =1,2,\dots, K$
        \EndIf
        \EndFor
    \State $x_k^g=\frac{1}{\Upsilon}\sum_{\upsilon=1}^{\Upsilon} x_k^{g,(\upsilon)}, \quad k =1,2,\dots, K$
    \State \Return ${\M}^g = (x_1^g, \dots, x_K^g)$
	\end{algorithmic} 
\end{algorithm}

\clearpage

\subsection{Proof of Theorem~\ref{ext_thm:consis}}
\label{apd:model_ext:thm}

It is easy to prove that function $g^E_k(x_k, T_+,\tilde{Q}_k)$ defined by Equation~\eqref{eq:ext_g_k_explicit} is a continuous, piece-wise linear, and concave function w.r.t $x_k\in  \real^+$ for any given $\tilde{Q}_k$ and $T_+$, $k= 1,2,\dots,K$. In addition, it is clear that $g^E_k(x_k, T_+,\tilde{Q}_k)$ is bounded above by $\sum_{j=1}^{ \tilde{n}^k} \tilde{q}^k_jB^k_{j}(T_+)$. Since $B^k(T_+,t)$ is a monotone convex decreasing function w.r.t $T_+$ and $t$, we have 
\begin{align*}
 g^E_k(x_k, T_+,\tilde{Q}_k)\le \sum_{j=1}^{ \tilde{n}^k} \tilde{q}^k_jB^k_{j}(T_+)\le \sum_{j=1}^{ \tilde{n}^k} \tilde{q}^k_jB^k_{j}(T)\le \sum_{j=1}^{ \tilde{n}^k} \tilde{q}^k_jB^k_{1}(T)=|\tilde{Q}_k|B^k_{1}(T).
\end{align*}

Let $C$ be $[0,\frac{W}{w_1}]\times\dots\times[0,\frac{W}{w_K}]$, which is an nonempty compact set. Next we prove that $\G^E(\M)$ is bounded above and continuous on $C$, and $\G^{\Upsilon,\Psi}(\M)$ converges to $\G^E(\M)$ uniformly on $C$ as $\Upsilon,\Psi\rightarrow\infty$. 

We first show $\G^E(\M)$ is finite on $C$. Denote $g^E(\M, T_+,\tilde{\Q}) \equiv \sum_{k=1}^K g^E_k(x_k, T_+,\tilde{Q}_k)$. Recall that we have shown $ g^E_k(x_k, T_+,\tilde{Q}_k)\le|\tilde{Q}_k|B^k_{1}(T)$. If $|\tilde{Q}_k|\le \frac{W}{w_k}$, we have $ g^E_k(x_k, T_+,\tilde{Q}_k)\le\frac{W}{w_k}B^k_{1}(T)$. Otherwise, note that $g^E_k(x_k, T_+,\tilde{Q}_k)$ is non-decreasing when $x_k\le|\tilde{Q}_k|$. Thus, we have $ g^E_k(x_k, T_+,\tilde{Q}_k)\le\frac{W}{w_k}B^k_{1}(T)$, when $\frac{W}{w_k}< |\tilde{Q}_k|$ and $x_k\in[0,\frac{W}{w_k}]$. Taking together, we have $ g^E_k(x_k, T_+,\tilde{Q}_k)\le\frac{W}{w_k}B^k_{1}(T)$ when $x_k\in[0,\frac{W}{w_k}]$. Moreover, it is obvious that $ g^E_k(x_k, T_+,\tilde{Q}_k)\geq -h^c_k \frac{W}{w_k}$ when $x_k\in[0,\frac{W}{w_k}]$. Thus, we immediately obtain that
\begin{align*}
|g^E(\M, T_+,\tilde{\Q})|&=|\sum_{k=1}^K g_k(x_k,T_+,\tilde{Q}_k)| \leq \sum_{k=1}^K \max\{B^k_{1}(T)\frac{W}{w_k}, h^c_k \frac{W}{w_k}\} = \sum_{k=1}^K \frac{W}{w_k} \max\{B^k_{1}(T), h^c_k\}.
\end{align*}
$g^E(\M, T_+,\tilde{\Q})$ is thus bounded above when $\M\in C$. Therefore, we have
\begin{align*}
|\G^E(\M)|&=|E_{T_+,\tilde{\Q}}(g^E(\M, T_+,\tilde{\Q}))|\le\sum_{k=1}^K \frac{W}{w_k} \max\{B^k_{1}(T), h^c_k\}< +\infty,
\end{align*}
which means $\G^E(\M)$ is bounded above. In addition, the continuity of $\G^E(\M)$ on $C$ is easy to prove by using Lebesgue Dominated Convergence Theorem as shown in the proof of Lemma~\ref{lm:uniformly}. By applying Proposition 7 in \citetsec{shapiro_monte_2003_a}, we can immediately show that $\G^{\Upsilon,\Psi}(\M)$ converges to $\G^E(\M)$ uniformly on $C$ as $\Upsilon,\Psi\rightarrow\infty$. 

Finally, since the feasible regions of problem~\eqref{pb:ext_new} and problem~\eqref{pb:ext_SAA} are both closed subsets of $C$ and their optimal solutions are contained in $C$, we complete the proof of Theorem~\ref{ext_thm:consis} via a direct application of Proposition 6 in \citetsec{shapiro_monte_2003_a}.

\clearpage

\section{Evaluation with Simulated Data}\label{apd:syn}

We employed simulation to evaluate the generalizability of our method. To this end, we simulated demand arrival data for a large number of runs, which enabled us to assess the performance of our and benchmark methods under various demand arrival scenarios. The demands for type 2 resources (e.g., food and water) are normally periodic. That is, the occurrences of these demands are well-dispersed and thus can be modeled by the self-correcting point process \citepsec{schoenberg_introduction_2010_a}. Specifically, the conditional intensity function of the self-correcting point process is given by
\begin{equation}
\label{eq:syn_sc}
    \lambda^{SC}(t)=\exp(\nu t-\sum_{t_i<t}\zeta),
\end{equation}
where $\nu>0$ represents a background trend of demand arrivals and $\zeta>0$ governs the intensity decrease due to existing demands.
Different from type 2 resources, the demands for type 1 resources (e.g., inflatable boat) and type 3 resources (e.g., pump) are generally not periodic.  
The occurrences of these demands are clustered whenever there are emergencies that need these resources. The Hawkes process is a powerful tool for capturing such clustering pattern of demand arrivals \citepsec{rizoiu_hawkes_2017_a}. 
Specifically, it has the following conditional intensity function:
\begin{equation}
\label{eq:syn_h}
    \lambda^{H}(t)=\lambda_0 + \beta \sum_{t_i<t}\exp(-\frac{t-t_i}{\sigma_H}),
\end{equation}
where $\lambda_0>0$ is a deterministic base occurrence rate, $\beta>0$ and $\sigma_H>0$ model the positive impact of early demands on the current intensity.
Having described the simulation of demand arrivals, we next elaborate the simulation of the quantities of requested resources in each demand. Specifically, the requested quantity of type $k$ resources in the $i$-th demand was drawn from a Poisson distribution with mean $\lambda_k^i$, $k=1,2,3$.
We denote ${\boldsymbol\lambda}^i = (\lambda^i_1, \lambda^i_2, \lambda^i_3)$. To simulate the concurrent correlations among different types of resources requested in a demand and the correlation between current demand and historical demands, we drew ${\boldsymbol\lambda}^i$ from a multivariate log-normal distribution: 
\begin{equation}
\label{eq:syn_lognorm}
{\boldsymbol\lambda}^i\sim {\displaystyle \operatorname {Multi-LogNormal} (\frac{1}{2} \log({\boldsymbol\lambda}^{i-1}),\Sigma)}, 
\end{equation}
where $$
{\boldsymbol\Sigma}= \begin{bmatrix}
 \sigma_{1}^2 & \rho_{12}\sigma_{1}\sigma_{2} & \rho_{13}\sigma_{1}\sigma_{3} \\
 \rho_{12}\sigma_{1}\sigma_{2} & \sigma_{2}^2 & \rho_{23}\sigma_{2}\sigma_{3} \\
 \rho_{13}\sigma_{1}\sigma_{3} & \rho_{23}\sigma_{2}\sigma_{3} & \sigma_{3}^2
\end{bmatrix}
$$
is the covariance matrix, $\sigma_{k}=0.5$ for $k=1,2,3$, and ${\boldsymbol\lambda}^0=(2,2,2)$. 
Considering positive correlations between type 2 resources and the other two types of resources, we set $\rho_{12}=\rho_{23}=0.5$. We set $\rho_{13}=-0.5$ because type 1 resources and type 3 resources could be substitutes of each other.
Following the literature \citepsec{xiao_wasserstein_2017_a, xiao_learning_2018_a}, we set $\nu=1$ and $\zeta=0.2$ for the self-correcting point process and $\lambda_0=1$, $\beta=0.8$, and $\sigma_H=1$ for the Hawkes process. 

We conducted the simulation for 200 runs. 
In each run, we simulated three sequences of demands. Each sequence consisted of demands for one resource type in a time period of 48 hours and was simulated according to its corresponding generative process. We then combined the three sequences of demands to obtain a simulated dataset. We employed demands from hour 0 to hour 36 in the dataset to train each compared method and evaluated its performance using demands occurred in (hour 36, hour 48] (i.e., $T=\text{hour } 36$ and $T_+=\text{hour } 48$). The evaluation parameters were set the same as those for the main experiment with the real world data (see Section~\ref{sec:eval:results}).
Table~\ref{tab:syn} reports the average performance of each compared method across 200 simulation runs. 
We excluded the ReR method in this and the rest comparisons because it is a reactive method and its performance is substantially inferior to those of the proactive methods as reported in Section \ref{sec:eval:results}.
As reported, our method significantly outperforms the benchmarks ($p<0.01$). Note the cost reductions by our method reported in this table are for one unit of resource. Considering the huge quantity of  resources needed for disaster response, our method can produce substantial amount of cost reductions. 

\begin{table}[H]
	\TABLE
	{ Performance Comparison with Simulated Data\label{tab:syn}}
	{\begin{tabular}{C{110pt} C{120pt} C{120pt}}
\hline
\textbf{Method}                                             & \begin{tabular}[c]{@{}c@{}}Average Unit Demand\\ Deprivation Cost   \end{tabular} & \begin{tabular}[c]{@{}c@{}}Cost Reduction \\by CNM-PRR \end{tabular} \\ \hline
\begin{tabular}[c]{@{}c@{}}CNM-PRR\\ (Our Method)  \end{tabular} & \$20.32 & \\
  LogNormMix-PRR &\$34.72  & 41.47\% \\
  A-NDTT-PRR & \$30.26 & 32.84\% \\  
  AttnMC-PRR   & \$28.57 & 28.88\% \\ 
  CTDRP-PRR   & \$45.79 & 55.61\% \\ 
  LR-NV  & \$52.19 & 61.06\% \\ 
  DL-NV   & \$45.37 & 55.20\% \\ 
  LogNormMix-IFCFS  & \$88.67 & 77.08\% \\
\hline
\end{tabular}}
	{}
\end{table}

\clearpage

\bibliographystylesec{informs2014}
\bibliographysec{bibs/Project-PRR-ap, bibs/Project-PRR-ap-hz}

\end{appendices}

\clearpage

\end{document}